\documentclass[journal]{IEEEtran}
\usepackage{graphicx} 
\usepackage{subfigure}
\usepackage{color}
\usepackage{soul} 
\usepackage{amsmath}
\usepackage{amsfonts}

\usepackage{changes}
\usepackage{hyperref}

\begin{document}

\title{Adaptively Enhancing Facial Expression Crucial Regions via Local Non-Local Joint Network}

\author{Guanghui~Shi,
	Shasha~Mao,~\IEEEmembership{Member,~IEEE,}
	~Shuiping Gou,~\IEEEmembership{Member,~IEEE,} Dandan Yan,~Licheng~Jiao,~\IEEEmembership{Fellow,~IEEE,}
	and~Lin~Xiong 
	\thanks{Manuscript received by \textbf{\textit{Machine Intelligence Research}}
	}
	\thanks{The paper can be accessed via \href{https://link.springer.com/article/10.1007/s11633-023-1417-9}{https://link.springer.com/article/10.1007/ s11633-023-1417-9}
	}
	\thanks{DOI: 10.1007/s11633-023-1417-9}
}

\markboth{Journal of \LaTeX\ Class Files,~Vol.~14, No.~8, August~2015}%
{Shell \MakeLowercase{\textit{et al.}}: Bare Demo of IEEEtran.cls for IEEE Journals}

\maketitle

\begin{abstract} 	
	Facial expression recognition (FER) is still one challenging research due to the small inter-class discrepancy in the facial expression data. In view of the significance of facial crucial regions for FER, many existing researches utilize the prior information from some annotated crucial points to improve the performance of FER.
	However, it is complicated and time-consuming to manually annotate facial crucial points, especially for vast wild expression images. Based on this, a local non-local joint network is proposed to adaptively light up the facial crucial regions in feature learning of FER in this paper. In the proposed method, two parts are constructed based on facial local and non-local information respectively, where an ensemble of multiple local networks are proposed to extract local features corresponding to multiple facial local regions and a non-local attention network is addressed to explore the significance of each local region. Especially, the attention weights obtained by the non-local network is fed into the local part to achieve the interactive feedback between the facial global and local information. Interestingly, the non-local weights corresponding to local regions are gradually updated and higher weights are given to more crucial regions. Moreover, U-Net is employed to extract the integrated features of deep semantic information and low hierarchical detail information of expression images. Finally, experimental results illustrate that the proposed method achieves more competitive performance compared with several state-of-the-art methods on five benchmark datasets. Noticeably, the analyses of the non-local weights corresponding to local regions demonstrate that the proposed method can automatically enhance some crucial regions in the process of feature learning without any facial landmark information.
\end{abstract}


\IEEEpeerreviewmaketitle

\section{Introduction}
\IEEEPARstart{E}{motion} is a complex state that integrates people's feelings, thoughts and behaviors \cite{1998expression}, and facial expression is one of the most direct signals to communicate their innermost thoughts. Therefore, facial expression recognition (FER) \cite{buck1974sex,smith1996spontaneous,corneanu2016survey,majumder2016automatic,xie2019adaptive} has attracted the attention of many researchers due to its important role in many practical application fields, such as human-computer interaction, recommendation system, patient monitoring, et al..
In general, facial expression is encoded into facial action units through facial action coding system \cite{AU,ekman2002facial,wang2018weakly}, and any expressions can be described through a set of facial action units.
As we know, some facial action units are crucial for FER \cite{ekenel2009facial}, such as the one located in regions around eyes and the mouth, since they are of more obvious actions compared with other facial regions (such as cheek and forehead).
In the following parts, we regard these crucial facial action units as facial crucial regions, shortened by FCRs. Fig.1 illustrates facial crucial regions of two facial images (ID1 and ID2) from six expressions, respectively. 
From Fig.1, it is found that the FCRs are more discriminative to determine the expression category of a facial image \cite{zhong2014learning}.

\begin{figure}[t]
	\centering
	\includegraphics[scale=0.45]{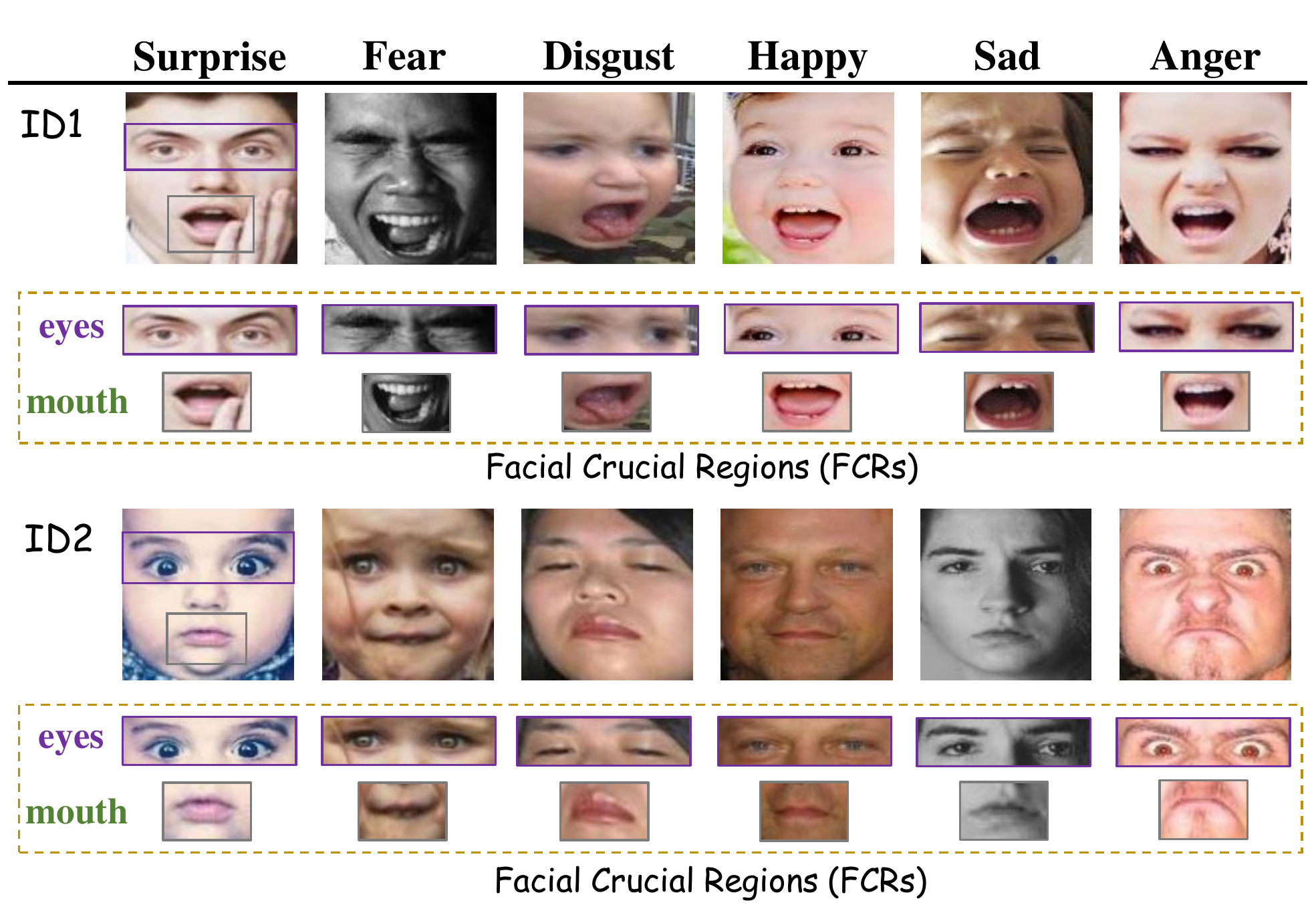}
	\caption
	{An illustration of facial crucial regions from six expressions, where two facial images (ID1 and ID2) are shown for each expression. The regions around eyes and mouths are cropped as examples of FCRs in the purple box and the green box, respectively.
	}\label{fig0}
\end{figure}
In view of the significance of FCRs, many studies \cite{fan2018multi,li2018occlusion,6998925,8974606} have been proposed based on applying the information of facial local regions, where the facial landmarks are employed as the prior information of facial crucial regions, whereas the landmarks are given by manually annotating for facial expression images. 
Early, most of FER researches \cite{Liu_2014_CVPR,shan2009facial,kotsia2006facial} focused on lab-collected expression datasets, such as CK+ \cite{CK}, MMI \cite{MMI}, JAFFE \cite{lyons1998japanese}, Oulu-CASIA \cite{zhao2011facial}. 
For lab-collected datasets, facial expressions images were collected from several or dozens of individuals under similar conditions (such as illumination, angle, posture, et al.), generally with a few uncontrollable factors.
Thus, it is easily achieved to manually annotate the landmark of FCRs for lab-collected datasets.

\begin{figure}[t]
	\centering
	\subfigure[]{
		\begin{minipage}[t]{0.25\linewidth}
			\centering
			\includegraphics[width=0.9\columnwidth]{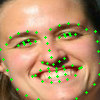}
		\end{minipage}%
	}
	\subfigure[]{
		\begin{minipage}[t]{0.25\linewidth}
			\centering
			\includegraphics[width=0.9\columnwidth]{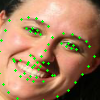}
		\end{minipage}%
	}
	\subfigure[]{
		\begin{minipage}[t]{0.25\linewidth}
			\centering
			\includegraphics[width=0.9\columnwidth]{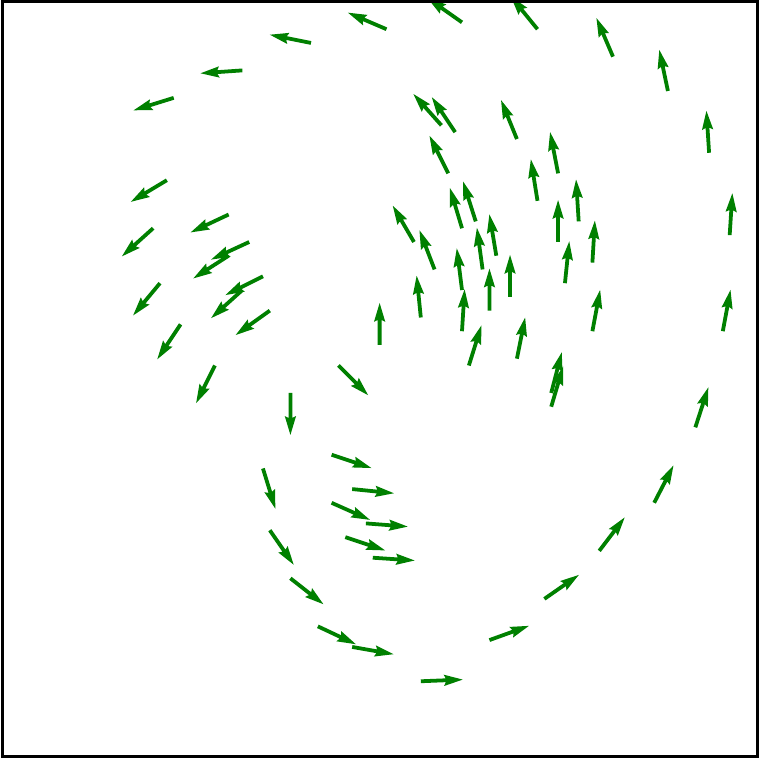}
		\end{minipage}%
	}
	\caption{Schematic diagram of the pixel deviations at image level when posture changing. To demonstrate this change, we measured the movement of 68 landmark points on the faces with different postures and the same identity. In figure (a) and (b), 68 landmark points are marked with a green cross, and figure (c) shows the movement of 68 landmark points.}
	\label{fig11}
\end{figure}

However, compared with the lab-controlled datasets, the wild expression datasets \cite{li2020deep} are collected under more complex and uncontrollable conditions, such as RAF-DB \cite{Li_2017_CVPR}, AffectNet \cite{AffectNet}, EmotionNet \cite{fabian2016emotionet}, et al.
For the wild expression datasets, especially including a vast of images, it is very complicated and time-consuming for manually annotating FCRs. 
Moreover, the postures of different faces vary greatly on the wild database.
One simple change of facial postures can cause multiple pixel deviations at the image level.
Fig.\ref{fig11} gives an example about the landmarks moving with the change of postures, where two expression images and their landmarks are from RAF-DB dataset \cite{Li_2017_CVPR}.
From Fig.\ref{fig11}, it is obvious that 68 landmark points of the image (a) are different from the image (b) and the landmarks are greatly shifted from (a) to (b), shown as the figure (c).
It implies that the position of FCRs varies with the change of facial postures. 
Inevitably, it increases the complexity of manually annotating landmarks for FER, especially for the wild dataset with a vast of images.
In view of this, it is considerable that whether the significance of FCRs or their features could be spontaneously enhanced in the training of deep FER or not, without any prior information, such as landmarks of FCRs. 

On the other hand, there exists a problem that some FCRs from different expression categories are similar{\color{blue}{,}} whereas some FCRs from one same category are very different.
From Fig.\ref{fig0}, it is obviously seen that the FCRs (including mouths) of ID1 from six expressions are similar with opening the mouth, which is absolutely different from ID2 with closing the mouth.
Similarly, for the crucial regions including eyes, ID1 and ID2 from the category (Fear) are different, whereas ID1 from the category (Surprise) and ID2 from the category (Anger) are similar.
It illustrates that FCRs of expression images belonging to the same category may be very different but FCRs from different categories are similar.
Distinctly, it is insufficient that only local information of facial expressions is utilized to construct one effective model for FER, especially for the wild dataset.
Hence, it is still important to utilize the global information of the facial expression while FCRs are enhanced in deep facial expression recognition.

\begin{figure}[t]
	\centering
	\includegraphics[scale=0.35]{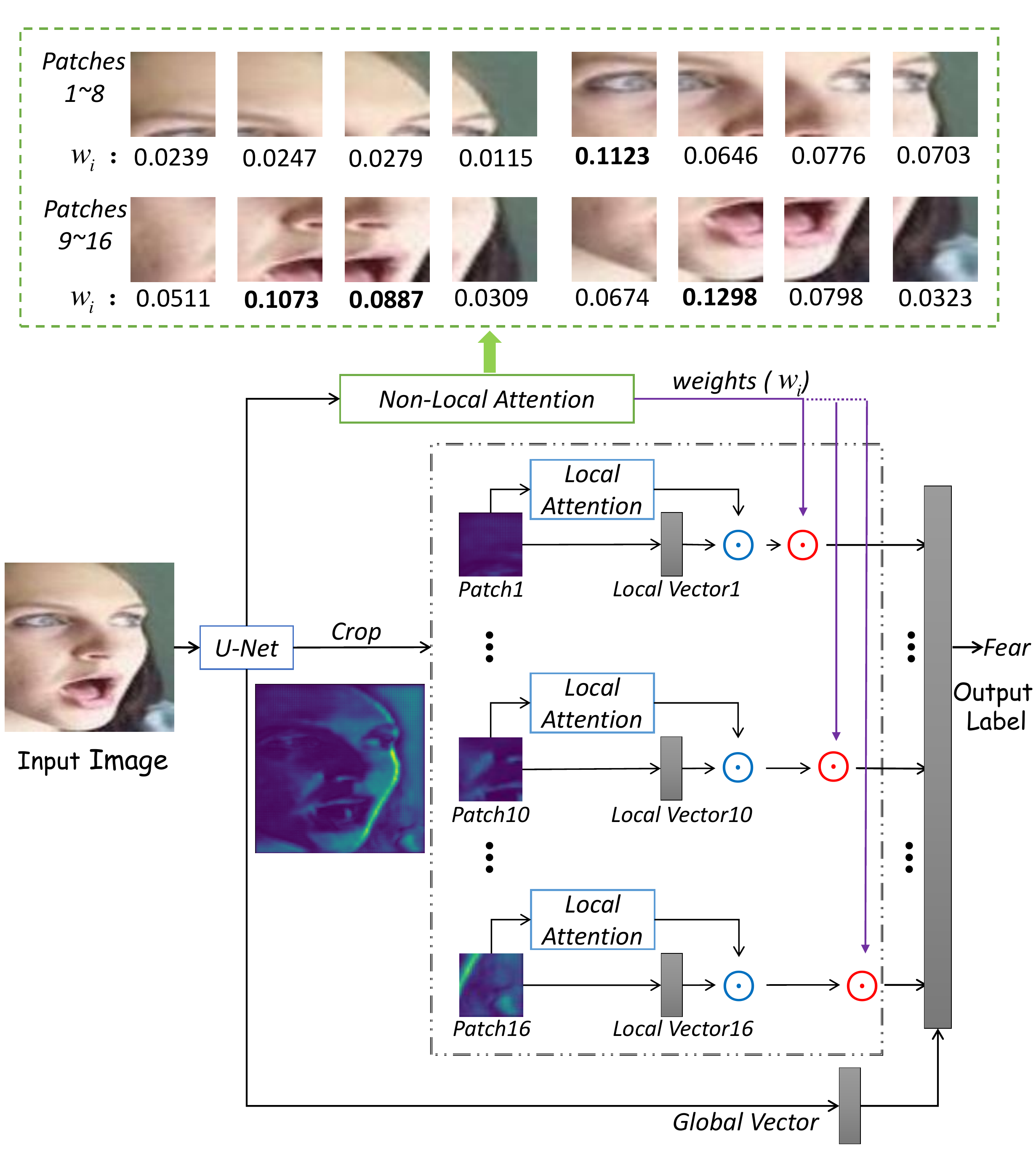}
	\caption
	{A simple view of the proposed model (LNLAttenNet). The part in the green dotted box shows the global weights corresponding to 16 local regions (from Patch 1 to Patch 16) obtained by LNLAttenNet, and the part under the green dotted box is a simple framework of LNLAttenNet.
	}\label{fig3}
\end{figure}

Based on the above analyses, we propose a new method of facial expression recognition in this paper, which constructs a local non-local joint network to adaptively enhance the facial crucial regions in the process of deep feature learning, shortened for LNLAttenNet.
In LNLAttenNet, the local and the non-local information of facial expressions are simultaneously considered to construct two parts of the network respectively: a local multi-network ensemble and a non-local attention network, and then the generated local and non-local feature vectors are integrated and jointly optimized in feature learning.
Specially, the attention weights obtained by the non-local part is regarded as the significance of facial local regions and fed into the local multi-network ensemble system to combine multiple local networks. Interestingly, we find that some facial crucial regions can be automatically enhanced in the process of deep feature learning by the proposed method. 
Moreover, U-Net is employed to generate feature maps where each pixel has large receptive field and the local region also contains the global information.
Fig.\ref{fig3} shows a simple view of LNLAttenNet. From Fig.\ref{fig3}, it is obvious that some crucial regions is given higher weights by LNLAttenNet, such as the 5th patch around the left eye (0.1123), the 10th, 11t and 14th patches around the mouth (0.0887, 0.1073 and 0.1298), which illustrates that some crucial regions are effectively enhanced by LNLAttenNet. Note that $w_i$ is the non-local attention weight corresponding to the $i^{th}$ local region and the initial weights are equal. More detailed descriptions will be introduced in the following parts.


Compared with stat-of-the art methods, our contributions are mainly three points:
\begin{itemize}
	\item We propose LNLAttenNet to automatically light up facial crucial regions in deep feature learning by utilizing the local and non-local information of facial expression simultaneously.
	To the best of our knowledge, it is the first work to study whether FCRs is directly explored and enhanced in feature learning of deep FER, where FCRs are automatically enhanced without any prior information for facial crucial regions or points. It effectively improves the problem that difficultly annotating for the wild dataset with a vast of facial images.

	\item In LNLAttenNet, an attention mechanism is introduced to construct the non-local attention network which explores the significance of local regions for FER from a global perspective of facial expression. The obtained attention weights corresponding to local regions are fed into the local multi-network ensemble system to integrate multiple local features, and then the integration of features obtained by multiple local networks is jointly optimized with the facial global feature. 
	
	\item Experimental results demonstrate that FCRs can be enhanced in deep feature learning by LNLAttenNet, which validates FCRs are exactly more discriminative local regions for FER. Moreover, it also implies that the model of deep FER can spontaneously focus on some crucial regions in the training process, which probably brings a new inspiration for designing deep FER methods. 
\end{itemize}

\begin{figure*}[t]
	\centering
	\includegraphics[width=0.95\textwidth]{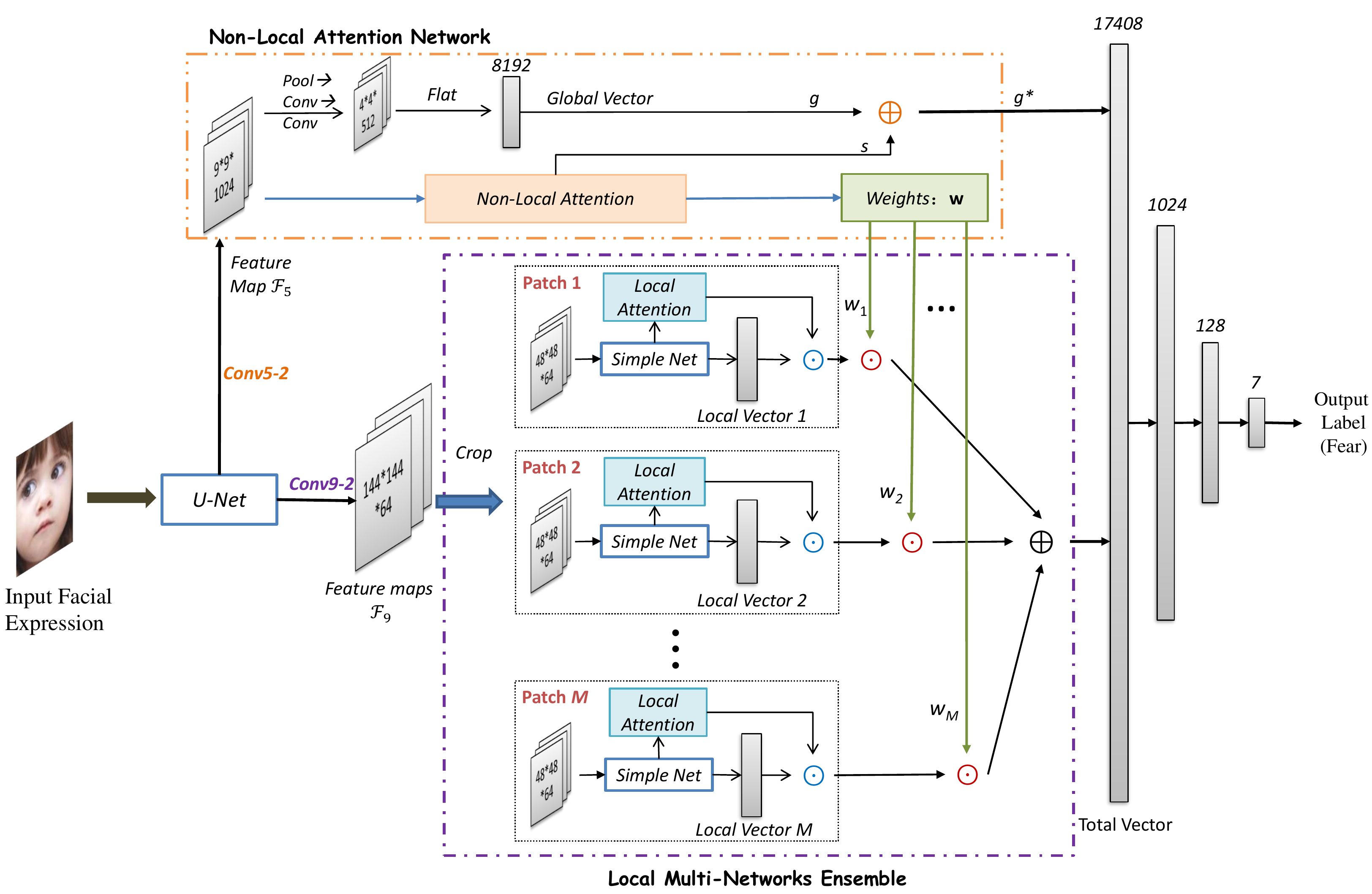}
	\caption{The framework of the proposed model (LNLAttenNet). LNLAttenNet uses U-Net to generate feature map with the same resolution as the input image. Then, its feature map (Conv9-2) is cropped into $M$ local patches to construct the local multi-networks ensemble model, where each patch is used to generate an individual network based on the structure of Simple Net. The feature map (Conv5-2) is used to construct the global attention network. Finally, the global and local features are integrated based on the global weights, and then three fully connected layers are followed.}
	\label{Fig2ProposedModel}
\end{figure*}

The rest of this manuscript is organized as follows. Section \ref{relatedworks} firstly introduces related works about deep facial expression recognition. Secondly, Section \ref{ProposedModelSection} introduces the detail of the proposed method. Then, experimental results and analyses are demonstrated to validate the performance of the proposed method in Section \ref{Experiments}. Finally, Section \ref{Conclusion} provides the conclusion as well as the prospects on future works.

\section{Related Works}\label{relatedworks}
Due to the excellent performance of deep learning, various deep networks have been applied in FER \cite{li2020deep}, such as VggNet \cite{VGG}, InceptionNet \cite{Inception}, ResNet \cite{Resnet}, et al. Based on this, many deep FER methods have been proposed to address different problems.
In \cite{hu2017ACM}, Hu \textit{et al.} firstly extended the idea of deep supervision to deal with FER in the wild.
The training of deep CNNs was softer and easier through the supervision not only to deep layers but also to intermediate layers and shallow layers, and a fusion structure was constructed where the feature ahead was used for the second-level supervision.
In \cite{Covariance_pool}, Acharya \textit{et al.} thought that the second-order statistic (such as covariance) were more suitable to catch the feature of the twisted facial expression. In their framework, a mainfold structure was constructed for covariance pooling to obtain a competitive performance for FER.
In \cite{li2019blended}, Li \textit{et al.} proposed a new deep manifold strategy for multi-label expressions,
and their proposed network focused on the ambiguity expressions and could learn the discriminative feature that was suitable for cross-database FER.

Considering that facial expression is determined by key regions, Fan \textit{et al.} \cite{fan2018multi} utilized the information of facial landmark points to select three sub-images around the eyes, mouth and nose. Then, three sub-images were encoded by three sub-networks, and the last pooling layer in each sub-network was concatenated with each other, which obtained better recognition performance compared with others.
In \cite{8273654,8545725}, the information of facial landmark is used to extract features and generate masks from specific locations to remove the pose variation.

In \cite{IPA2LT}, it was taken into account that there are inevitably labeling errors and deviations between different databases due to the subjectivity of labeling facial expressions.
Therefore, when existing methods make use of multiple databases to expand the training set, their performance cannot be continuously improved.
In order to solve this inconsistency between different databases, an Inconsistent Pseudo Annotations to Latent Truth (IPA2LT) framework is proposed to train a model from multiple inconsistent databases and large scale unlabeled images.
The IPA2LT essentially constructs the ensemble at label level.
Each image in the model has the same number of labels as the number of data sources, in which only one label is original and others are pseudo.
Existing methods for FER have been almost satisfying on analyzing the frontal faces but fail to attain a good performance on partially occluded faces collected in the wild.
Some facial expressions are ambiguous and have multi-labels.
In \cite{gan2019facial}, Gan \textit{et al.} proposed a new framework based on CNN with the supervision of soft labels, where hard labels are used to construct soft labels with a novel label-level perturbation.
In this framework, soft labels were obtained to eliminate the similarity between faces of different emotions, and multiple basic classifiers were trained and then combined.
Moreover, some GAN-based methods have been proposed to generate expressional images for FER \cite{De-Gan_2018_CVPR, Joint-GAN_2018_CVPR, CycleEmotionGAN_2019_AAAI} or usually focus only on generating new facial expression images \cite{ImgtoVid_2019_AAAI, EF-GAN_2020_CVPR, GANimation_2018_ECCV, StarGAN_2018_CVPR}.
In \cite{De-Gan_2018_CVPR}, a novel approach is proposed to learn facial expressions by extracting the expressive component through a de-expression procedure where the corresponding neutral expression is generated by the trained generative model by given a facial image with arbitrary expressions.
In \cite{ImgtoVid_2019_AAAI}, a user-controllable approach is proposed so as to generate video clips of various lengths from a single face image and the lengths and types of the expressions are controlled by users.

In \cite{li2018occlusion}, Li \textit{et al.} proposed a CNN with attention mechanism (ACNN) to detect the occlusion of facial regions and paid attention to the most discriminative regions, where ACNN used the information of 24 facial landmark points to select the key regions at the feature level.
In \cite{barros2017emotion}, Barros \textit{et al.} investigated the emotion-driven attention mechanisms from the view of videos.
In \cite{wang2018two}, Wang \textit{et al.} proposed two-level attention mechanism to extract emotion-related features, which was based on global information, not involving the local regions.
Similarly to \cite{barros2017emotion,wang2018two,li2018occlusion}, the attention mechanism is also involved in this work, whereas the essence of algorithms is very different.
Here, our purpose is to adaptively enhance the significance of facial crucial regions based on the attention weights in feature learning obtained by the non-local attention network from the view of multiple local regions, where the attention weights corresponding to each local regions are obtained by the non-local attention network.

\section{Local Non-Local Joint Network for Facial Expression Recognition}\label{ProposedModelSection}
In this paper, we propose a Local Non-Local Attention Joint Network for FER to adaptively light up more crucial local regions of facial expression, named by LNLAttenNet.
The overall framework of LNLAttenNet is visually shown in Fig.\ref{Fig2ProposedModel}.
In Fig.\ref{Fig2ProposedModel}, one facial expression image is used as the initial input instance of the proposed network, and its size is 144$\times$144 as same as our implemented experiments.

\begin{figure}[t]
	\centering
	\includegraphics[width=1.02\columnwidth]{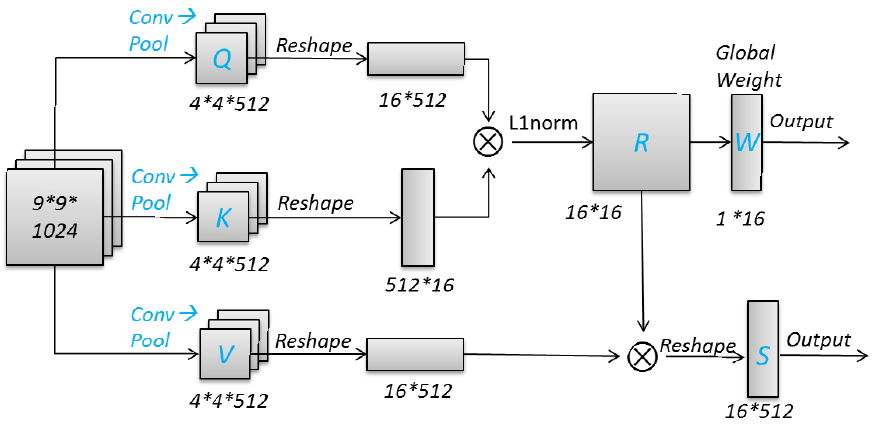}
	\caption{Overview of the Non-Local attention model.} 
	\label{Fig5GlobalAttention}
\end{figure}

In LNLAttenNet, U-Net is firstly employed to extract the feature maps integrating the deep semantic information and the low hierarchical detail information of facial expression images.
For the facial expression dataset, when the regional integration is carried out \cite{fan2018multi}, the inter-class discrepancy is smaller and the intra-class discrepancy is larger, as shown in Fig.\ref{fig0}.
The structure of U-Net \cite{ronneberger2015u,falk2019u,8309343}, the top-down architecture with lateral connections for introducing details into high-level semantic feature maps, has been proved that local regions in last few layers are of the large receptive field and the global information, which is important and useful for ambiguous objects recognition \cite{FPN,FCN}. Therefore, U-Net is beneficial to alleviate the negative impact of the regional integration, but it does not mean that the proposed method is restricted to U-Net.
Actually, one model with the similar structure to U-Net can be employed in our proposed method, such as FPN \cite{FPN}.

As shown as Fig.\ref{Fig2ProposedModel}, facial expression images are inputted to the proposed model. By U-Net, two different feature maps are generated for the initial input image, located in the last layer (Conv9-2) and the intermediate layer (Conv5-2) of U-Net, respectively. In the following parts, we use $\mathcal{F}_5$ and $\mathcal{F}_9$ to express the feature maps from Conv5-2 and Conv9-2 of U-Net, respectively.
Then, the generated feature maps $\mathcal{F}_5$ and $\mathcal{F}_9$ are utilized to construct two parts of LNLAttenNet, where the map $\mathcal{F}_5$ is utilized as the input to construct the non-local part (the {\textsl{Non-Local Attention Network}}) and the map $\mathcal{F}_9$ is employed as the input to construct the local part (the {\textsl{Local Multi-Networks Ensemble System}}).
In the local part, an ensemble of multiple networks is applied to generate and integrate multiple individual networks corresponding to different facial local regions respectively.
By the non-local attention network, an attention weight $w_i$ ($i=1,...,M$) is obtained corresponding to the $i^{th}$ local region of the facial expression, and then the vector $\bf{w}$ ($[w_1,...,w_M]^T$) are used as the weights of multiple local networks to combine $M$ local vectors and meanwhile boost the significance of local regions in the process of deep feature learning.
Finally, the non-local attention network and the local ensemble network are jointly optimized by integrating local and non-local features in three fully connected layers of LNLAttenNet.
More detailed descriptions of the proposed method will be introduced as follows.

\subsection{Non-Local Attention Network}
For facial expression recognition, there is small inter-class discrepancy and large intra-class discrepancy on expression images, as shown in Fig.\ref{fig0}.
Therefore, facial crucial regions are regarded as more discriminative regions which determine the categories of facial expression, such as regions around the mouth (eyes) rather than the cheek.
However, it is tough to estimate which regions are more crucial without the assistance from manually annotated crucial points.
Based on this, we construct the {\textsl{Non-Local Attention Network}} to automatically mine more discriminative regions from the whole facial expression, visually shown in the box with orange dot lines of Fig.\ref{Fig2ProposedModel}.

In Fig.\ref{Fig2ProposedModel}, the feature map $\mathcal{F}_5$ (Conv5-2) is generated by U-Net as the global information of the facial image to construct the non-local attention network.
The Conv5-2 is with the minimum resolution and the maximum receptive field, which means that $\mathcal{F}_5$ is not affected by each local patch but contains the relationship between local patches implicitly.
It is useful to mine more crucial regions based on the global information from the whole face.


\begin{figure}[t]
	\centering
	\includegraphics[width=0.95\columnwidth]{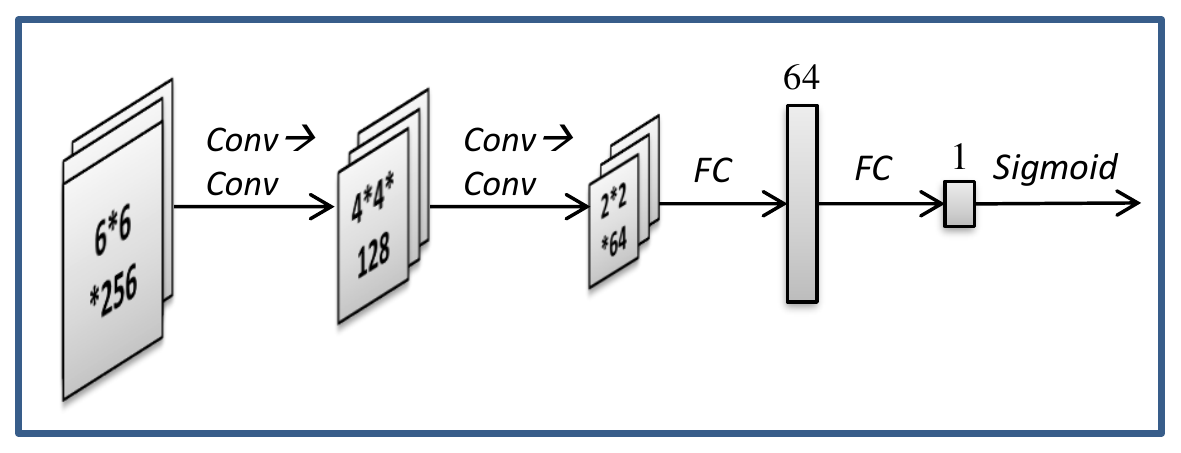}
	\caption{Overview of the local attention.} 
	\label{Fig5LocalAttention}
\end{figure}

\subsubsection{Global Attention}
Inspired by \cite{First_atte,non_local}, we construct a non-local attention model based on three branches, shown as in Fig.\ref{Fig5GlobalAttention}.
First, the input is the map $\mathcal{F}_5$ containing the global information of facial expression in Fig.\ref{Fig5GlobalAttention}.
Based on $\mathcal{F}_5$, three feature maps $\mathcal{Q}$, $\mathcal{K}$ and $\mathcal{V}$ are generated by one convolution layer and one pooling layer, respectively.
Note that three maps are with a special resolution{\footnote{
		This special resolution is set in order to expediently calculate the correlation between each patch. For example, when the number of cropped local regions is set as 16 ($M=16$) in our experiments, the special resolution is $4*4$ ($n=4$), as shown in Fig.\ref{Fig5GlobalAttention}.
}}
with $n*n$ in this model, where $M=n^2$ and $M$ is the number of cropped local regions.
Then, the maps $\mathcal{Q}$ and $\mathcal{K}$ are reshaped as $\bf{Q}^*$ and $\bf{K}^*$, shown as in Fig.\ref{Fig5GlobalAttention}, and a multiplication operation is followed to get a matrix $\bf R$ which reflects the correlation among local regions.
Compared with \cite{First_atte,non_local}, the relevance of each region (patch) in LNLAttenNet is not as strong as each frame in video or each word in sentence, and thus $L_1$ normalization is adopted to limits the sum of each row of $R$ to 1 instead of the softmax function.
Finally, a vector is calculated via averaging the each column of the correlation matrix $\bf R$, regarded as the non-local attention weights $\bf{w}^g$ assigned to $M$ local regions.

Furthermore, the map ${\bf{V}}$ is reshaped as ${\bf{V}}^*$, and the feature vector ${\bf{s}}$ is obtained by multiplying ${\bf{V}}^*$ by the correlation matrix ${\bf{R}}$, which is the self-attention form in \cite{First_atte,non_local}. In order to make the matrix ${\bf{R}}$ reflect the correlation among local regions, ${\bf{s}}$ is flattened and added to the non-local vector ${\bf{g}}$ (shown in Fig.\ref{Fig2ProposedModel}).
Meanwhile, a function is given to trade off two vectors ${\bf{g}}$ and ${\bf{s}}$, shown as
\begin{equation}
{\bf{g}}^* =(1-\alpha)\cdot {\bf{g}}+\alpha\cdot flat({\bf{s}}),
\end{equation}
where ${\bf{g}}^*$ expresses the new non-local vector and $\alpha$ is the hyper-parameter to adjust the ratio of ${\bf{s}}$.
In experiments, we will give an analysis for the parameter $\alpha$.

\subsection{Local Multi-Networks Ensemble}
\begin{figure}[t]
	\centering
	\includegraphics[width=1\columnwidth]{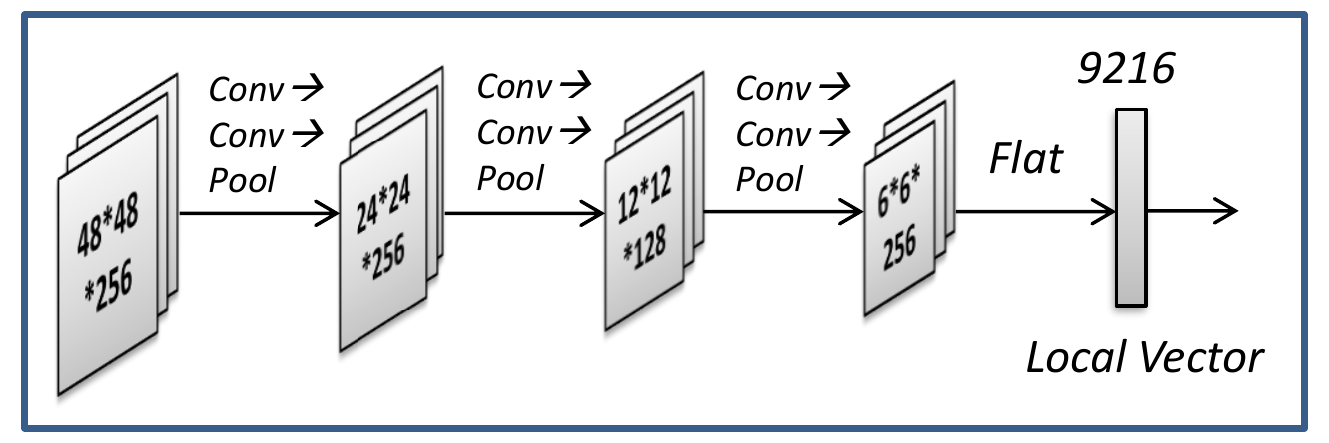}
	\caption{The structure of Simple Network}
	\label{FigSimpleNet}
\end{figure}

The feature map ($\mathcal{F}_9$) is employed as the input to construct the part: {\textsl{Local Multi-Networks Ensemble}}, shown as in Fig.\ref{Fig2ProposedModel}.
The reason of using the map $\mathcal{F}_9$ is that each pixel is of the large receptive field and the rich sementic information in Conv9-2, where $\mathcal{F}_9$ is with the same resolution as the initial input image.
In the part of Local Multi-Networks Ensemble, the feature map $\mathcal{F}_9$ is firstly divided into $M$ patches (including different local regions) with the same dimension (set as 48*48*64 in our experiments).
Then, $M$ patches are trained by Simple Network{\footnote{The basic structure of Simple Network is shown in Fig.\ref{FigSimpleNet}, composed of six convolution layers and three pooling layers.}} to generate $M$ individual networks $\{{\mathcal{IN}}_1,...,{\mathcal{IN}}_M\}$, respectively.
Specially, for each individual network, the local attention mechanism is added to enhance the feature vector of each local region.
Finally, $M$ local feature vectors are combining with the non-local attention weights obtained by Non-Local Attention Network.

%

\subsubsection{Local Attention}
In practice, it is found that the useful information is decreased when partial regions in one patch are missed or obscured. It means that less attention should be given to them.
In view of this, a local attention mechanism is adopted in each individual network to weaken the significance of useless regions.
The local attention model is encoded by four convolution layers and two fully connected layers, and its structure is shown in Fig.\ref{Fig5LocalAttention}.
Note that two convolution layers are not padded in order to reduce the computational complexity.
In the local attention model, its input is the output of the last pooling layer in Simple-Net,
and its output is one value between 0 and 1 obtained via the sigmoid function, regarded as the local attention weight $w_i^l$ of each individual network, which represents the amount of information in each local patch can flow to the next level.
If the facial local region is obscured or missed, the information that it contains for expression recognition will be reduced, and then the weight value of the local attention is also reduced to alleviate the effect of patches including the obscured region. Furthermore, the weights will be multiplied by the corresponding local vector as the output feature of each local network.
More visual illustrations can be found in the part of experiments.


\begin{table*}[ht]
	\begin{center}
		\begin{tabular}{c|c|c|c|c|c|c|c|c}
			\hline
			\multicolumn{3}{c|}{$Q$} & \multicolumn{3}{|c|}{$K$} &\multicolumn{3}{|c}{$V$}\\
			\hline
			Operation & Activate    & Output shape &Operation & Activate    & Output shape &Operation & Activate    & Output shape \\
			\hline
			Conv 1$\times$1  s:1& ReLu& 9*9*512  &Conv 1$\times$1  s:1  & ReLu & 9*9*512   &Conv 1$\times$1  s:1 & ReLu  & 9*9*512 \\
			MaxPooling 2$\times$2 s:2   &-  & 4*4*512 & MaxPooling 2$\times$2 s:2   &-  & 4*4*512 &MaxPooling 2$\times$2 s:2   &-  & 4*4*512\\
			Reshape &- &16*512  &Reshape &- &512*16  &Reshape &- &16*512  \\
			\hline
			
		\end{tabular}
	\end{center}
	\caption{The structure of Non-Local attention.}
	\label{non-local-atte}
\end{table*}

\subsubsection{Combination of Multiple Local Networks}
According to the non-local attention weights ${\bf{w}}^g$ and the local attention weights ${\bf{w}}^l$, the local feature vectors given by $M$ individual networks $\{{\mathcal{IN}}_1,...,{\mathcal{IN}}_M\}$ are aggregated by the formula

\begin{equation}
{\bf{f}}_{en}=\sum_{i=1}^{M} w_i^g*w_i^l{\bf{f}}_i,
\end{equation}
where ${\bf f}_{en}$ expresses the ensemble feature vector, ${\bf{f}}_i$ expresses the feature vector given by ${\mathcal{IN}}_i$ corresponding to the $i^{th}$ local region, $w_i^g$ is the non-local attention weight of the $i^{th}$ local region, and $w_i^l$ expresses the local attention weight of the $i^{th}$ local region.
In experiments, we will give an analysis for the number $M$ of local patches.

\subsection{Joint Optimization of LNLAttenNet}
In Fig.\ref{Fig2ProposedModel}, the non-local feature vector ${\bf{g}}^*$ is produced by the non-local attention network, and the local vector ${\bf{f}}_{en}$ is obtained by the local multi-network ensemble.
Inspired by \cite{PSPN}, we think that the global information of an input image is essential, and each local patch can get large receptive field and the global information by embedding U-Net, which makes it easier to classify the similar patch of facial expression of different categories.
Moreover, Conv5-2 is encoded to a global vector with 8192 dimension by two convolution layers and one pooling layer.
Then, the non-local vector ${\bf{g}}^*$ is concatenated with the local vector ${\bf{f}}_{en}$ to obtain the total vector as the feature of the first fully connected layer and is jointly optimized, and the dimension of the integrated feature vector is 17408 shown as in Fig.\ref{Fig2ProposedModel}.
In LNLAttenNet, three full connect layers are implemented, 
and the loss function is formulated as
\begin{equation}
L = loss_{entropy} + \gamma loss_{l2},
\end{equation}
where $loss_{entropy}$ expresses the cross entropy loss, $loss_{l2}$ is the l2 regularization loss, and $\gamma$ is the hyper-parameter controlling the balance between two losses.
The cross entropy is calculated as:
\begin{equation}
loss_{entropy} =\frac{1}{N} \sum_{n=0}^{N-1} \sum_{c=0}^{C-1} \mathbb{L}(l_n = c ) \cdot log(p_n^i),
\end{equation}
where $C$ is the number of categories, $N$ is the number of the input image, and $\mathbb{L}$ is the function that determines whether the input is correct.
$p_n^i$ is the $i^{th}$ component of the output of the last softmax layer of the $n^{th}$ image,
and $l_n$ is the label of the $n^{th}$ input image.
The l2 regularization loss is computed by $loss_{l2} = \lambda \cdot {||W||}^2$, where $W$ is the parameters of our model and $\lambda$ is set as 0.0001 in the following experiments.

\section{Experiments and Analyses}\label{Experiments}
In this section, we will validate the performance of the proposed method from several items:
1) the performance comparison with state-of-the-art methods on benchmark datasets, 2) the analyses of Non-Local Attention, 3) the visualization of Local Attention, 4) the change of the parameter $\alpha$, 5) the performance of LNLAttenNet with different $M$, and 6) the analyses for overlapped pixels between local regions, respectively.

\begin{table}[t]
	\begin{center}
		\begin{tabular}{c|c|c}
			\hline
			Operation                  & Activate    & Output shape  \\
			\hline
			Conv 3$\times$3  s:1         & ReLu        & 48*48*64  \\
			Conv 3$\times$3  s:1          & ReLu        & 48*48*64  \\
			MaxPooling 2$\times$2 s:2    &-            & 24*24*64\\
			\hline
			Conv 3$\times$3  s:1         & ReLu        & 24*24*128  \\
			Conv 3$\times$3  s:1          & ReLu        & 24*24*128  \\
			MaxPooling 2$\times$2 s:2    &-            & 12*12*128\\
			\hline
			Conv 3$\times$3  s:1         & ReLu        & 12*12*256  \\
			Conv 3$\times$3  s:1          & ReLu        & 12*12*256  \\
			MaxPooling 2$\times$2 s:2    &-            & 6*6*256\\
			\hline
		\end{tabular}
	\end{center}
	\caption{The structure of SimpleNet.}
	\label{SimpleNet}
\end{table}

\begin{table}[t]
	\begin{center}
		\begin{tabular}{c|c|c}
			\hline
			Operation                  & Activate    & Output shape  \\
			\hline
			Conv 3$\times$3  s:1 No padding         & ReLu        & 4*4*256  \\
			Conv 1$\times$1  s:1           & ReLu        & 4*4*128  \\
			\hline
			Conv 3$\times$3  s:1 No padding         & ReLu        & 2*2*128  \\
			Conv 1$\times$1  s:1           & ReLu        & 2*2*64  \\
			Reshape &- & 256\\
			\hline
			Full connect & - & 64 \\
			Full connect & - & 1  \\
			Sigmoid      & - & 1\\
			\hline
		\end{tabular}
	\end{center}
	\caption{The structure of local attention.}
	\label{Local-atte}
\end{table}

\subsection{Databases and Setups}
In experiments, we employ five FER datasets to evaluate the performance of LNLAttenNet: RAF-DB \cite{Li_2017_CVPR}, SFEW \cite{SFEW}, AffectNet \cite{AffectNet}, CK+ \cite{CK} and MMI \cite{MMI}.
\begin{itemize}
	\item \textbf{RAF-DB} contains 29672 facial images downloaded from the Internet. For the RAF-DB dataset, the facial landmarks are manually annotated via the crowdsourcing method with basic or compound expressions. In experiments, we use the basic database including 12,271 training and 3,068 testing images.
	\item \textbf{SFEW} contains the statistic images selected from the movie clips with spontaneous expressions, where the labels of training set and validation set are given. Therefore, 958 training images are used as the training set and 436 validation images are as the testing set in experiments.
	\item \textbf{AffectNet} contains 450,000 images with 10 categories, where each image is annotated by one volunteer. In experiments, we use 287,401 images with neutral and six basic emotions, where 283,901 images are selected as the training set and 3,500 images are selected from the validation set as the testing set.
	\item \textbf{CK+} contains 593 sequences from 123 volunteers, where 309 sequences have been annotated with six basic emotions. The emotion in each sequence goes from neutral to peak and then to neutral again. In view of this, we select the first frame of each sequence with the label of neutral and the peak frame of each sequence with the target label to generate 618 experimental images.
	\item \textbf{MMI} is recorded from 30 objects with rich details of annotations, and 398 images are generated by selecting the first frame of each sequence with the label of neutral and one peak frame of each sequence.
\end{itemize}


For RAF-DB and SFEW datasets, their training sets are directly used to train the model and testing sets are used to evaluate the performance.
For AffectNet dataset, its training set is used to train the model, and its validation set is used as the testing set, since the testing set of AffectNet is not given the annotated labels \cite{AffectNet}. For CK+ and MMI datasets,  we adopt the five-fold cross-validation scheme to evaluate the recognition performance, in order to make a fair comparison with other methods. 		
Additionally, in order to fairly compare with the state-of-the-art methods of FER, we initialized the parameters of U-Net by Xavier initializer \cite{glorot2010understanding} rather than pre-training.
In experiments, the original images are resized to 144$\times$144, and the training images are augmented by standard approaches, such as image flips and random cropping.
The number $M$ of local regions is set as 16, and each patch (local region) overlaps about 16 pixels with its adjacent patches, and the parameter $\alpha$ is set as 0.7 in Eq.(1).
The size of the epoch is set to 24, the initial learning rate is 0.0003, and the weight decay is set as 0.95 each epoch. 

In Tables. \ref{non-local-atte}, \ref{Local-atte} and \ref{SimpleNet}, we give the structures of the non-local attention network, the local attention and the simple net, respectively.
For the non-local attention network, we only show the convolution layer and the pooling layer, and the operations such as reshaping and matrix multiplication are not shown.
All experiments are implemented on the framework of Tensorflow and GTX 2080Ti with 11G memory.

\begin{figure}[t]
	\centering
	\includegraphics[width=0.7\columnwidth]{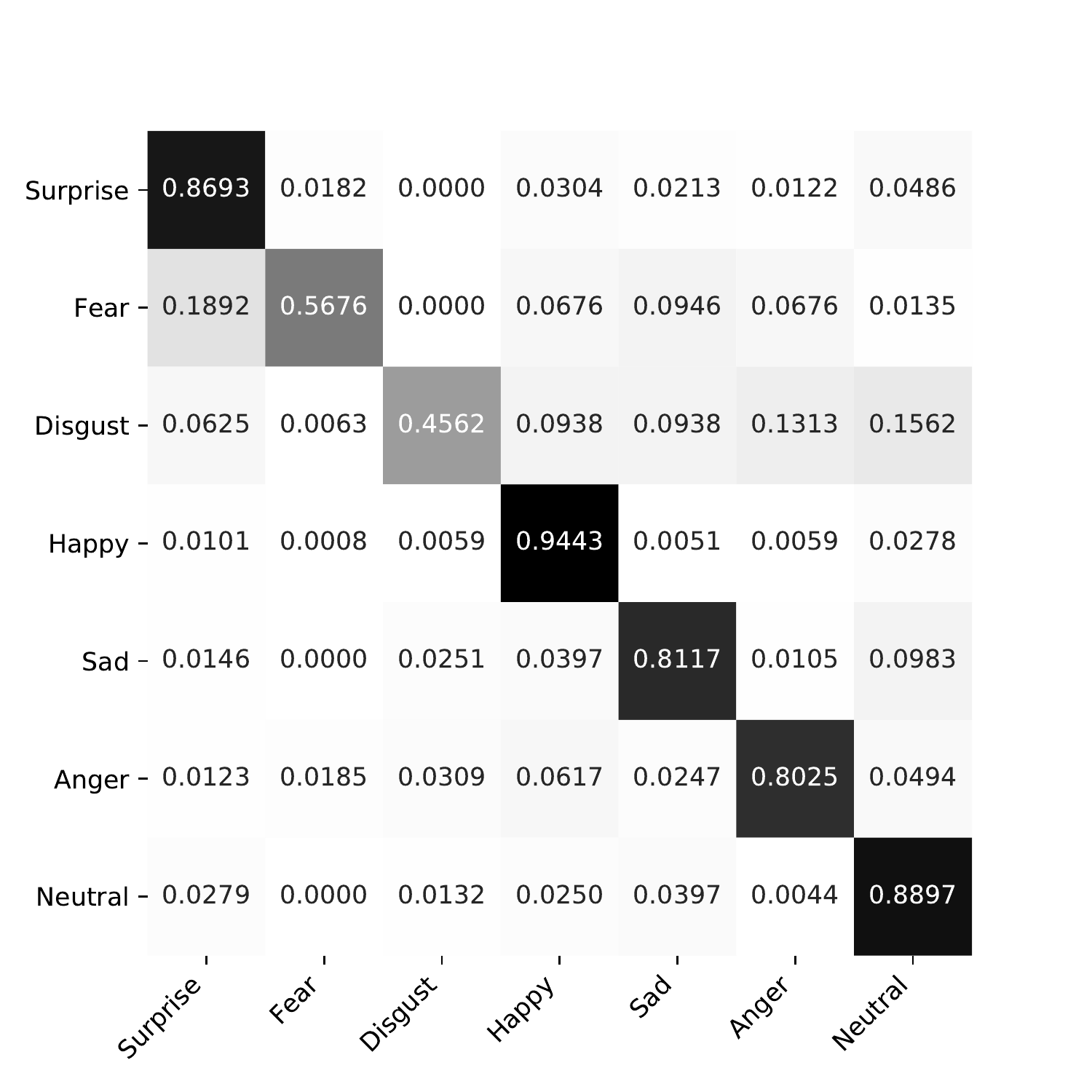}
	\caption{Confusion matrix of the proposed model (LNLAttenNet) on RAF-DB database.}\label{fig7}
\end{figure}
\begin{figure}[t]
	\centering
	\includegraphics[width=0.7\columnwidth]{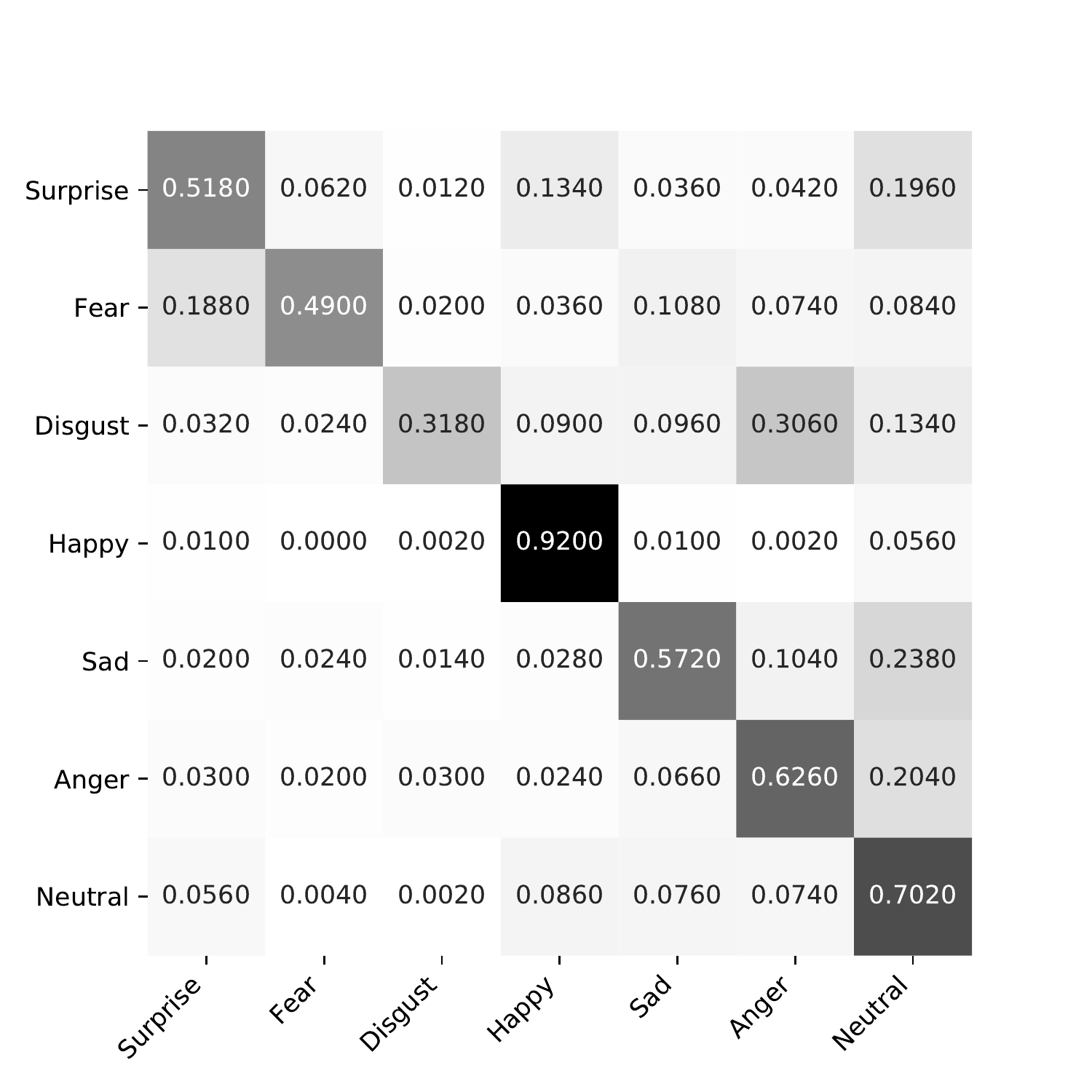}
	\caption{Confusion matrix of the proposed model (LNLAttenNet) on AffectNet database.}\label{fig8}
\end{figure}

\subsection{Comparisons with State-of-the-Art Methods}
In order to validate the performance of the proposed method, we firstly give a comparison with eight state-of-the-art methods on five datasets.
Eight compared methods are VGG16 \cite{VGG}, DLP-CNN \cite{Li_2017_CVPR}, NAL \cite{NAL}, Soft-CNN \cite{gan2019facial}, CenterLoss \cite{wen2016discriminative}, gACNN \cite{li2018occlusion}, LDL-ALSG \cite{chenlabel20} and IPA2LT \cite{IPA2LT}, where VGG16 is applied as the baseline method in experiments.
\begin{itemize}
	\item {\bf DLP-CNN} \cite{Li_2017_CVPR} decomposes the image structurally rather than spatially into regions (parts) which are discriminative for matching. According to the representations over the regions, it aggregates discriminative features for classification.
	\item {\bf NAL} \cite{NAL} utilizes a noise adaptation layer to address the problem of noise labels.
	\item {\bf Soft-CNN} \cite{gan2019facial} fuses the latent label probability distribution predicted by the trained model to obtain soft labels with a novel label-level perturbation strategy.
	\item {\bf{CenterLoss}} \cite{wen2016discriminative} minimizes the center loss calculated by the distance between each data and its corresponding class center to reduce the intra-class discrepancy.
	\item {\bf{gACNN}} \cite{li2018occlusion} uses 24 facial landmarks as the attention mechanism to conduct multi-region ensemble at the feature level.
	\item {\bf{LDL-ALSG}} \cite{chenlabel20} considers the subjectivity of human annotators and the ambiguous expression labels and then leverages the topological information of the labels from related but more distinct tasks, such as AU recognition and facial landmark detection, to explore the label distribution of facial expressions.
	\item {\bf{IPA2LT}} \cite{IPA2LT} employs an inconsistent pseudo annotations framework to solve the inconsistent annotations between different facial expression databases.
\end{itemize}

Noticeablely, IPA2LT \cite{IPA2LT} applies both RAF and AffectNet as the training set, differently from our method (LNLAttenNet) and other compared methods where only the training set of one dataset is employed to train a model.
In LNLAttenNet, both non-local attention and local attention mechanisms are utilized.
Thus, we also make a comparison with \textit{three special cases of our model}: the model without both local and non-local attention ({\bf{Model-S}}), the model only with local attention ({\bf{Model-Local}}), and the model only with non-local attention ({\bf{Model-NonLocal}}).
Table \ref{tab1} shows the experimental results of 12 models, where the highest accuracy is bold for each dataset.
All results are the average of the last 10 epochs.

\begin{table*}[t]
	\caption{Accuracy (\%) of the proposed method (LNLAttenNet) compared with state-of-the-arts methods.}\label{tab1}
	\small 
	\begin{center}
		\begin{tabular}{l|ccccc|c}
			\hline
			Methods              & AffectNet     & RAF-DB        & SFEW          & CK+            & MMI  & average \\
			\hline\hline
			VGG16{\cite{VGG}}              & 51.11   & 80.96   & 54.45         & 90.37   & 63.21 & 68.02\\
			DLP-CNN\cite{Li_2017_CVPR}     & 54.47   & 80.89   & -             & -       & - & -\\
			NAL\cite{NAL}                  & 55.97   & 84.22   & {\bf{58.13}}  & 91.20   & 64.71 & 70.85\\
			Soft-CNN\cite{gan2019facial}   & 56.77   & 85.20   & 55.73   & -   &-     & -\\
			CenterLoss\cite{wen2016discriminative}   & 57.37   & 84.42   & 56.19   & 95.48   & - & -\\
			gACNN\cite{li2018occlusion}        & 58.78         & 85.07         & -             & 97.03           & - &-\\
			LDL-ALSG\cite{chenlabel20}    & {\bf{59.35}}   & 85.53   & 56.50   & 93.08   & {\bf{70.49}} & 72.99\\
			\hline
			Model-S               & 56.26   & 83.80   & 54.82   & 94.14   & 63.52   & 70.51 \\
			Model-Local           & 57.63   & 84.55   & 56.42   & 96.44   & 65.42   & 72.09 \\
			Model-NonLocal        & 58.09   & 85.04   & 55.73   & 96.63   & 66.56   & 72.41 \\
			LNLAttenNet           & 59.28   & {\bf{86.15}}   & 57.80   & {\bf{98.18}}   & 68.75   & {\bf{74.03}}\\
			\hline
			IPA2LT\cite{IPA2LT}   & 55.11   & \bf{\emph{{86.77}}}   & \bf{\emph{58.29}}   & 91.67   & 65.61   & 71.49\\
			\hline
		\end{tabular}
	\end{center}	
\end{table*}

\begin{figure*}[htb]
	\centering
	\includegraphics[width=0.11\textwidth]{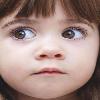}
	\includegraphics[width=0.11\textwidth]{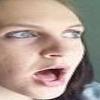}
	\includegraphics[width=0.11\textwidth]{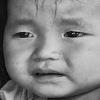}
	\includegraphics[width=0.11\textwidth]{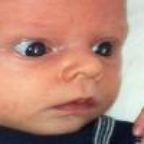}	
	\includegraphics[width=0.11\textwidth]{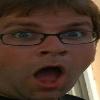} 		
	\includegraphics[width=0.11\textwidth]{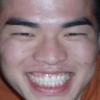}
	\includegraphics[width=0.11\textwidth]{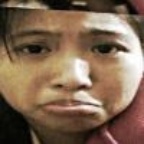}
	\includegraphics[width=0.11\textwidth]{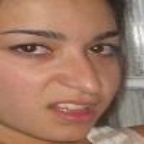}\\	
	\includegraphics[width=0.11\textwidth]{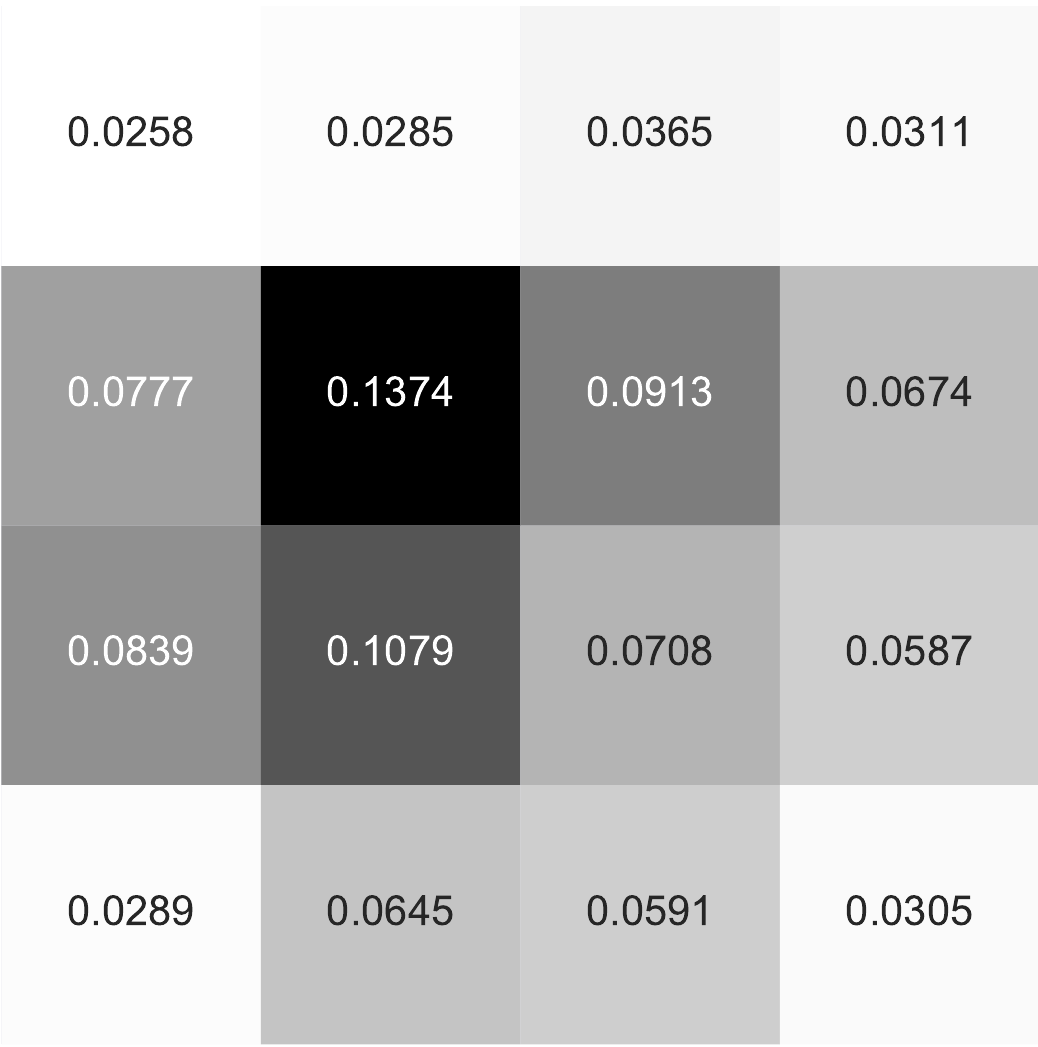}
	\includegraphics[width=0.11\textwidth]{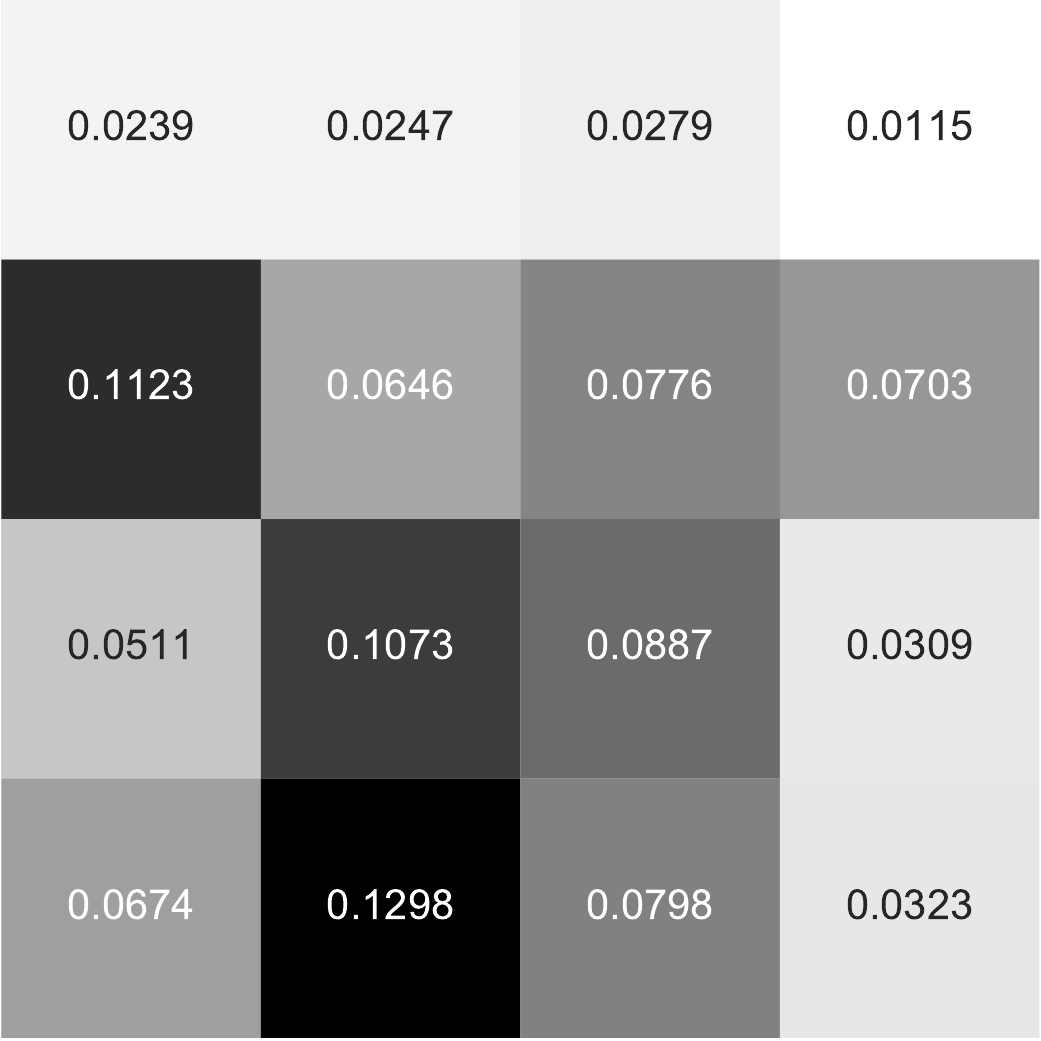}
	\includegraphics[width=0.11\textwidth]{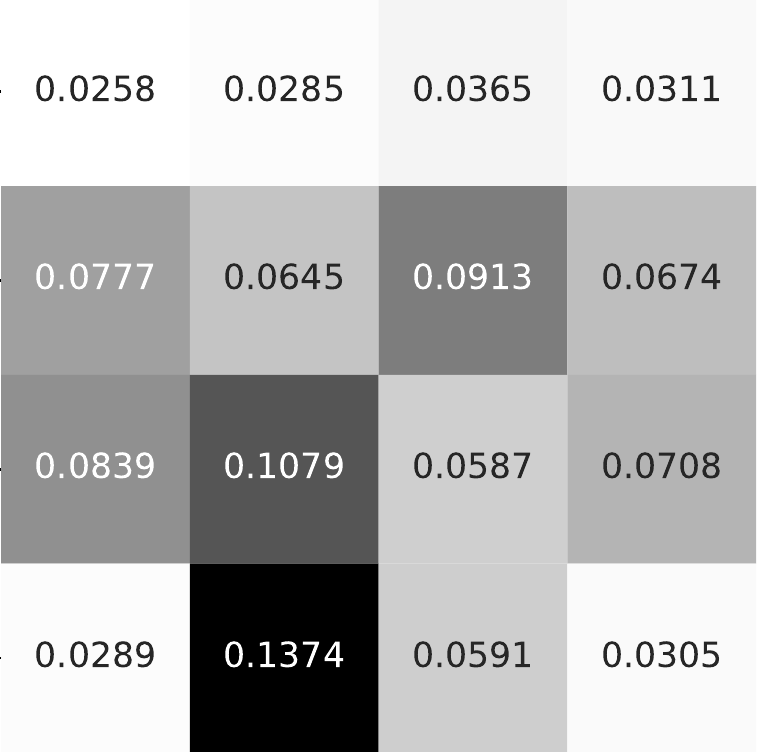}
	\includegraphics[width=0.11\textwidth]{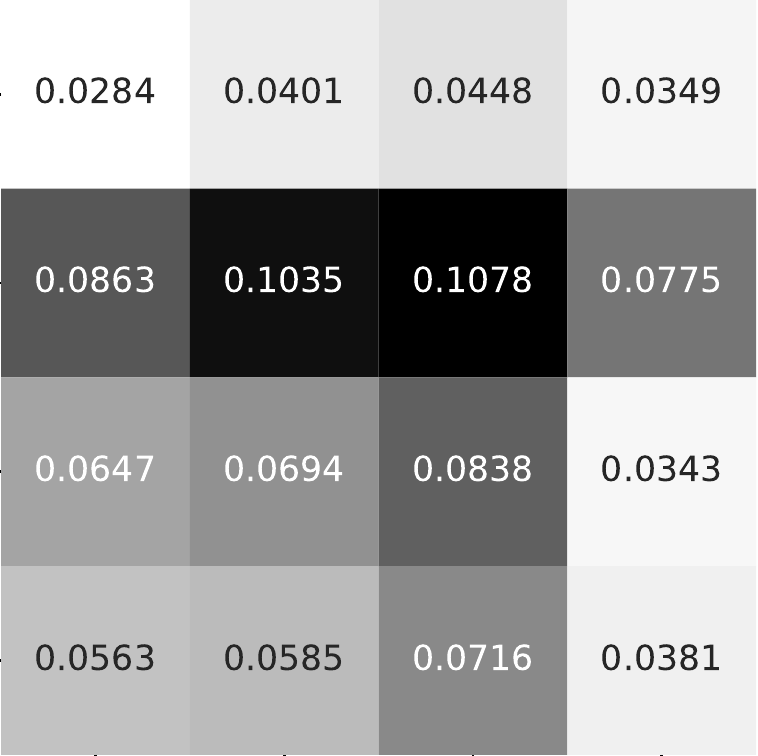}	
	\includegraphics[width=0.11\textwidth]{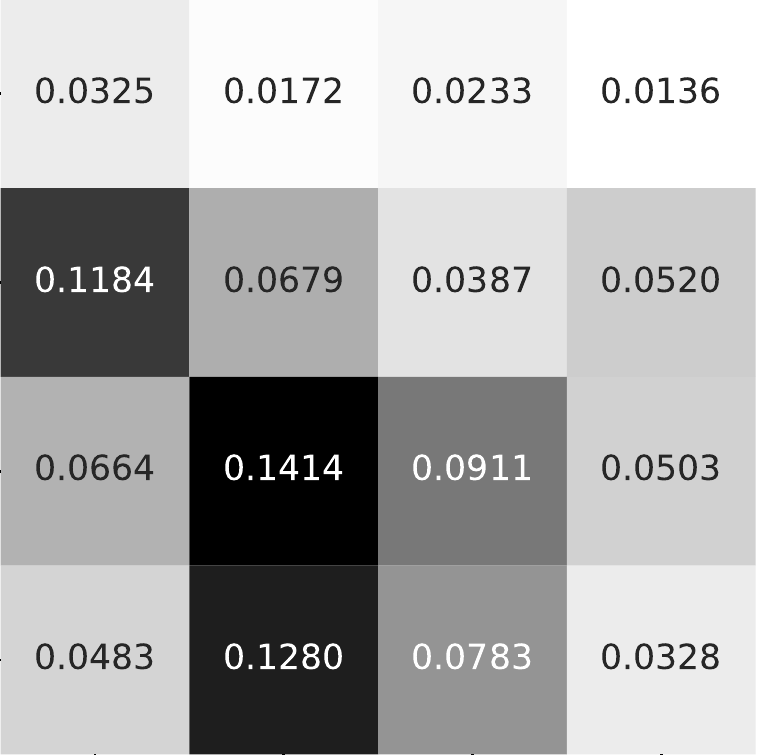} 	
	\includegraphics[width=0.11\textwidth]{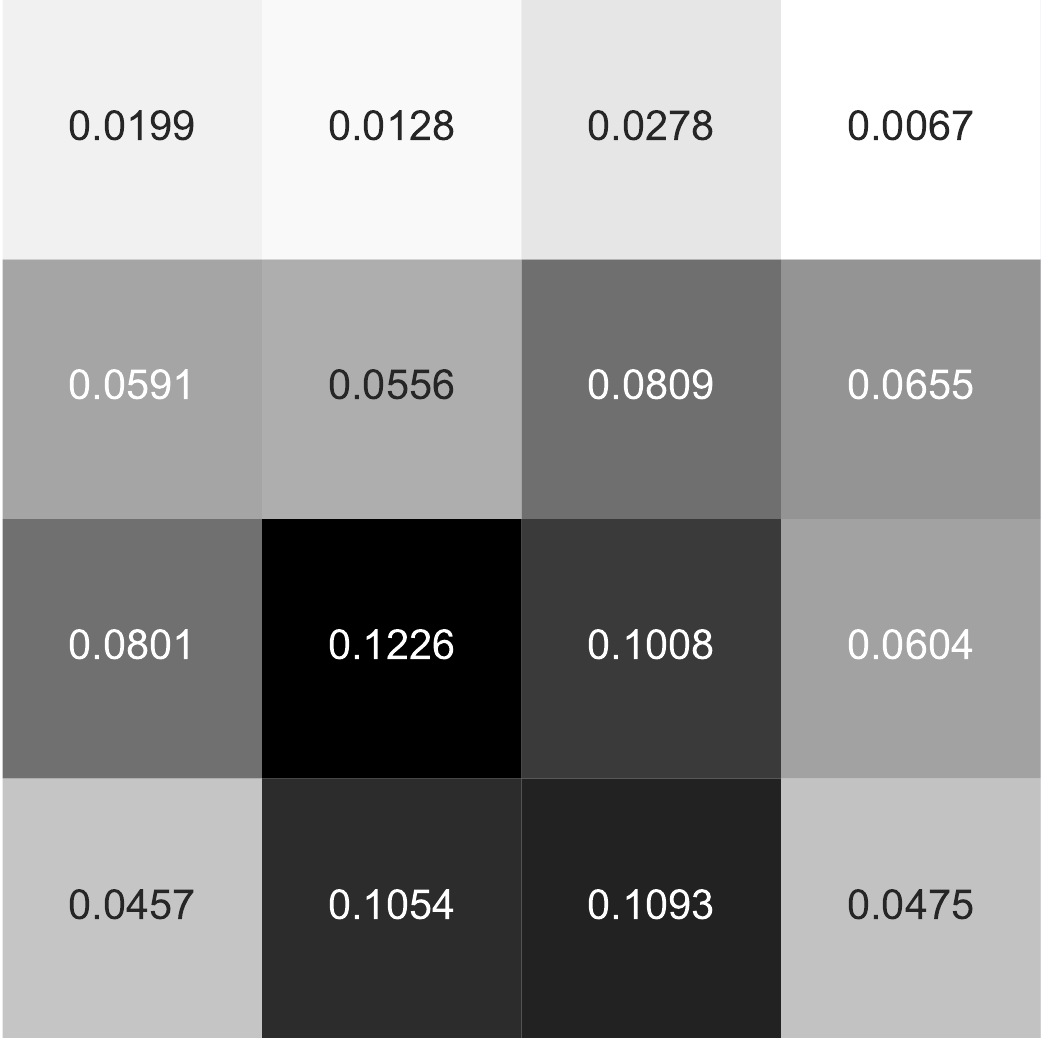}
	\includegraphics[width=0.11\textwidth]{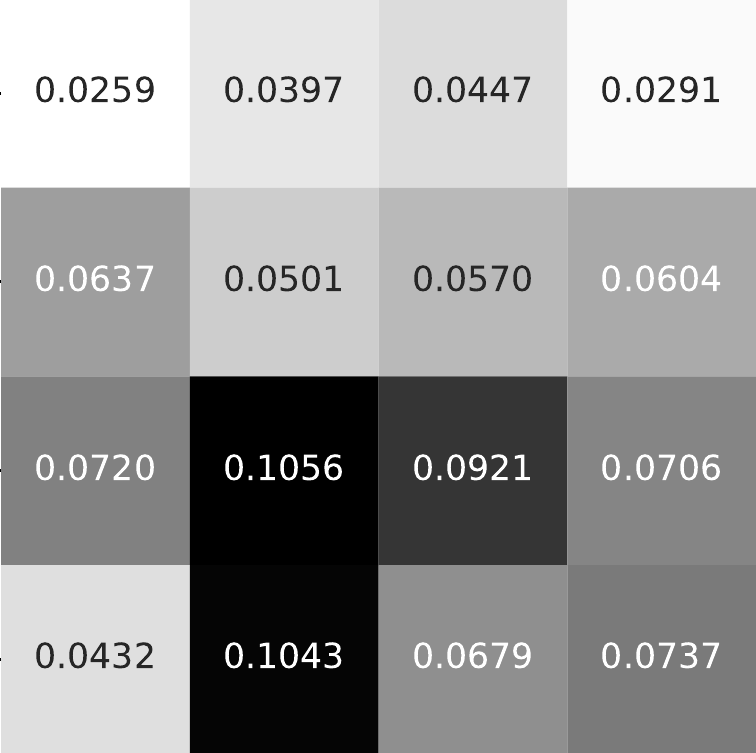}	
	\includegraphics[width=0.11\textwidth]{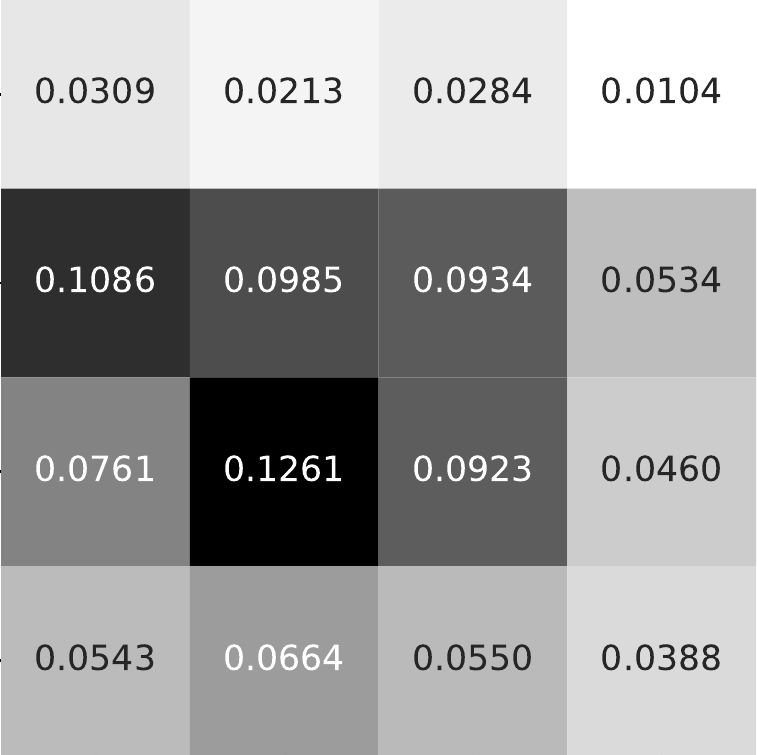}\\
	\includegraphics[width=0.11\textwidth]{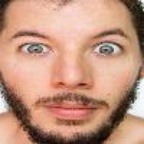}
	\includegraphics[width=0.11\textwidth]{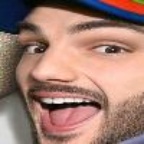}
	\includegraphics[width=0.11\textwidth]{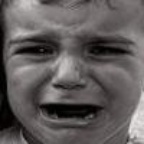}	
	\includegraphics[width=0.11\textwidth]{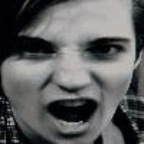}
	\includegraphics[width=0.11\textwidth]{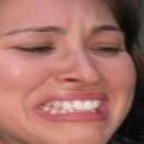}
	\includegraphics[width=0.11\textwidth]{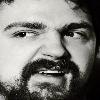}
	\includegraphics[width=0.11\textwidth]{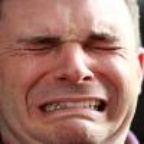}
	\includegraphics[width=0.11\textwidth]{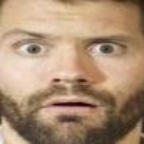}\\
	\includegraphics[width=0.11\textwidth]{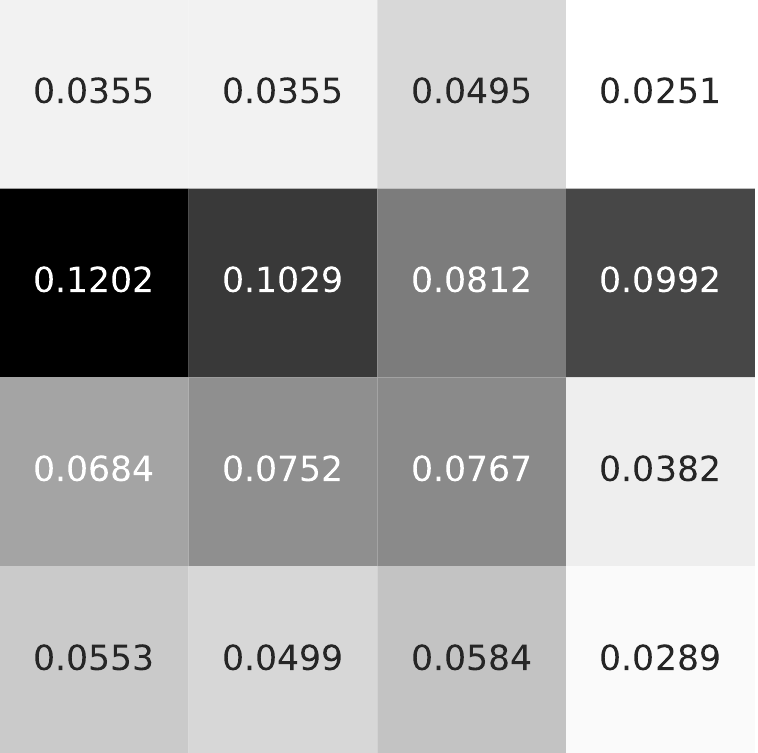}
	\includegraphics[width=0.11\textwidth]{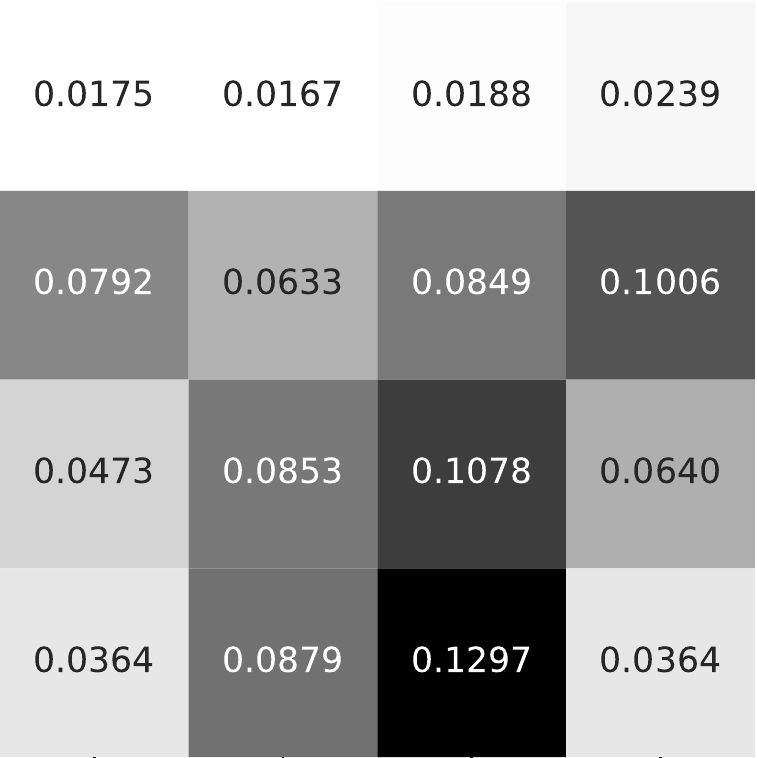}
	\includegraphics[width=0.11\textwidth]{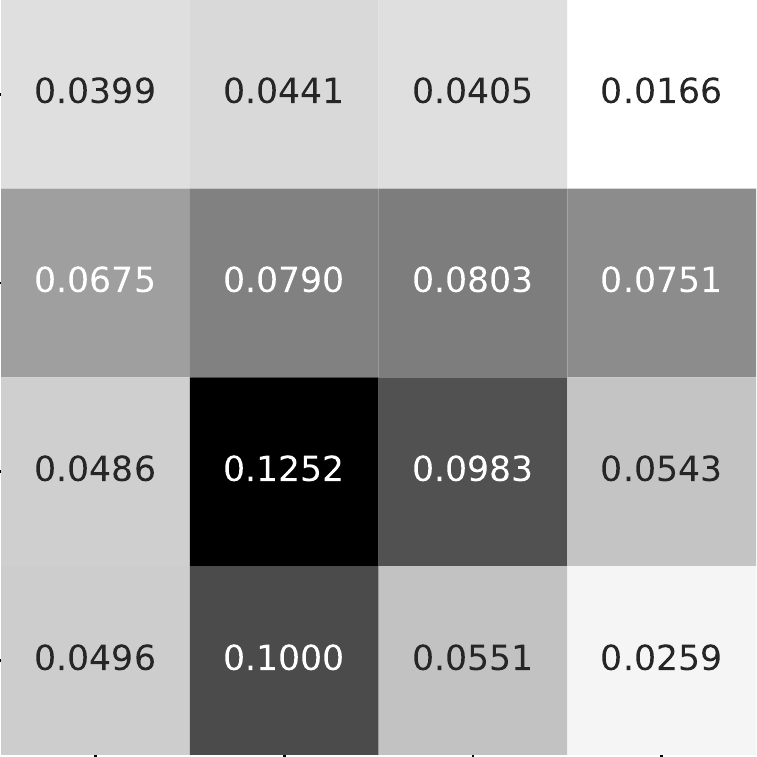}	
	\includegraphics[width=0.11\textwidth]{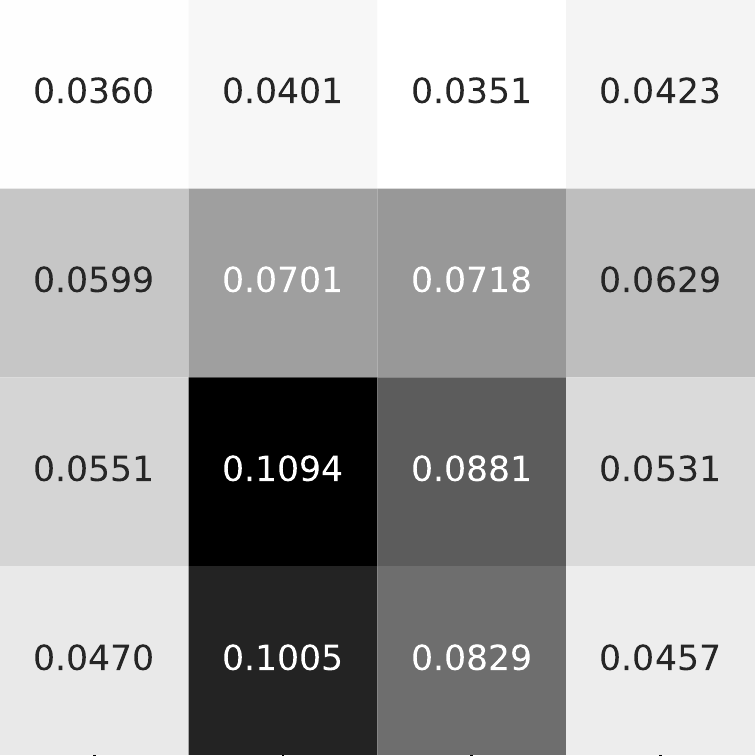}	
	\includegraphics[width=0.11\textwidth]{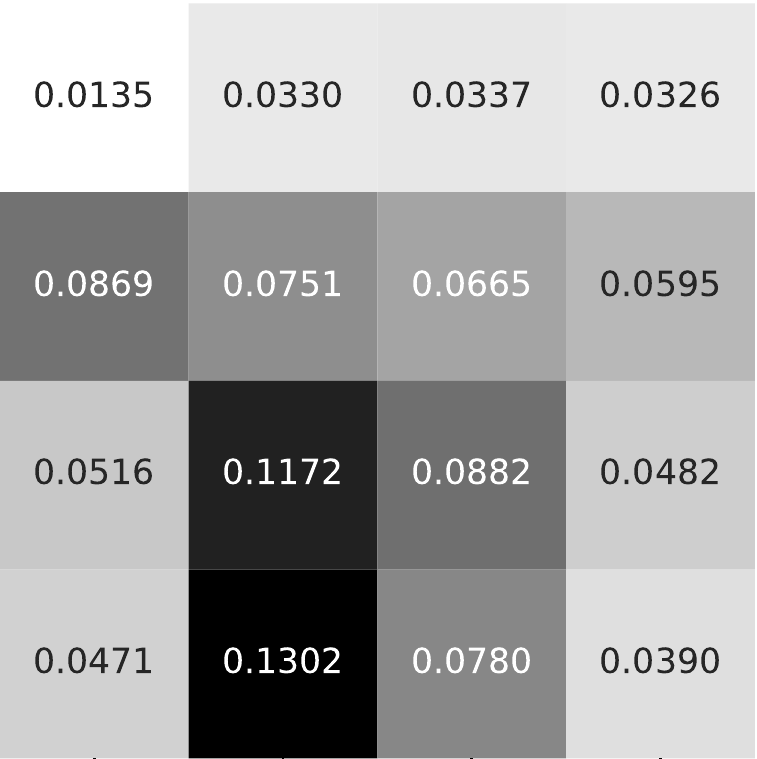}
	\includegraphics[width=0.11\textwidth]{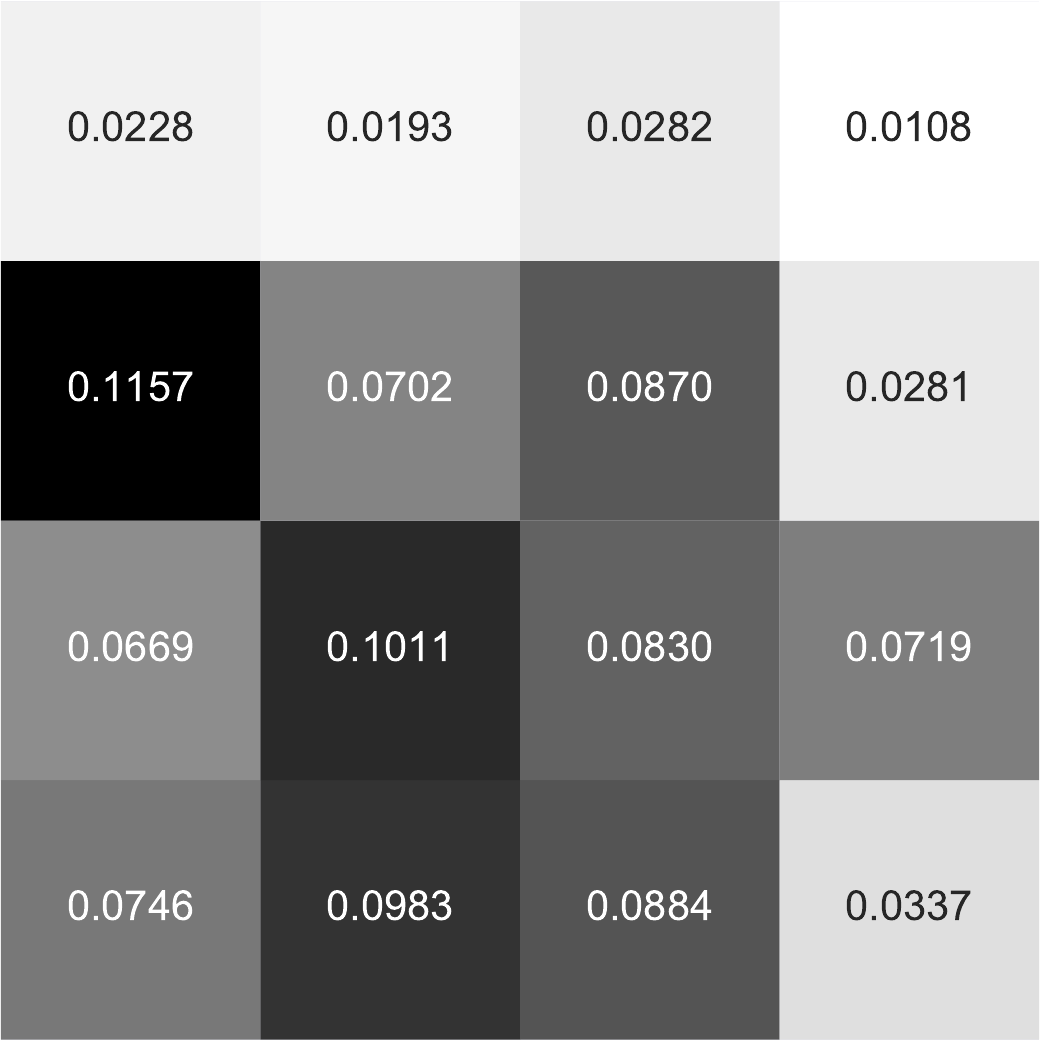}
	\includegraphics[width=0.11\textwidth]{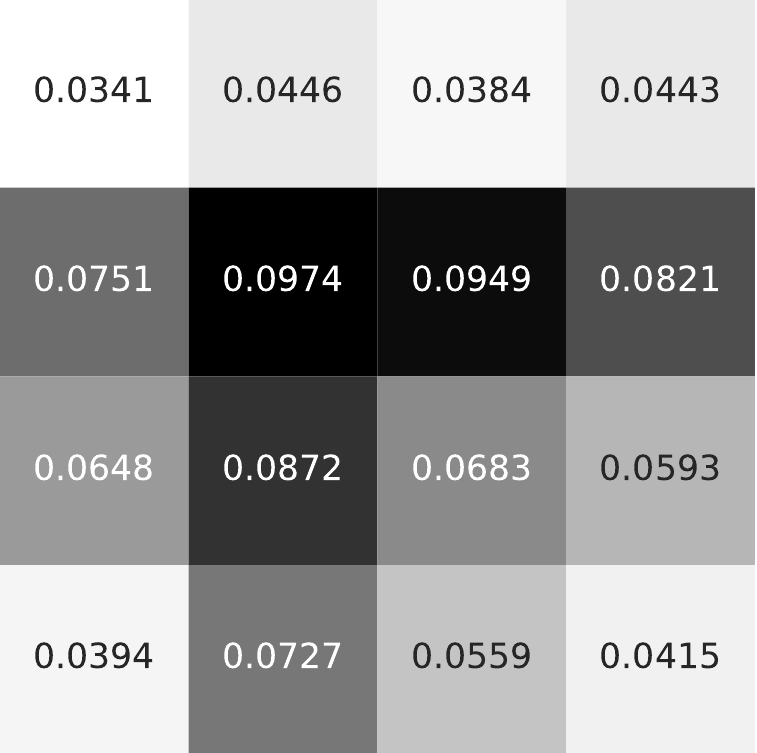}
	\includegraphics[width=0.11\textwidth]{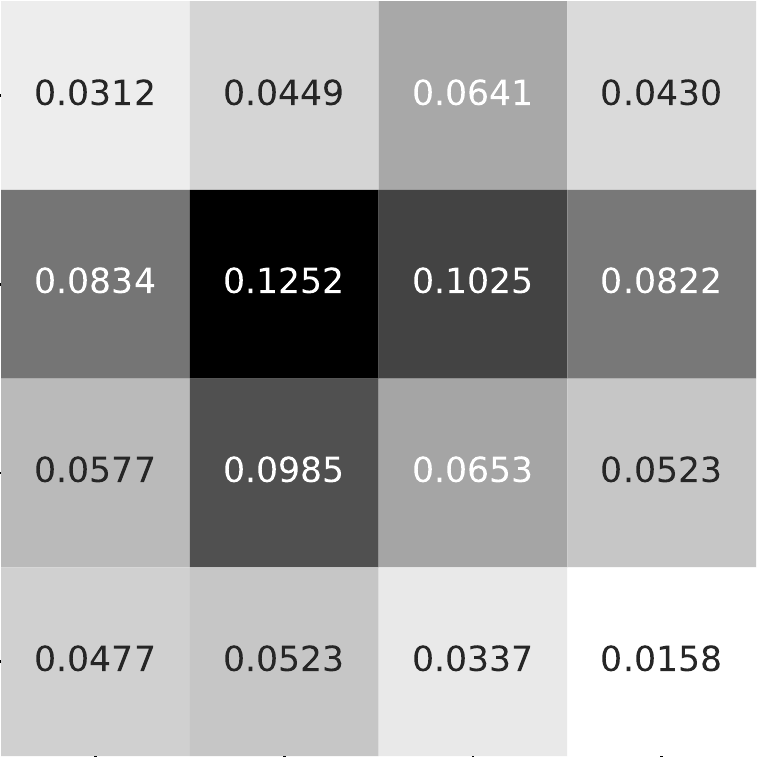}
	\caption{Non-Local weights of 16 local regions of one face in RAF-DB obtained by the proposed model. The first and third lines show the facial images, and the second and forth lines show the Non-Local weights of 16 local regions corresponding to images.}\label{FigNonLocalWeights}
\end{figure*}

From Table \ref{tab1}, it is obviously seen that the performance of the proposed method (LNLAttenNet) is superior to all compared methods except LDL-ALSG and IPA2LT on AffectNet, RAF-DB, CK+, MMI and SFEW.
Differently to LNLAttenNet, IPA2LT\cite{IPA2LT} utilizes two big datasets (RAF and AffectNet) as the training set, which results in its obtaining better performance. But, LNLAttenNet still achieves a competitive performance on two datasets (RAF-DB and SFEW) and outperforms on three datasets (AffectNet, CK+ and MMI) compared with IPA2LT.
Compared with LDL-ALSG\cite{chenlabel20}, LNLAttenNet outperforms on RAF-DB, SFEW and CK+, ties on AffectNet and loss on MMI.
In the last column of Table \ref{tab1}, we also show the average of accuracies for five datasets given by each method in the last. It is found that LNLAttenNet obtains the highest average of accuracies: 74.03\%, which illustrates LNLAttenNet can obtain a more competitive performance of FER on all of five datasets than eight compared methods.

Furthermore, it is found that Model-S is inferior to all of Model-Local, Model-NonLocal and LNLAttenNet, which demonstrates that the attention mechanism is meaningful for improving the performance of FER in our model.
Meanwhile, Model-NonLocal is slightly better than Model-Local but obviously inferior to LNLAttenNet, which also demonstrates our model jointly utilizing local and non-local information of facial expression is more effective.
In short, the experimental results illustrate that adaptively enhancing the facial crucial regions in feature learning by LNLAttenNet is effective for improving the performance of FER.

Considering that RAF and AffectNet datasets have a large amount of images, we also shows the confusion matrices for them in Fig.\ref{fig7} and Fig.\ref{fig8}, respectively.
According to the confusion matrices, it is observed that the categories (fear and surprise) are easily distinguishable for RAF-DB (shown in Fig.\ref{fig7}) and the categories (disgust and anger) are easily distinguishable for AffectNet (shown in Fig.\ref{fig8}).

\begin{figure}[t]
	\subfigure[]{
		\begin{minipage}[t]{0.98\linewidth}	
	\centering
	\includegraphics[width=0.3\columnwidth]{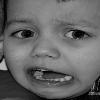}
	\includegraphics[width=0.3\columnwidth]{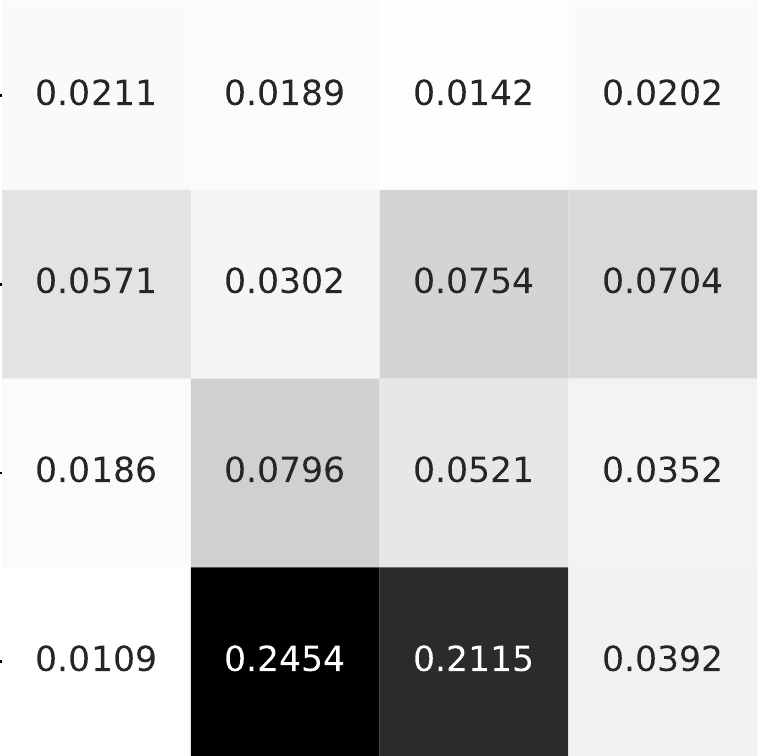}\\
	\vspace{.1in}
	\includegraphics[width=0.32\columnwidth]{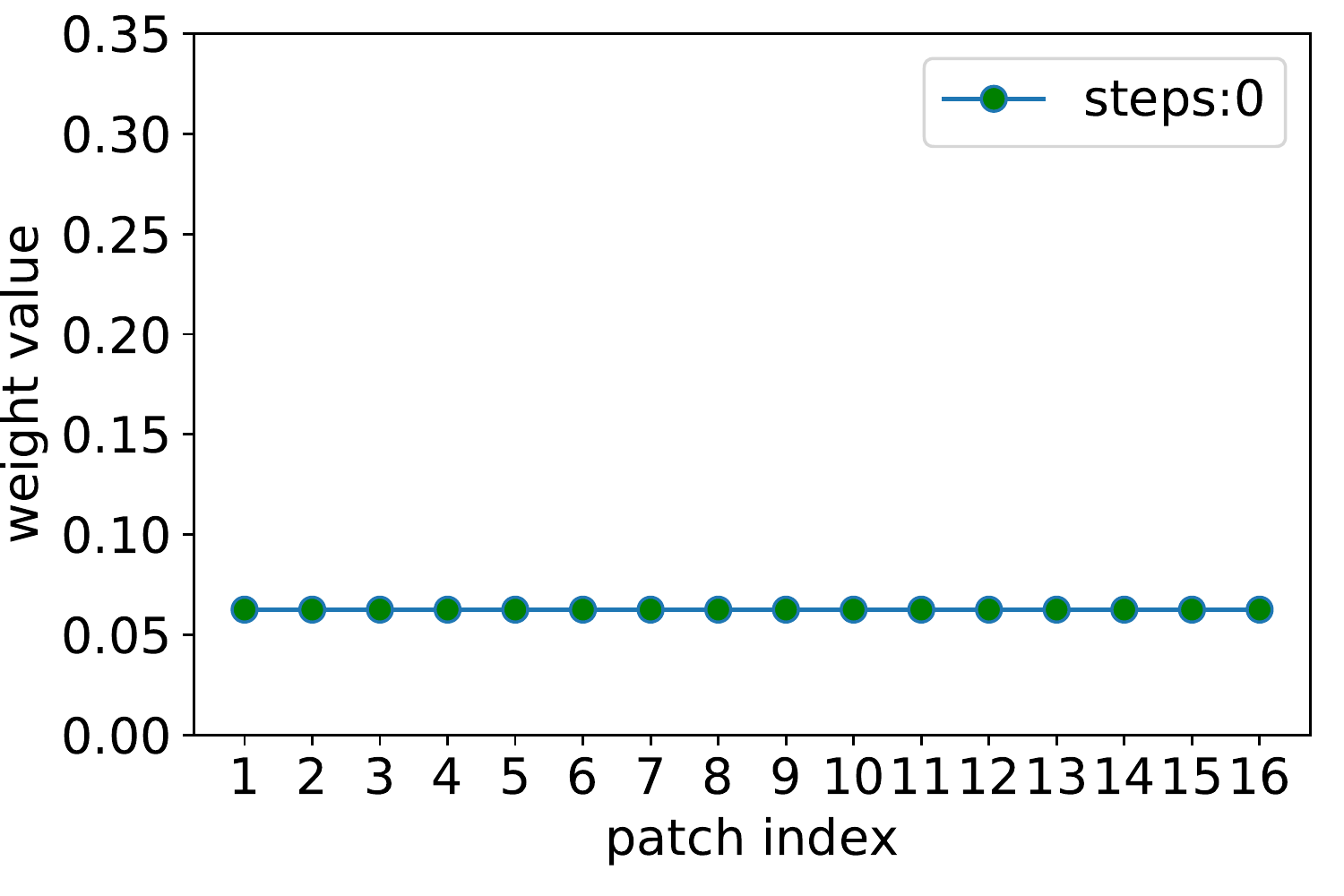}
	\includegraphics[width=0.32\columnwidth]{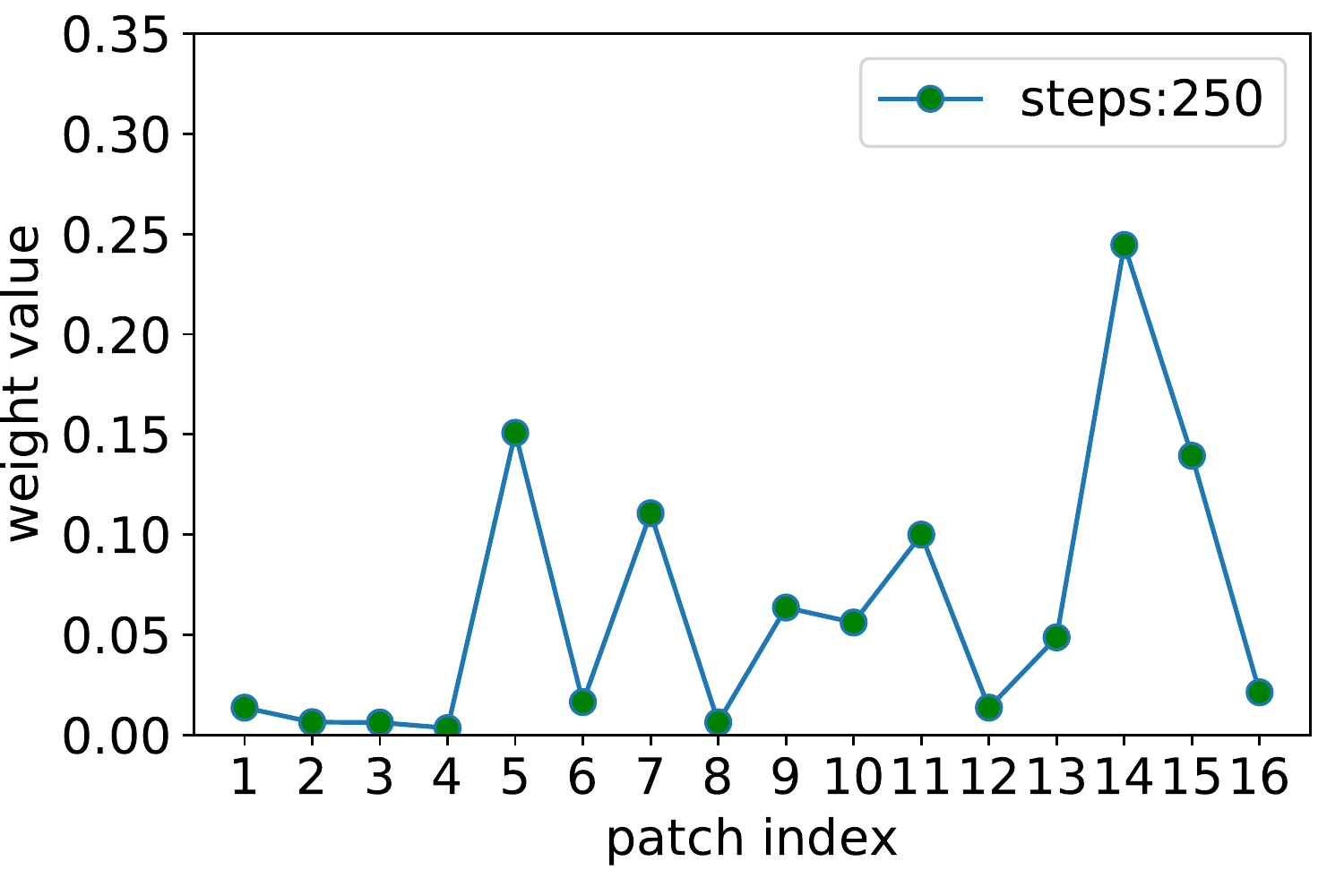}
	\includegraphics[width=0.32\columnwidth]{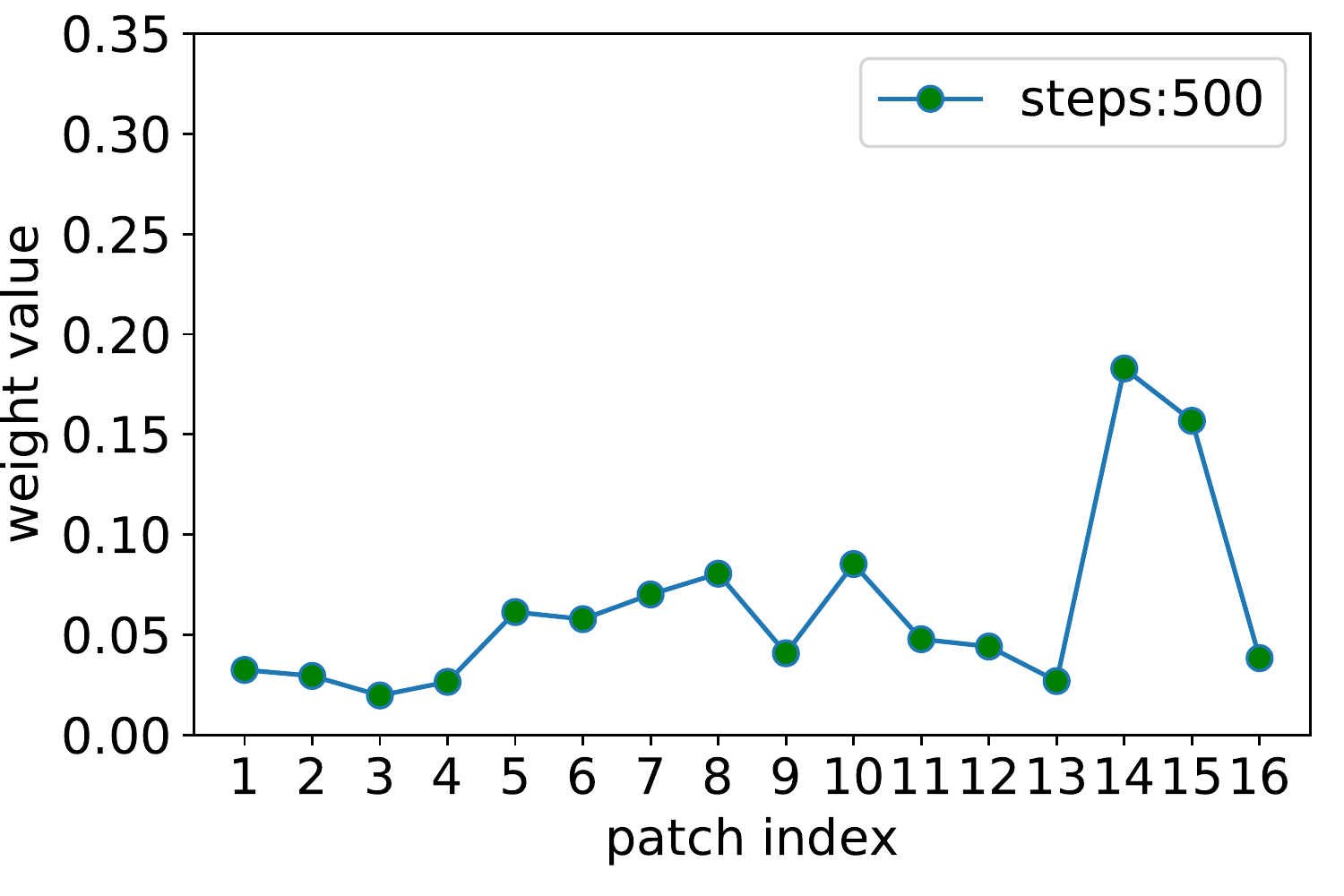}\\
	\vspace{.05in}
	\includegraphics[width=0.32\columnwidth]{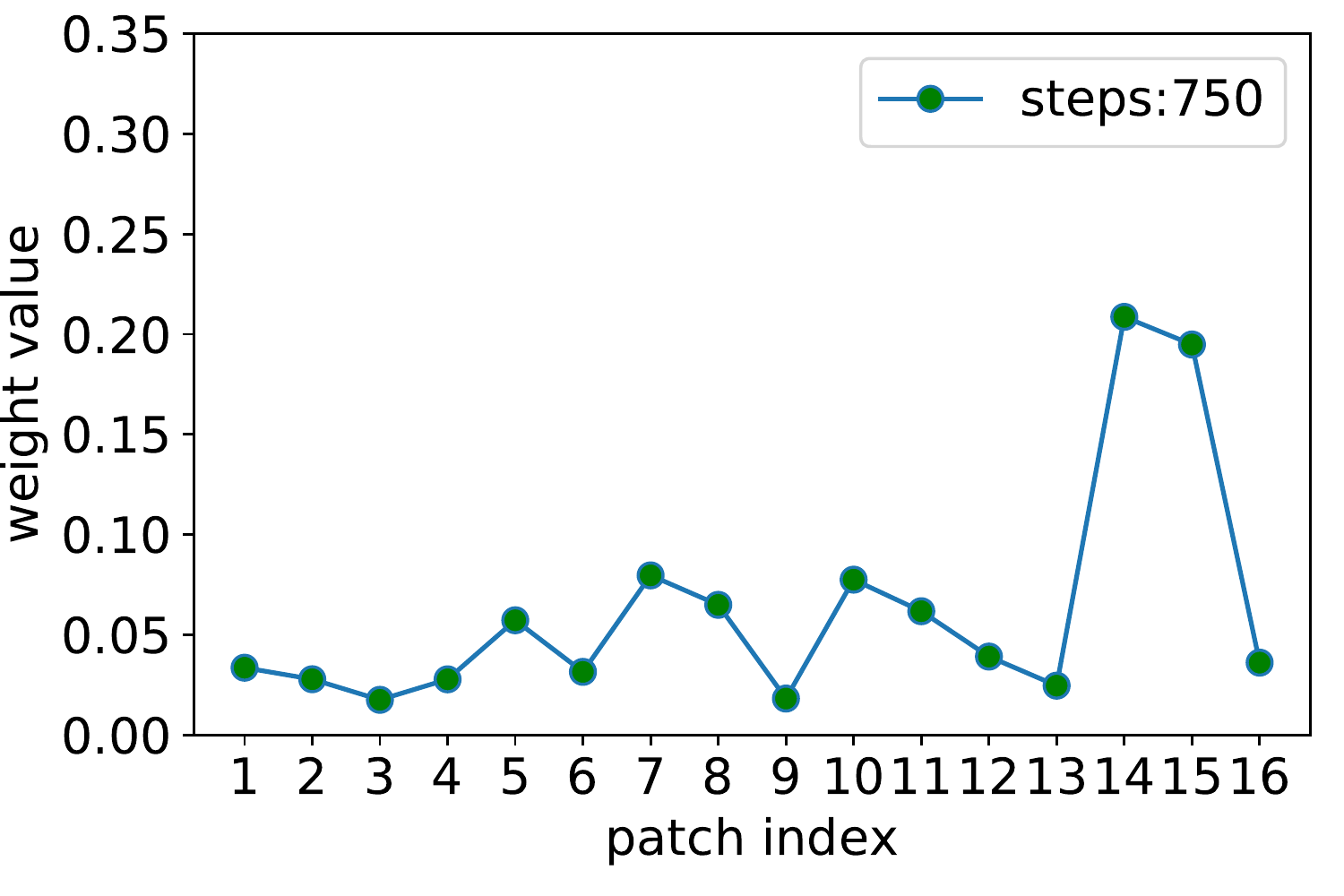}
	\includegraphics[width=0.32\columnwidth]{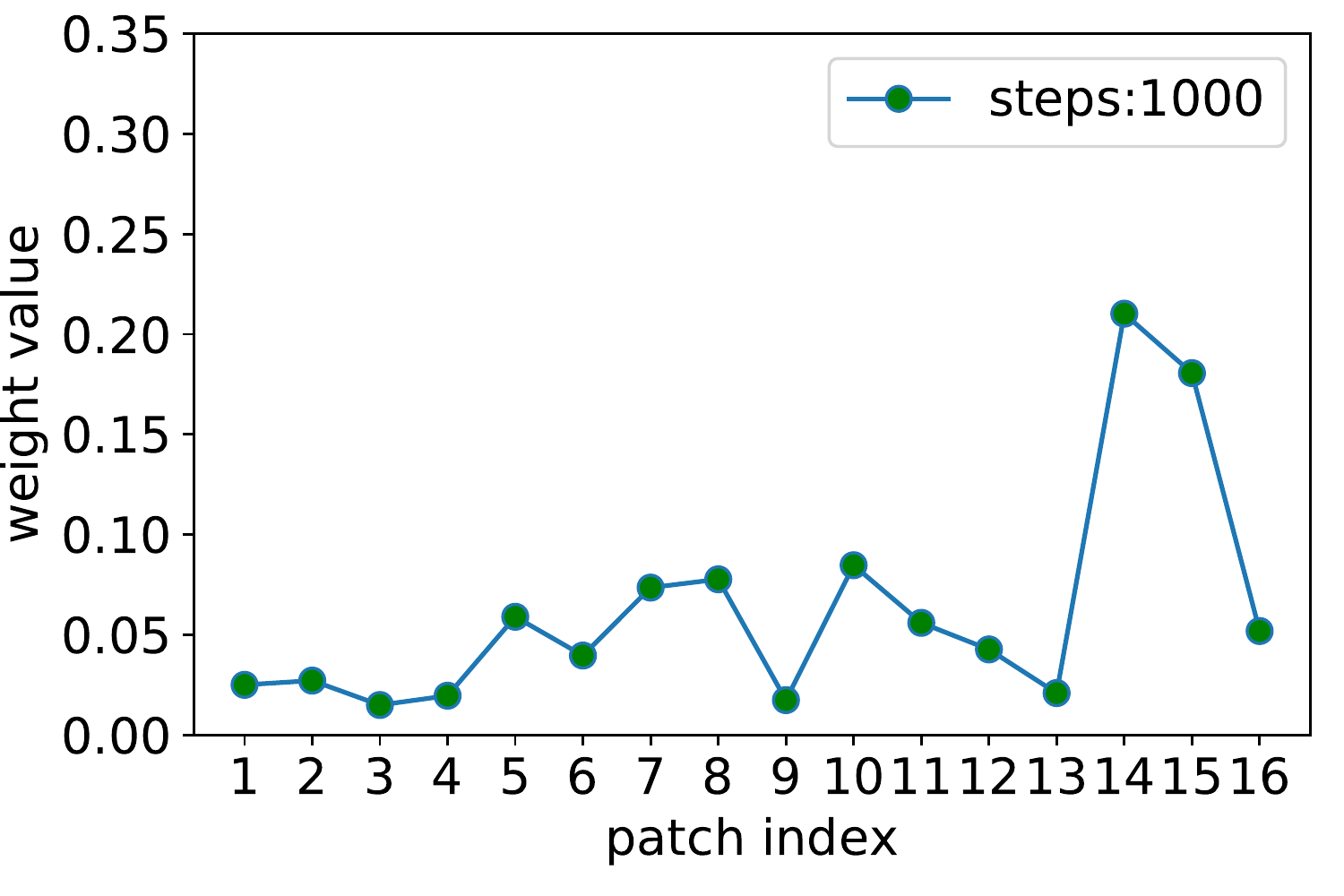}
	\includegraphics[width=0.32\columnwidth]{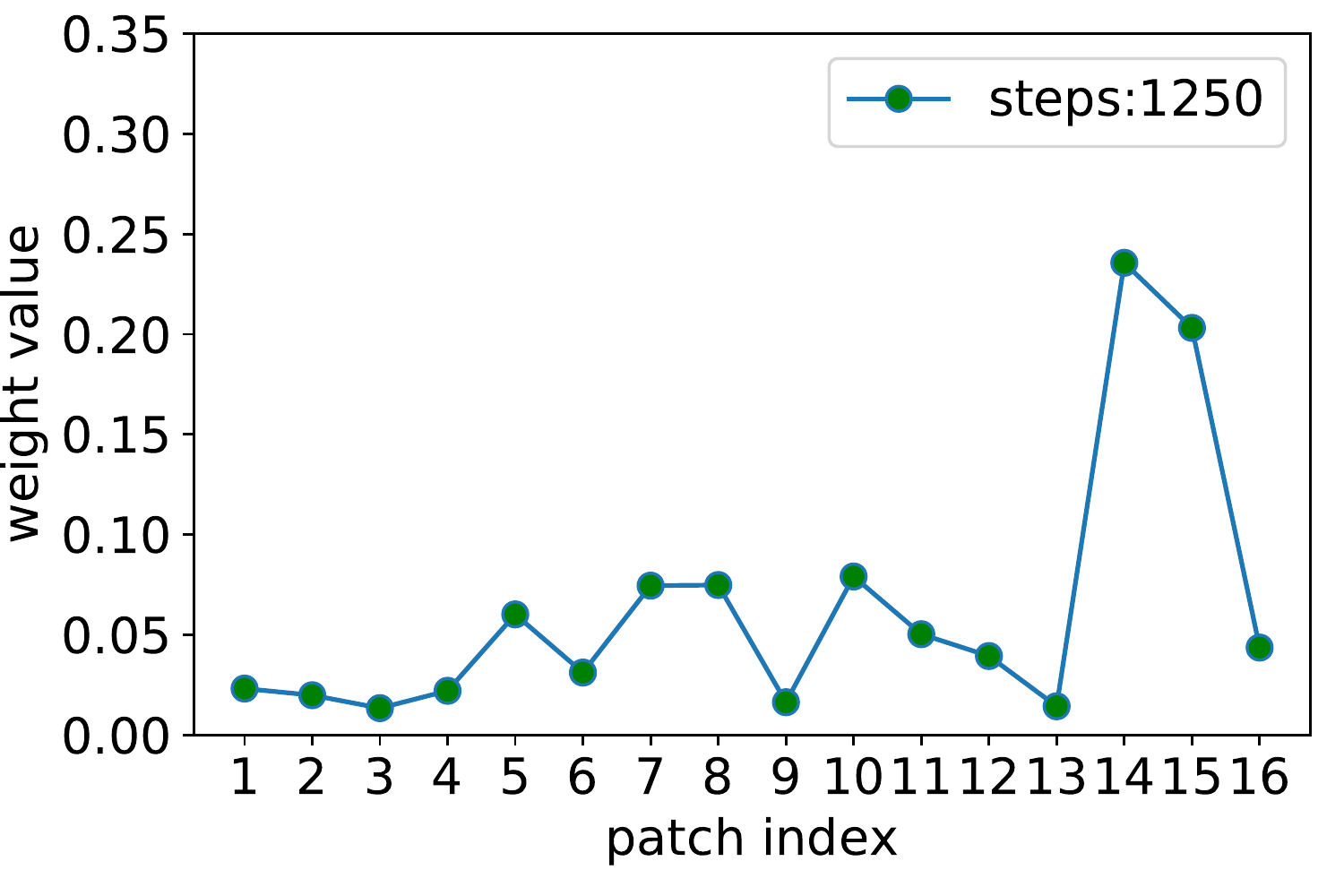}\\
	\vspace{.08in}
	\includegraphics[width=1.0\columnwidth]{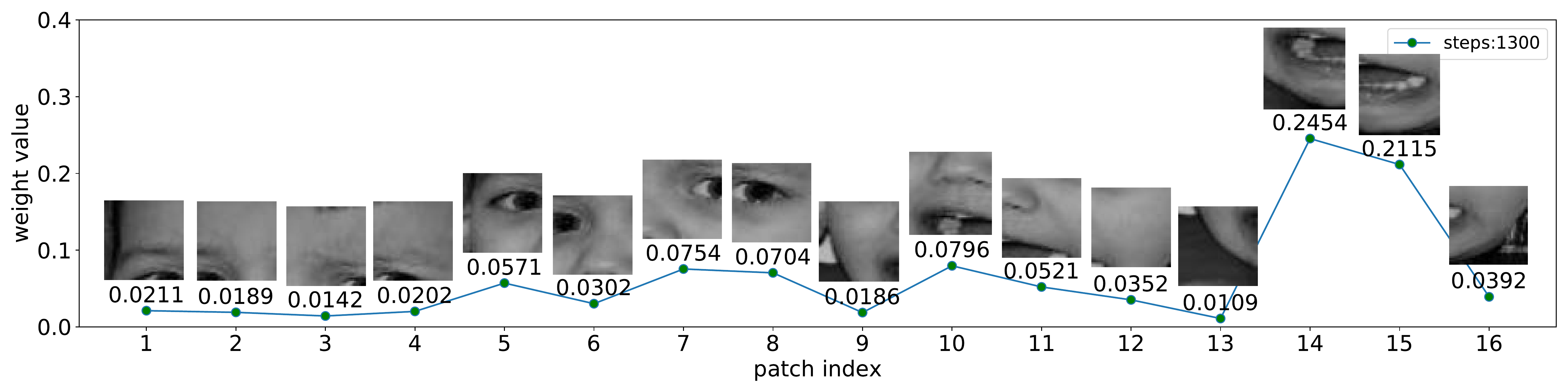}
\end{minipage}
}
\subfigure[]{
	\begin{minipage}[t]{0.98\linewidth}
		\centering
		\includegraphics[width=0.3\columnwidth]{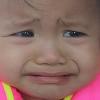}
		\includegraphics[width=0.3\columnwidth]{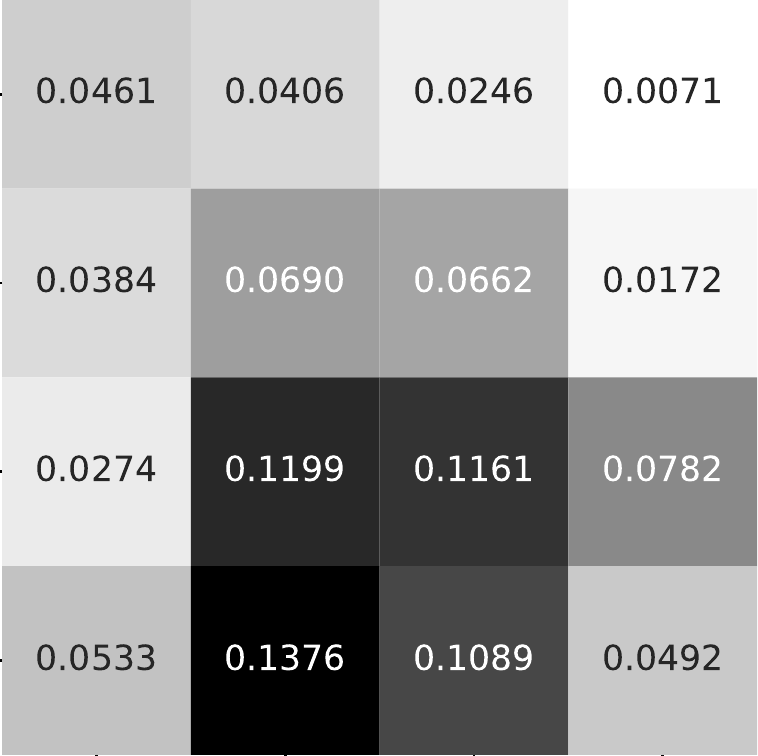}\\
		\vspace{.1in}
		\includegraphics[width=0.32\columnwidth]{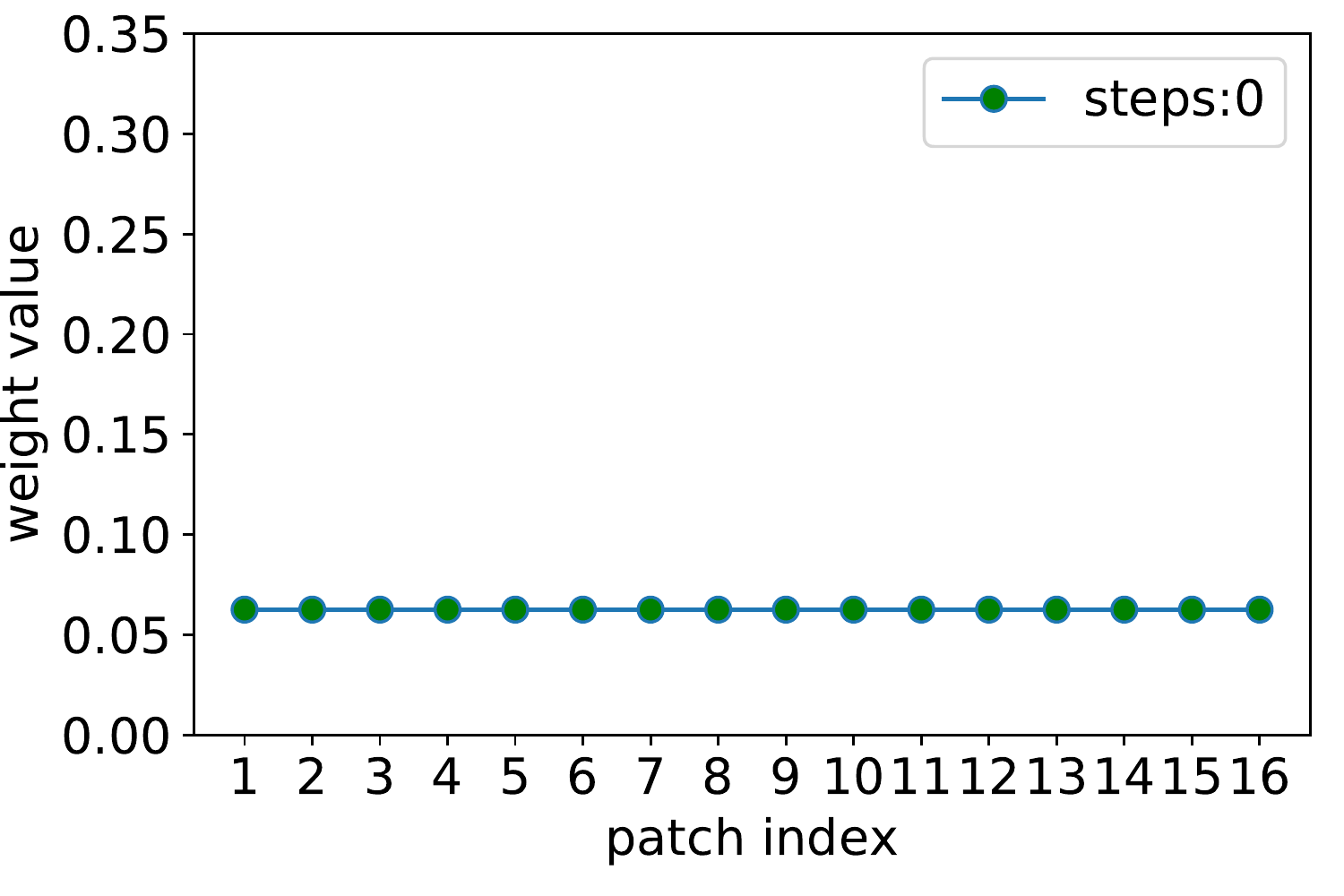}
		\includegraphics[width=0.32\columnwidth]{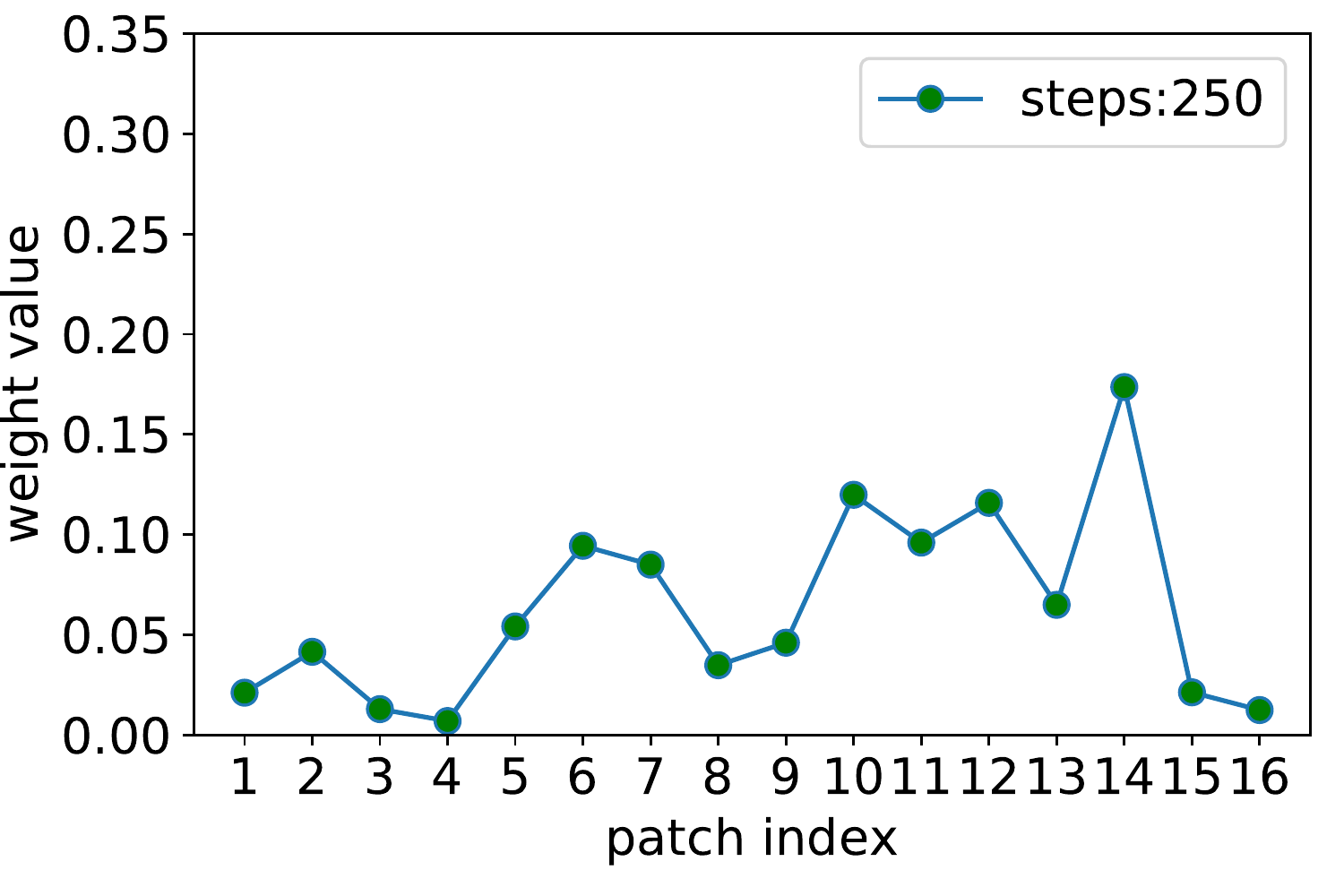}
		\includegraphics[width=0.32\columnwidth]{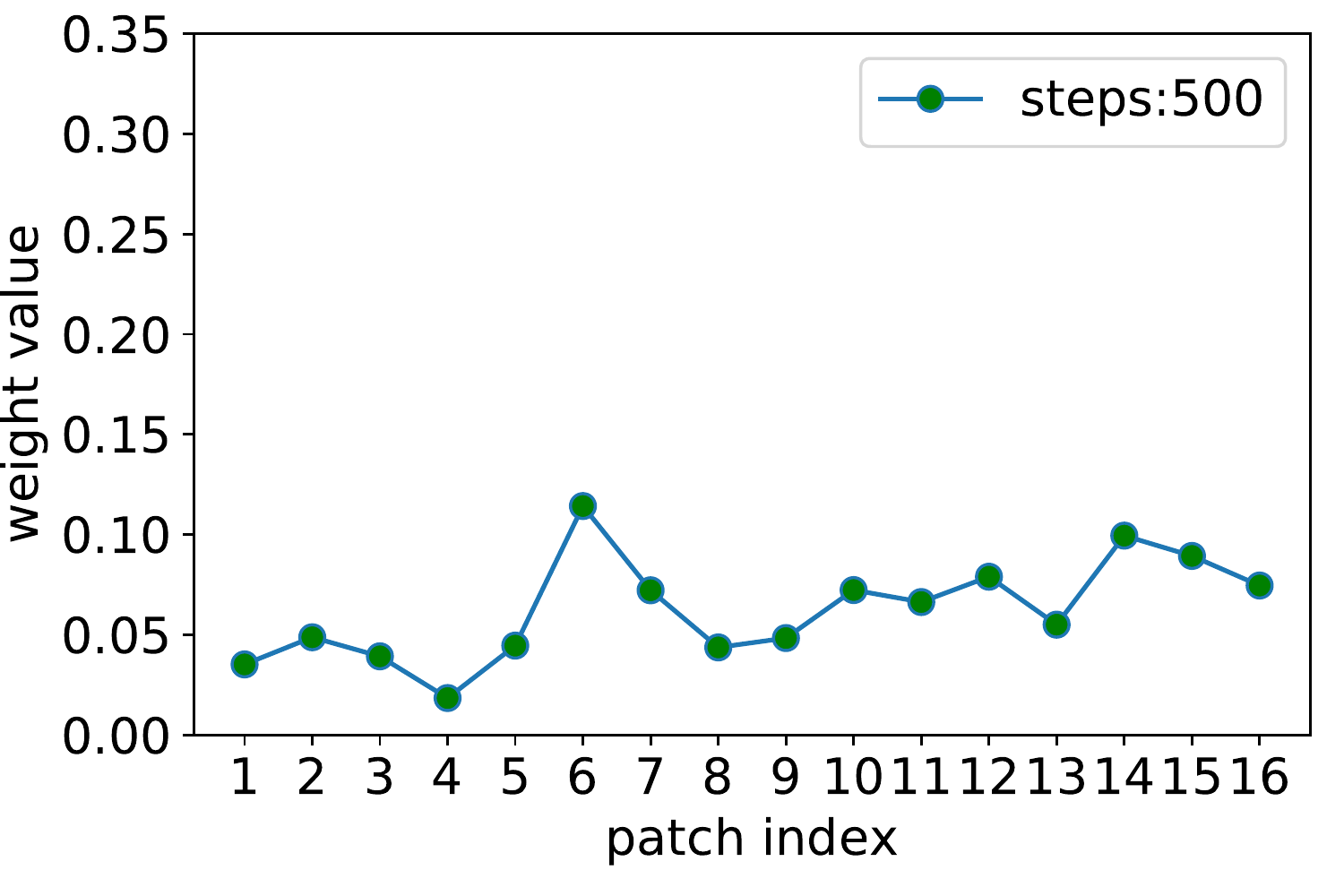}\\
		\vspace{.05in}
		\includegraphics[width=0.32\columnwidth]{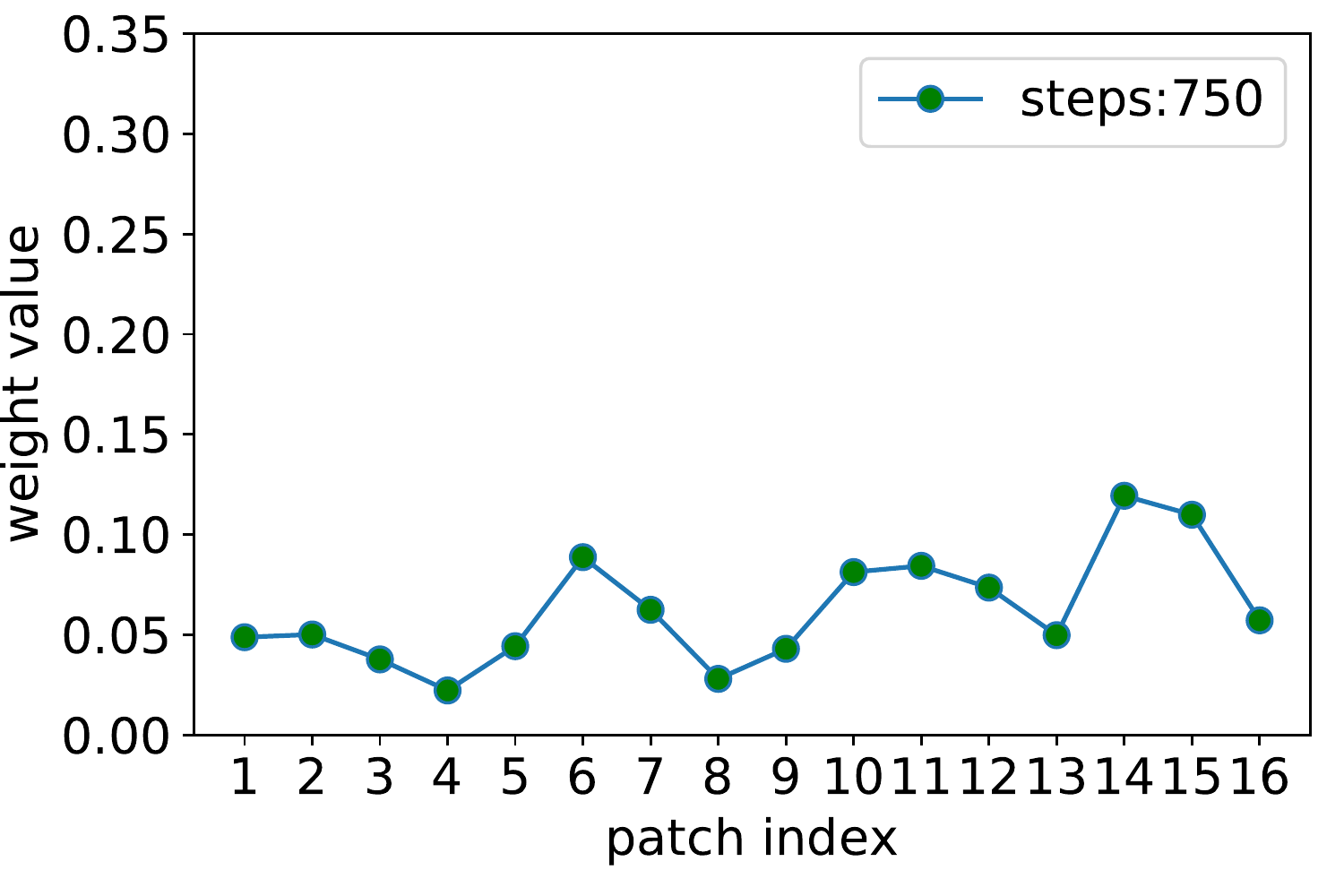}
		\includegraphics[width=0.32\columnwidth]{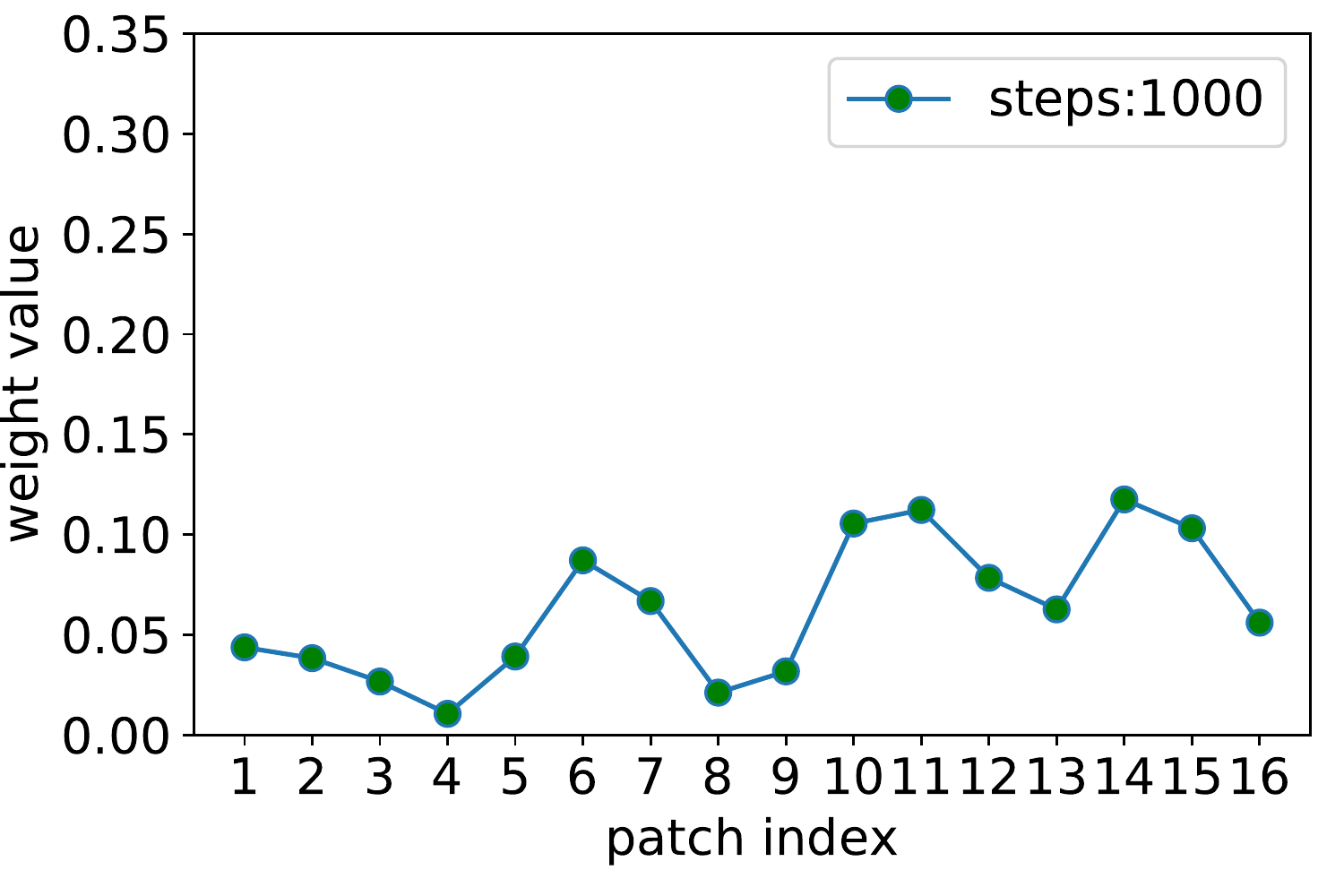}
		\includegraphics[width=0.32\columnwidth]{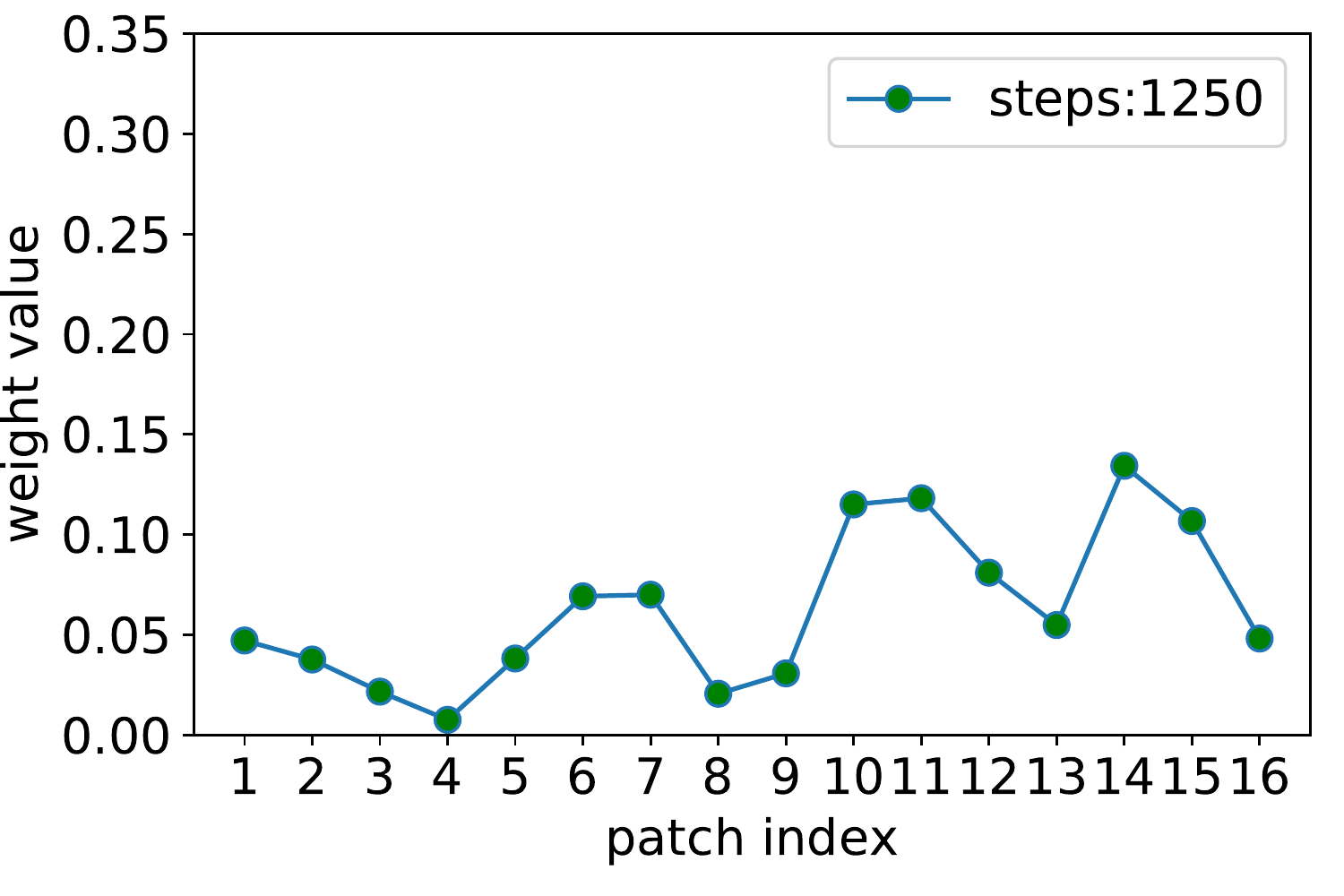}\\
		\vspace{.08in}
		\includegraphics[width=1.0\columnwidth]{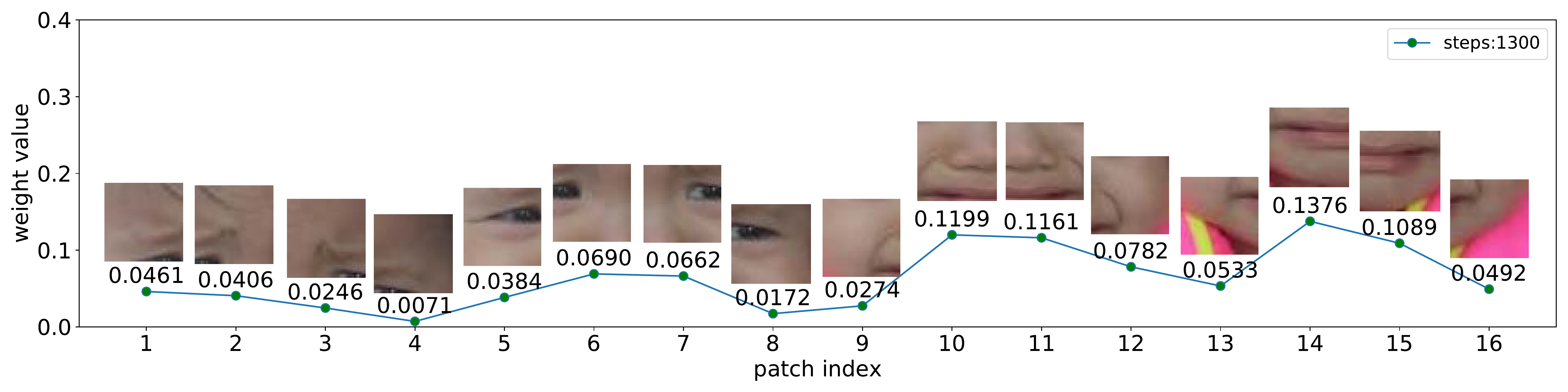}
\end{minipage}	
}	
	\caption{16 non-local weights of two input images. In the first row, the input image and the non-local weights corresponding to each patches is shown. In the second and third rows, the six figures  show the non-local weights of the input images at different training stages,respectively. The last row shows the final non-local weights obtained by our model.}\label{FigChangWeights}
\end{figure}

\begin{figure*}[ht]
	\centering
	\subfigure[]{
		\begin{minipage}[t]{0.78\linewidth}
			\centering
			\includegraphics[width=0.24\textwidth]{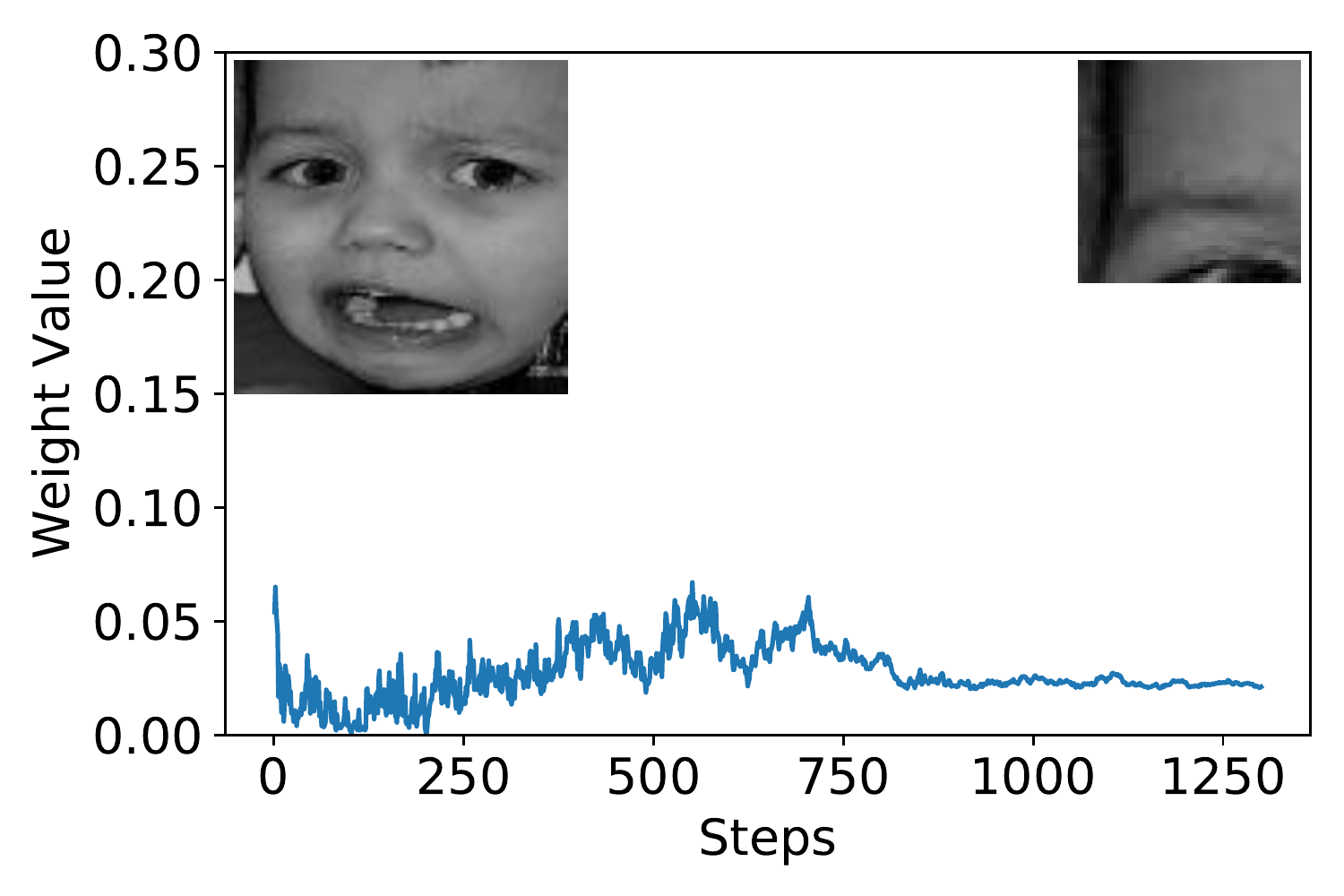}
			\includegraphics[width=0.24\textwidth]{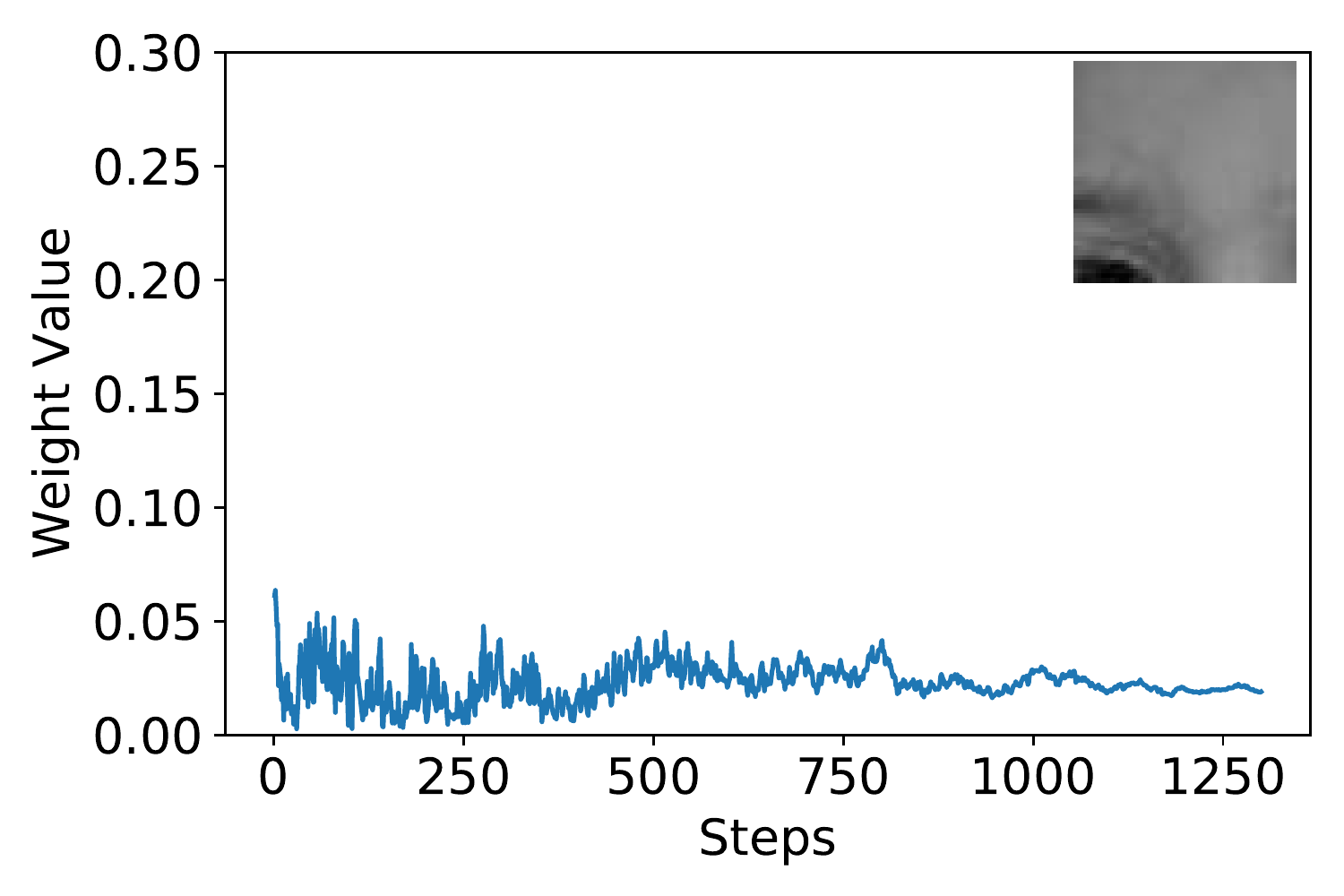}
			\includegraphics[width=0.24\textwidth]{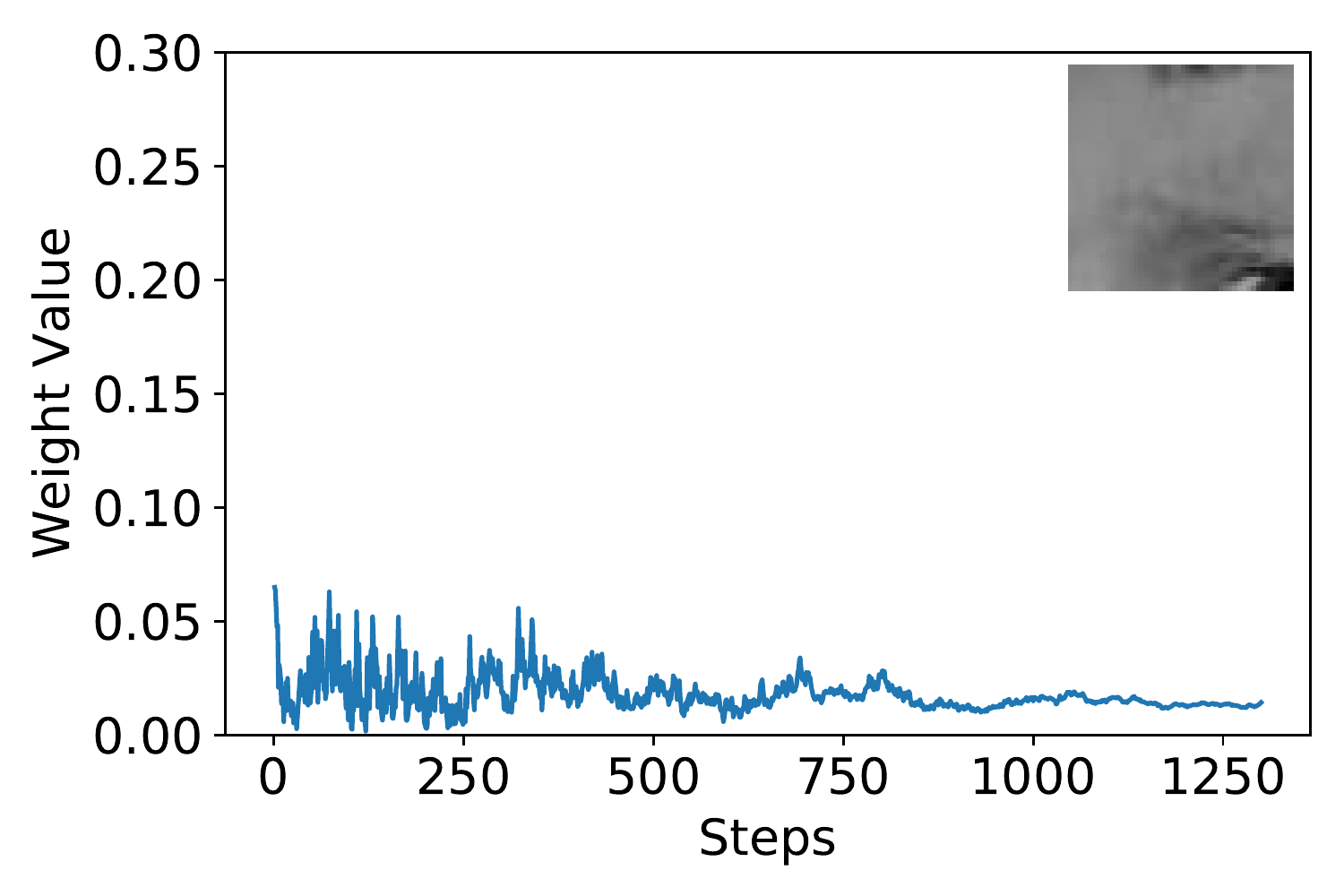}
			\includegraphics[width=0.24\textwidth]{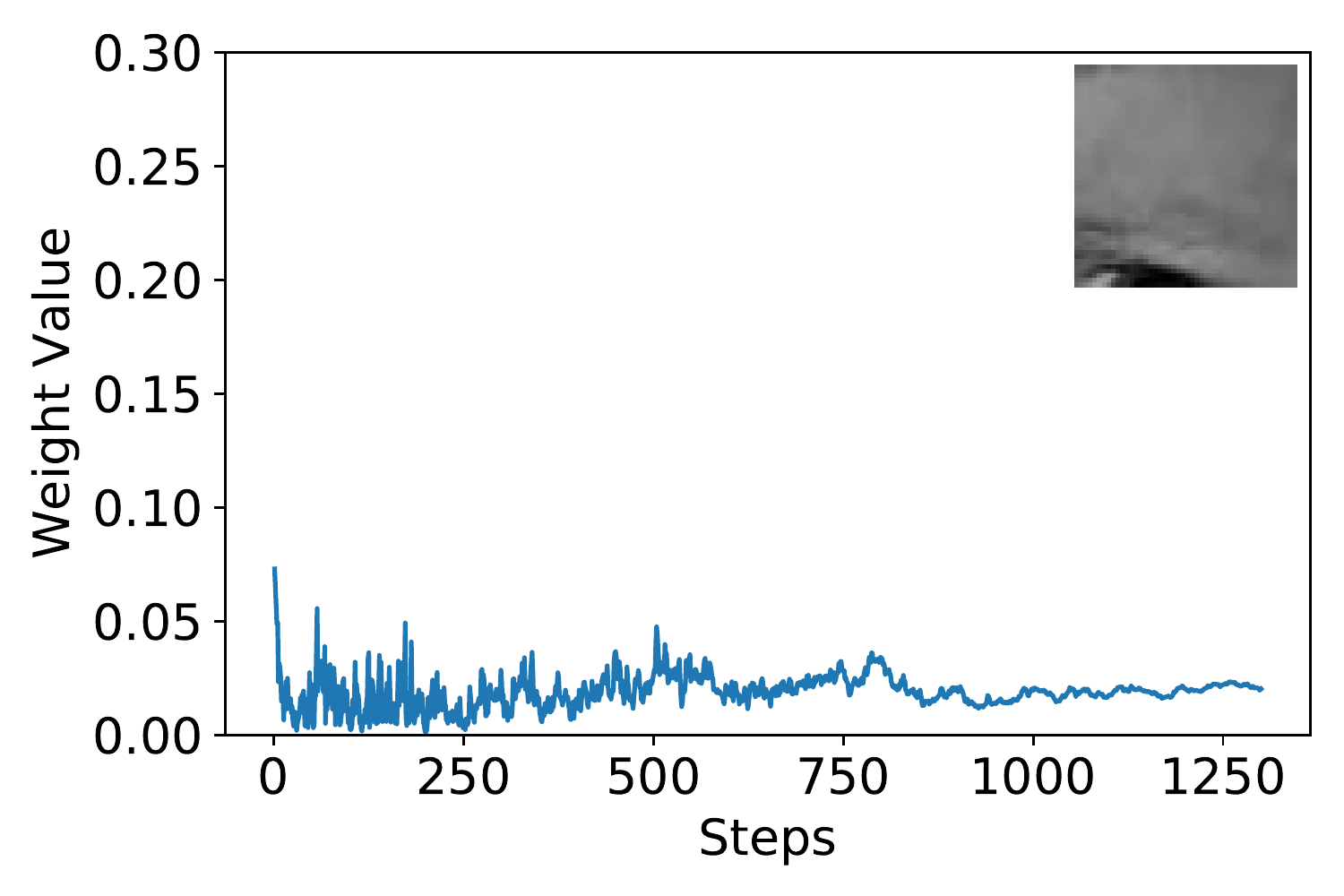}\\
			
			\includegraphics[width=0.24\textwidth]{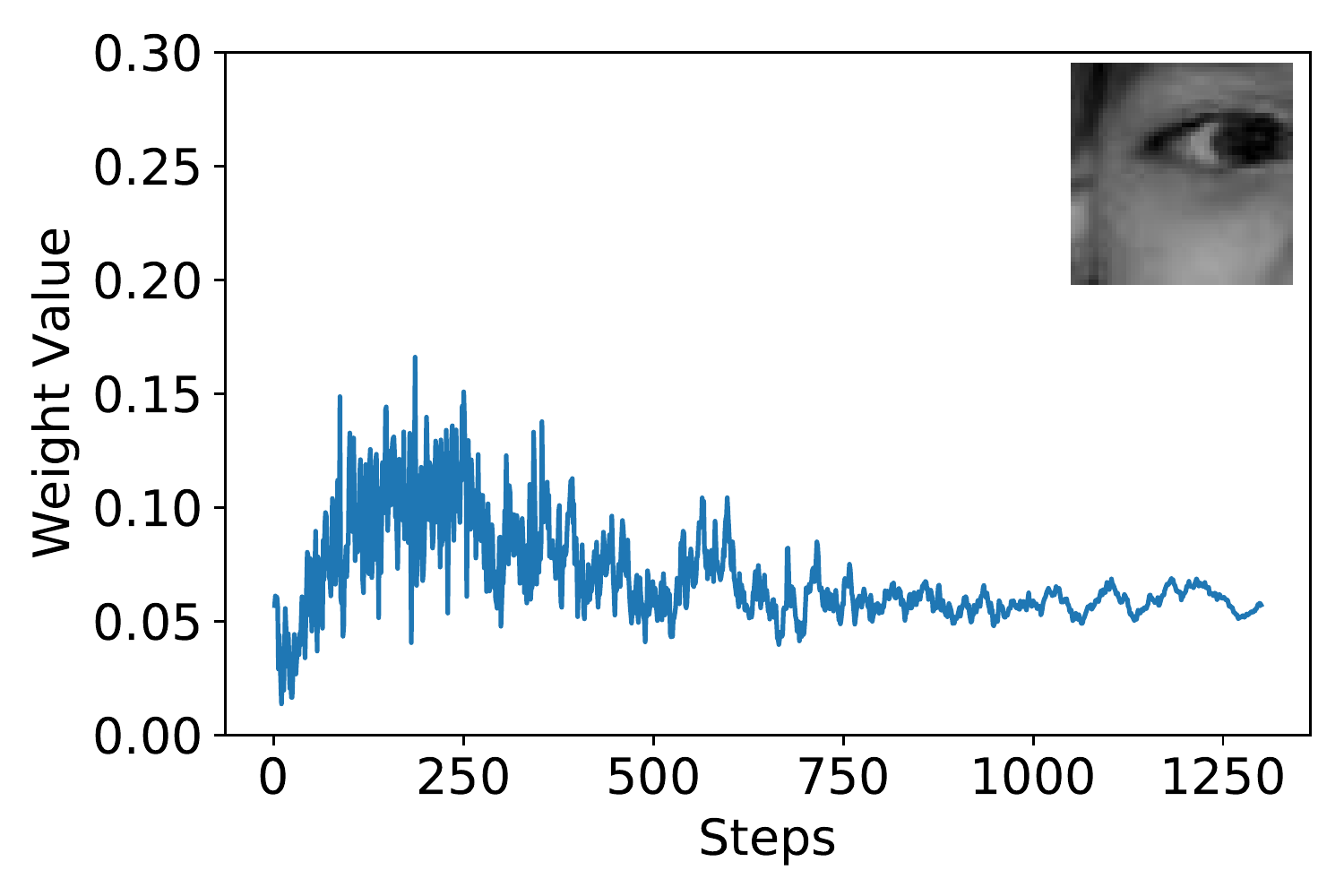}
			\includegraphics[width=0.24\textwidth]{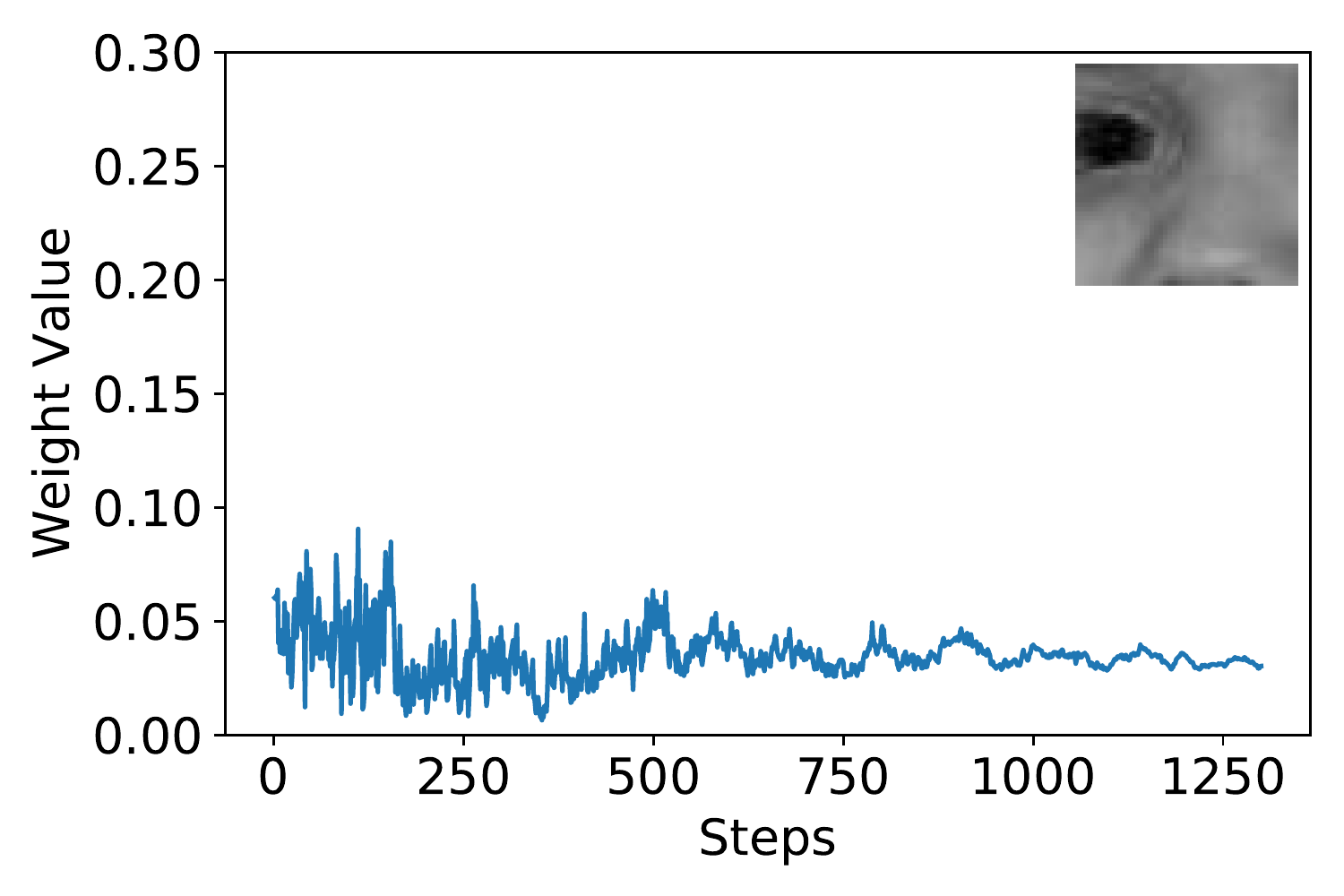}
			\includegraphics[width=0.24\textwidth]{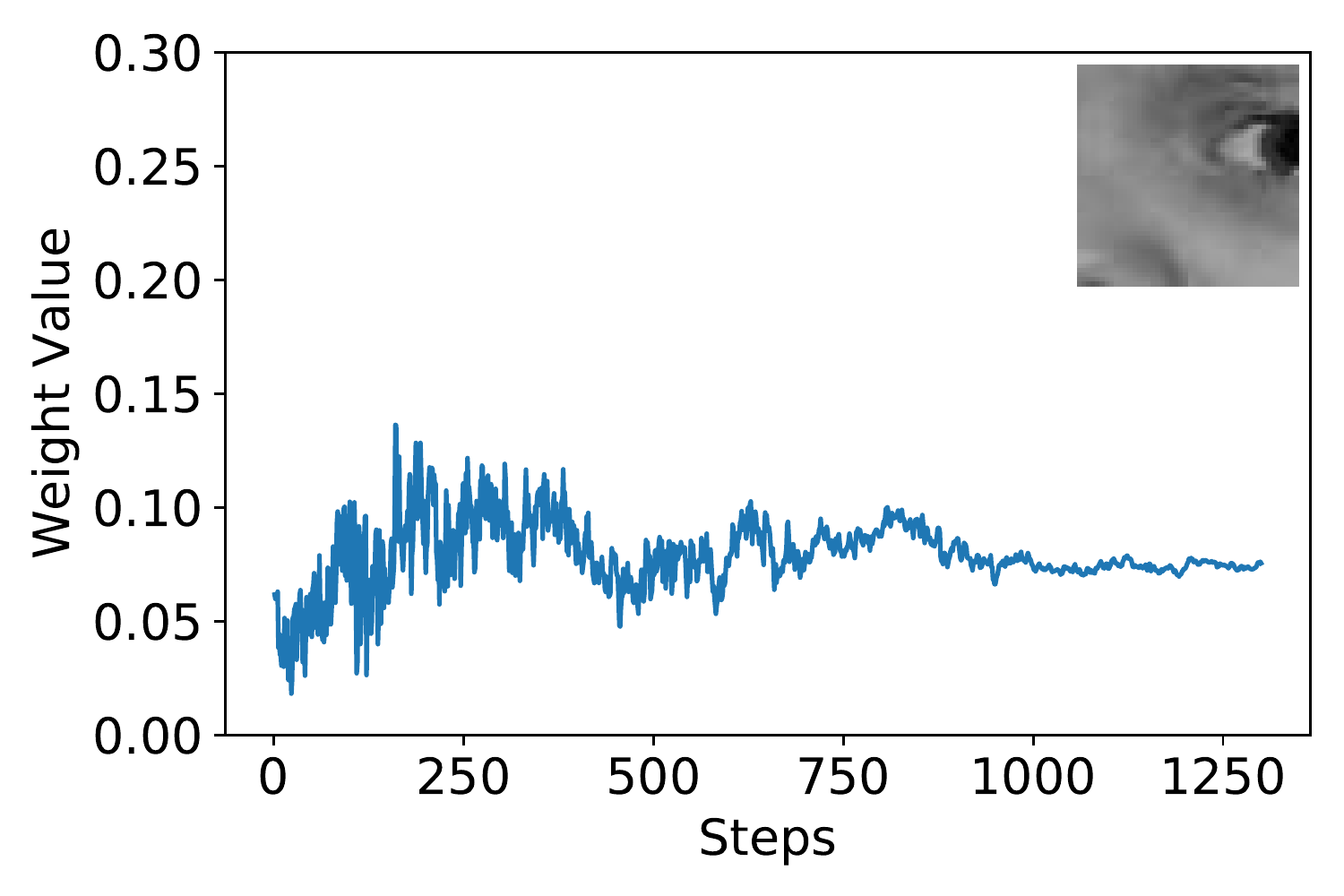}
			\includegraphics[width=0.24\textwidth]{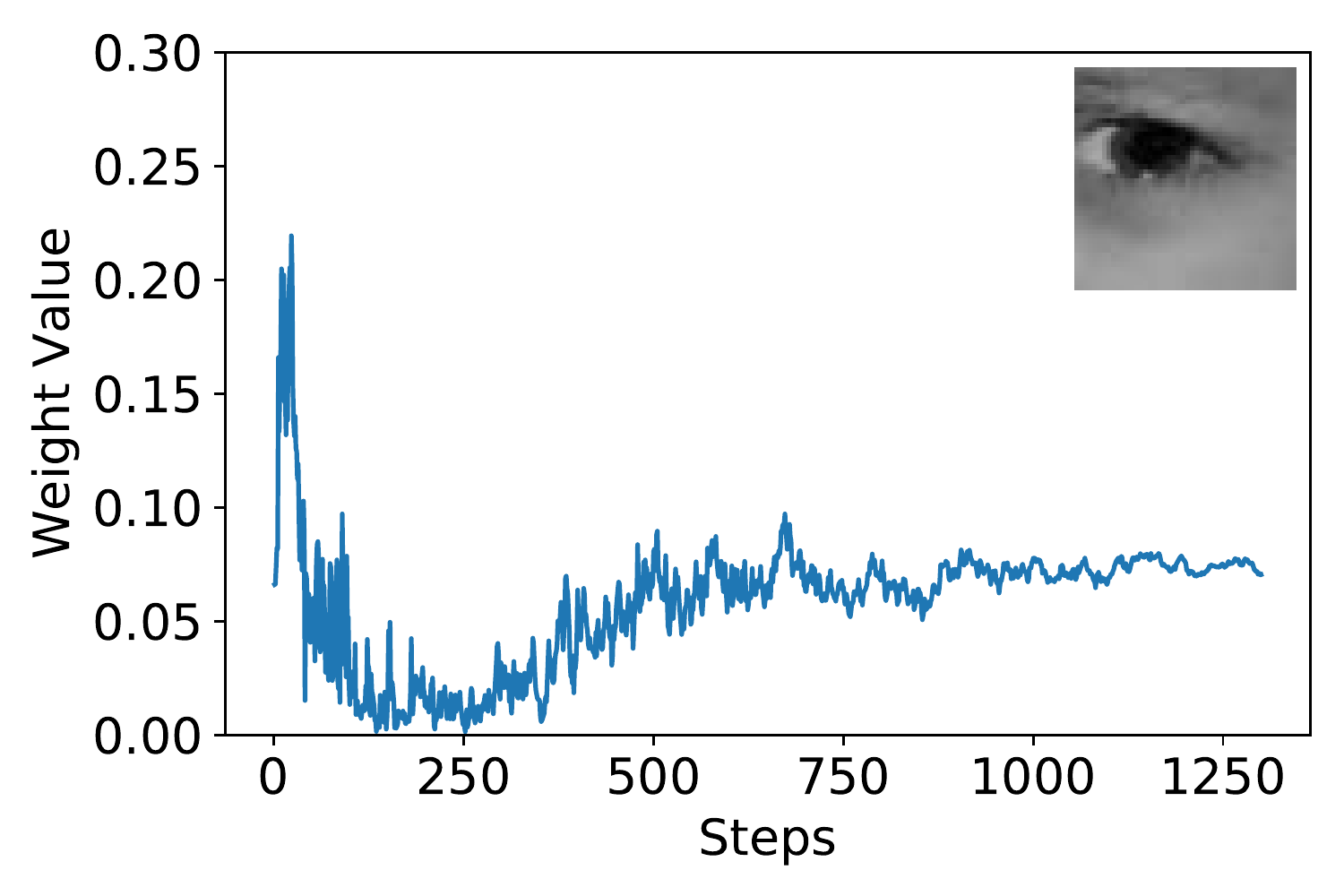}\\
			
			\includegraphics[width=0.24\textwidth]{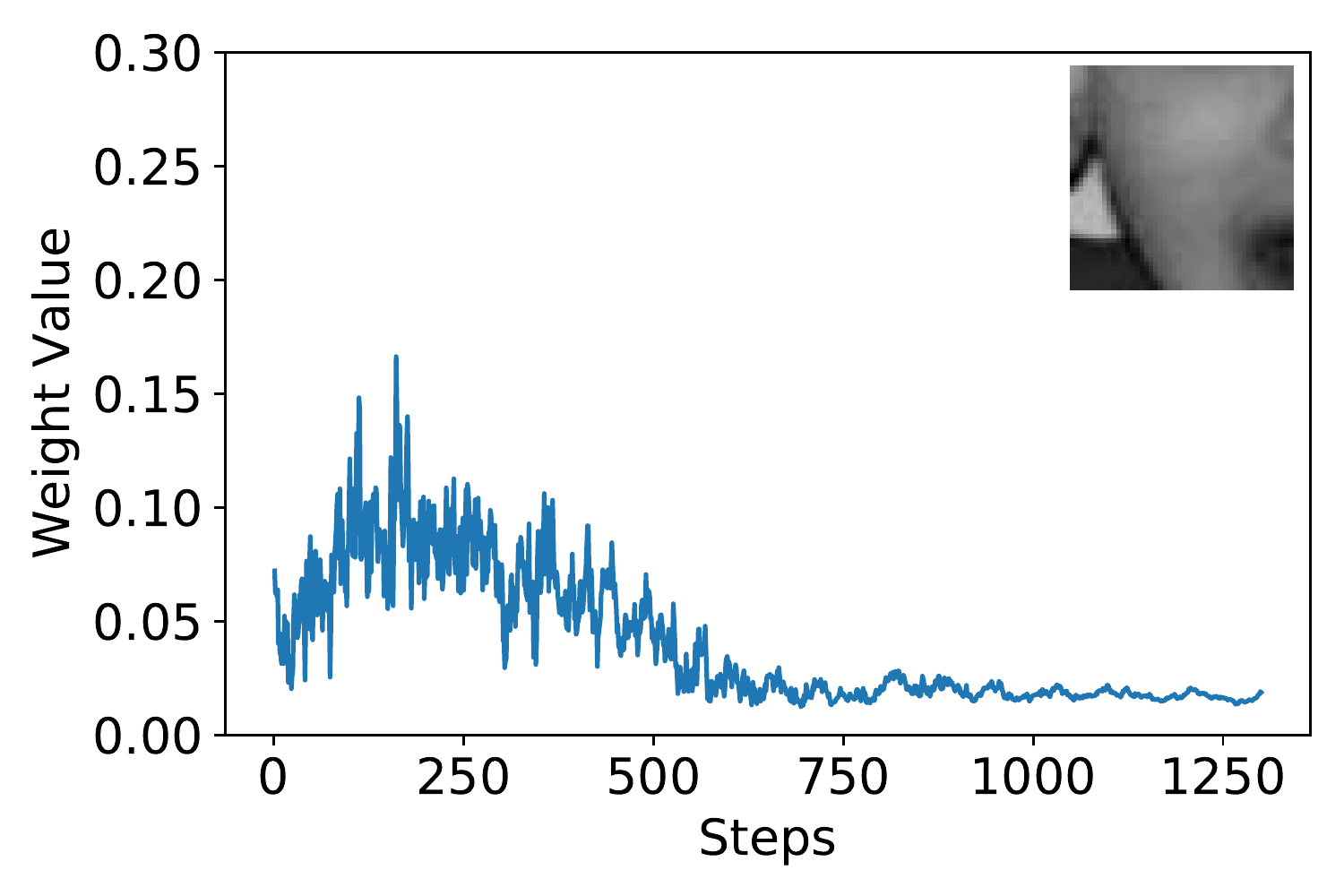}
			\includegraphics[width=0.24\textwidth]{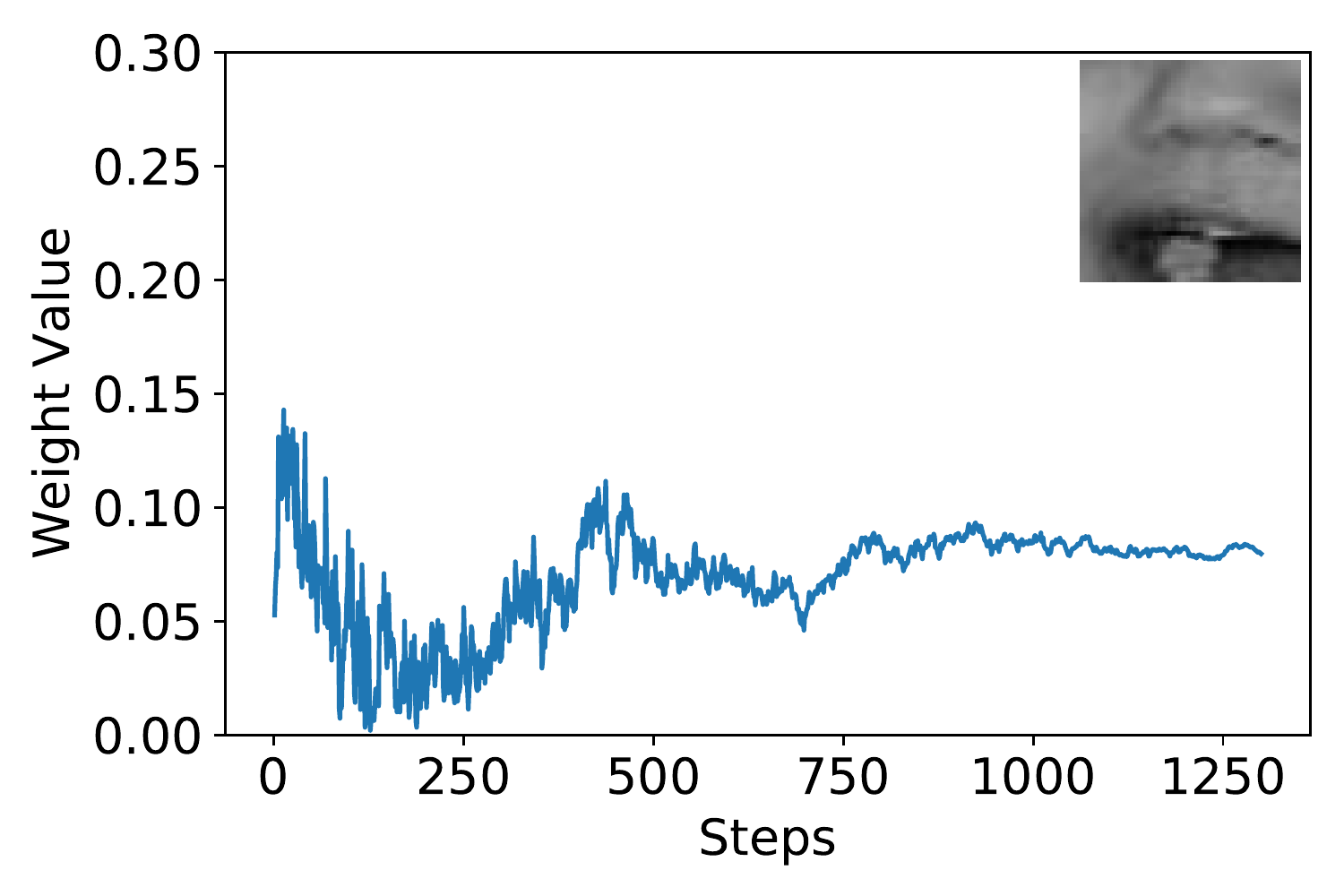}
			\includegraphics[width=0.24\textwidth]{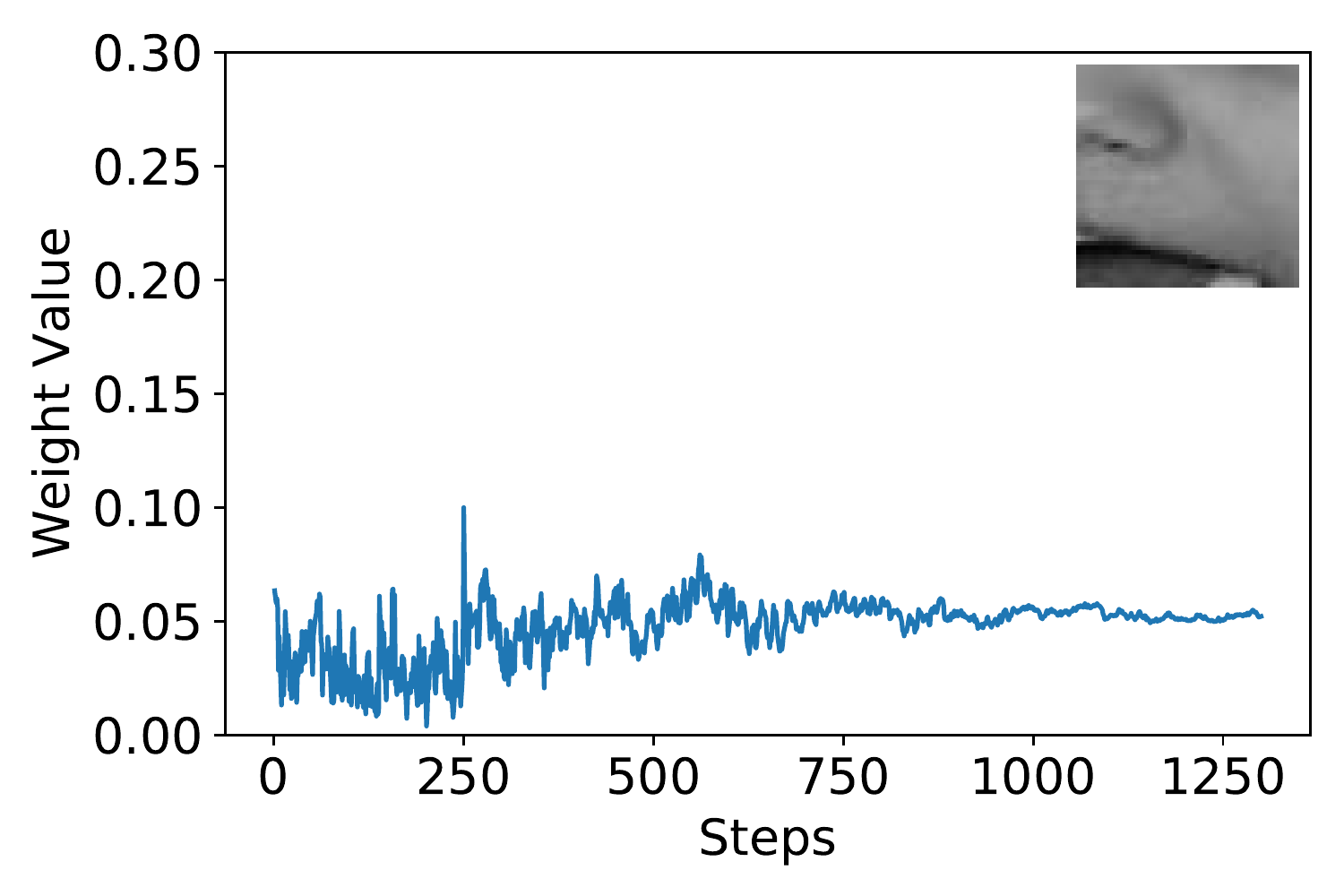}
			\includegraphics[width=0.24\textwidth]{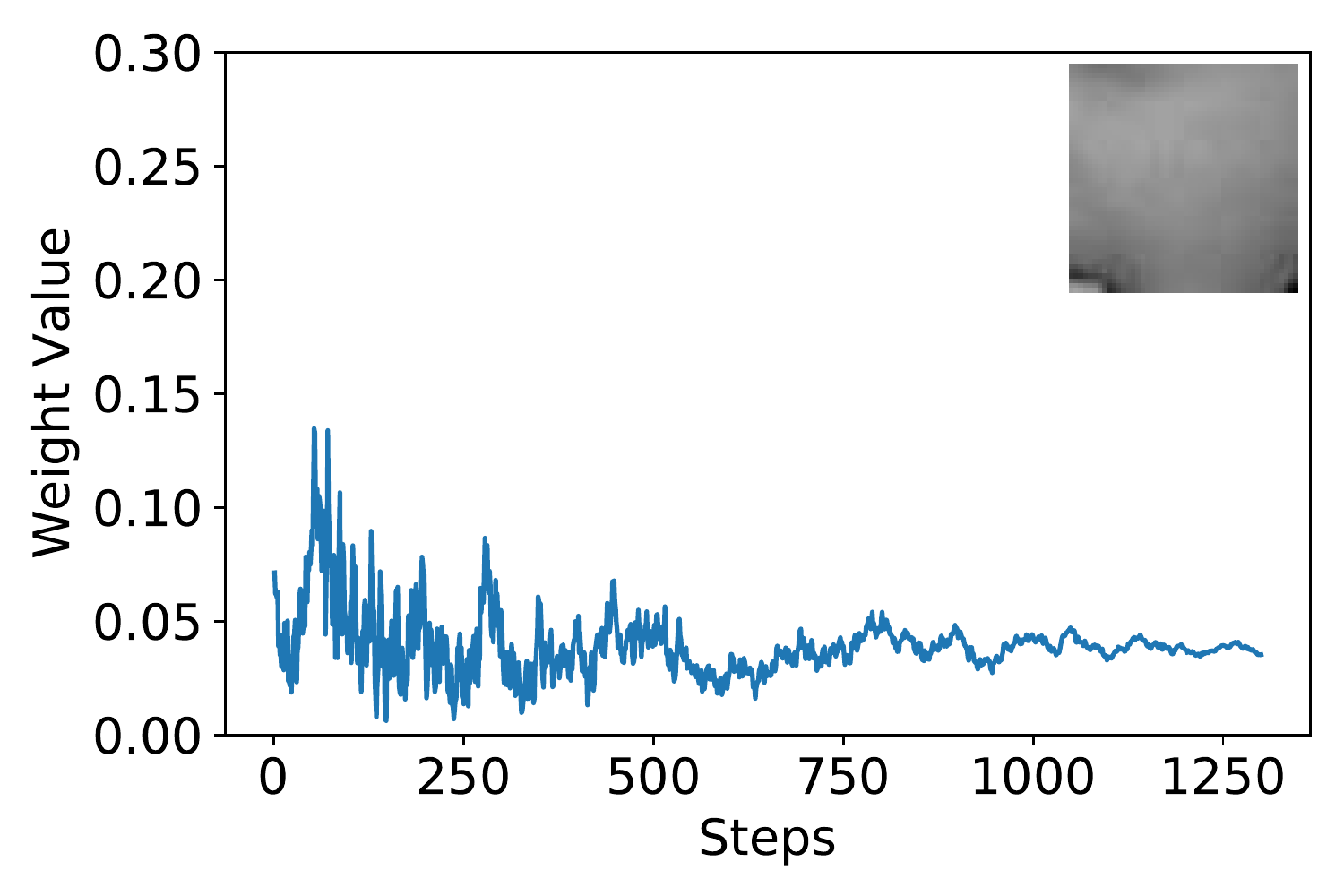}\\
			
			\includegraphics[width=0.24\textwidth]{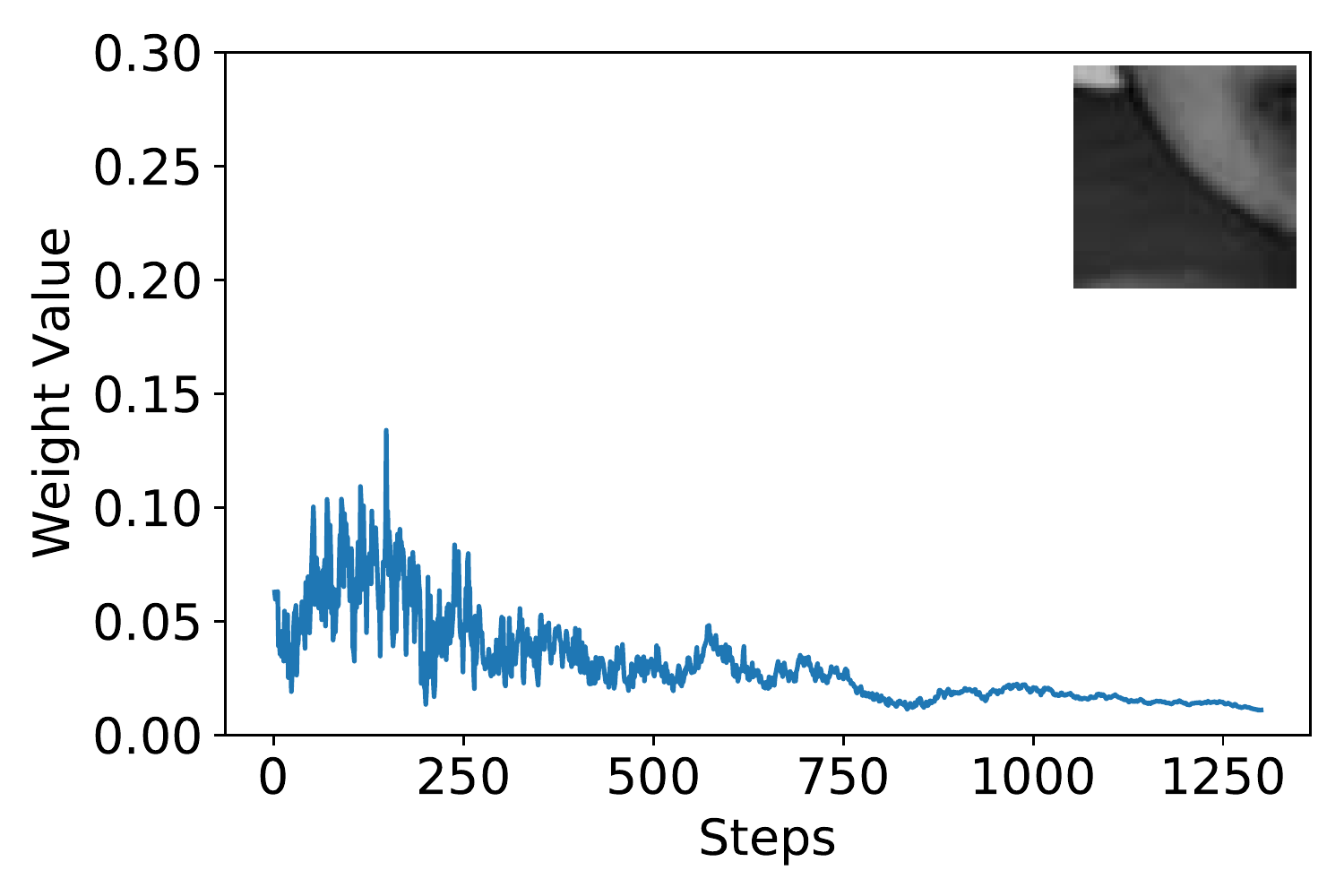}
			\includegraphics[width=0.24\textwidth]{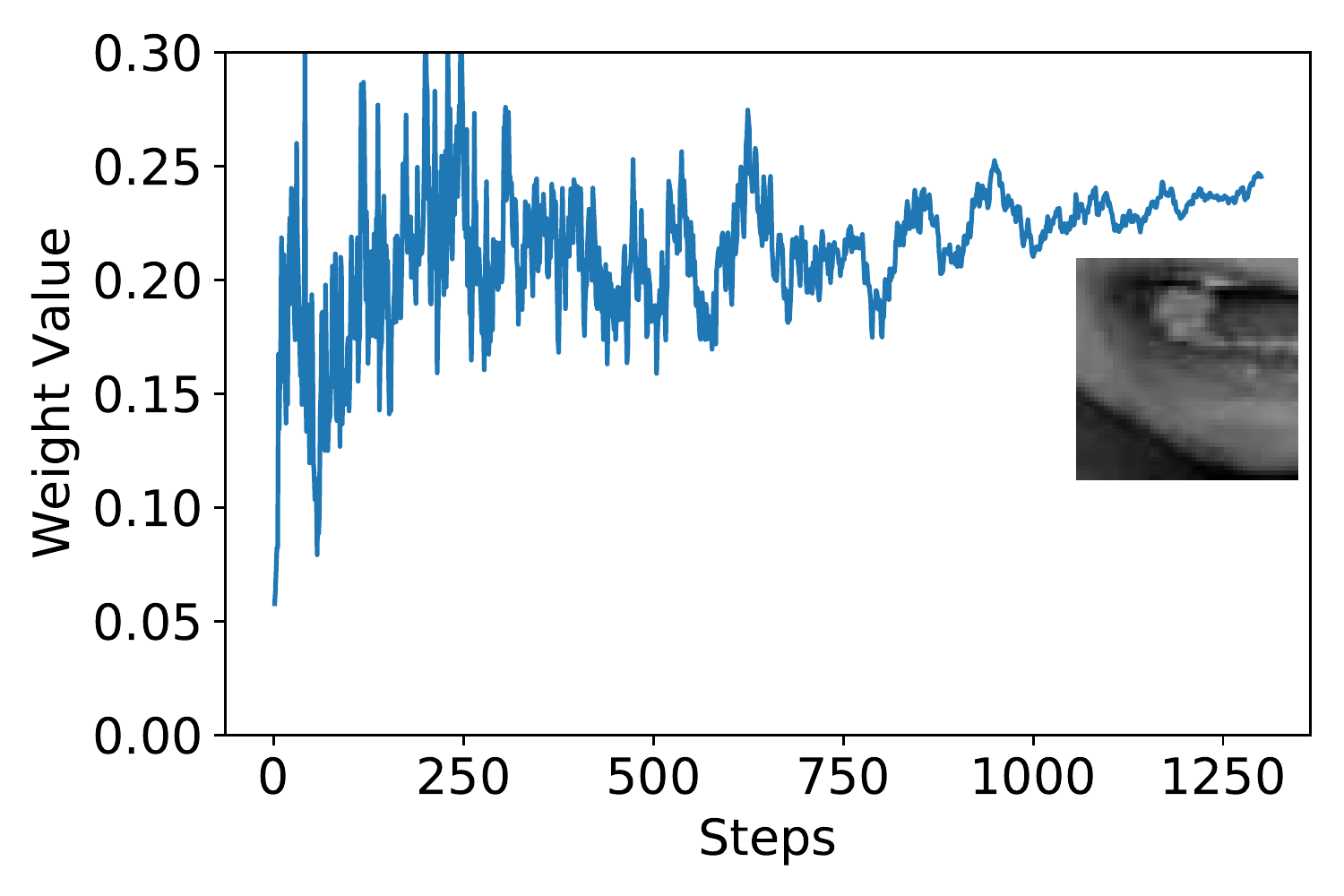}
			\includegraphics[width=0.24\textwidth]{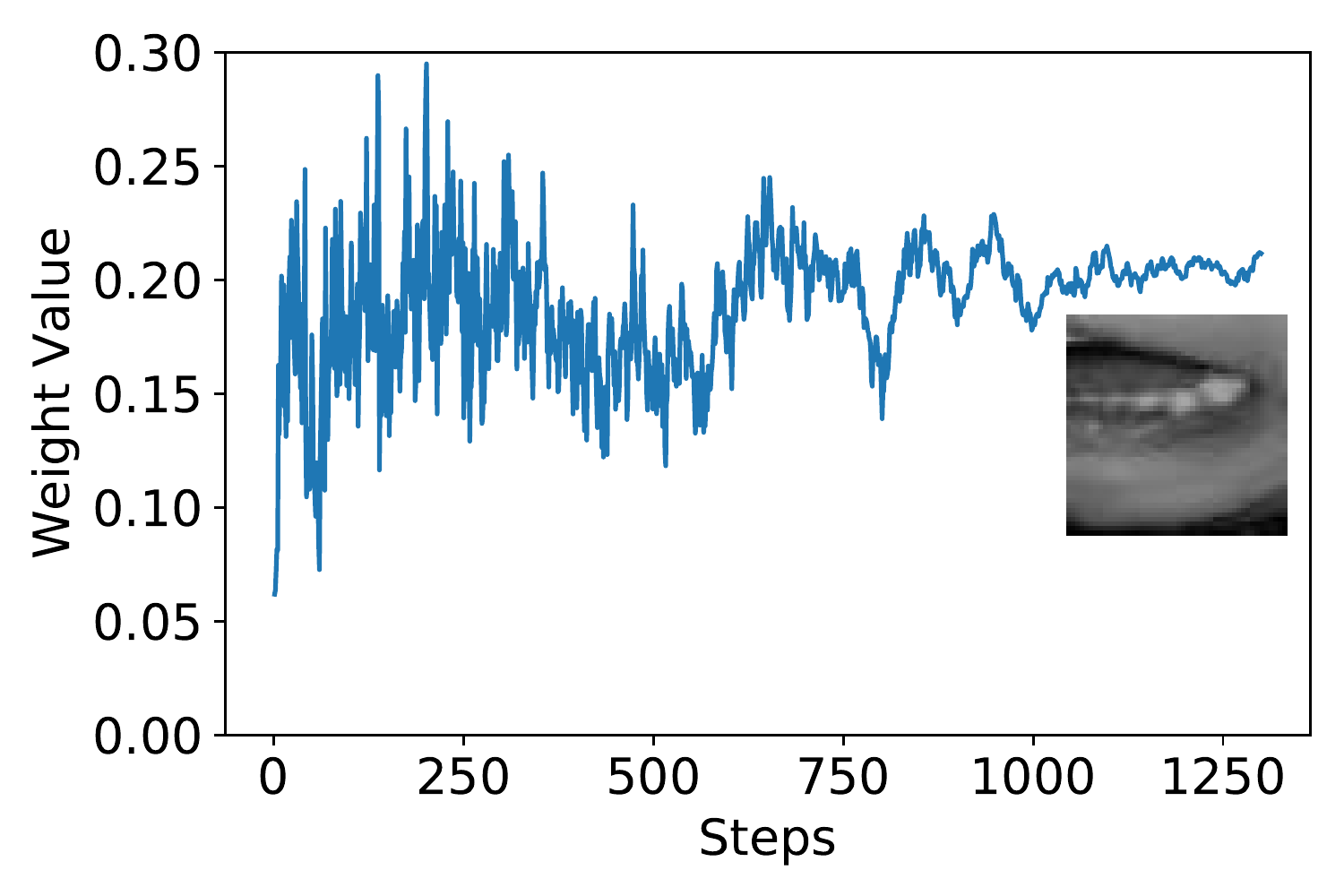}
			\includegraphics[width=0.24\textwidth]{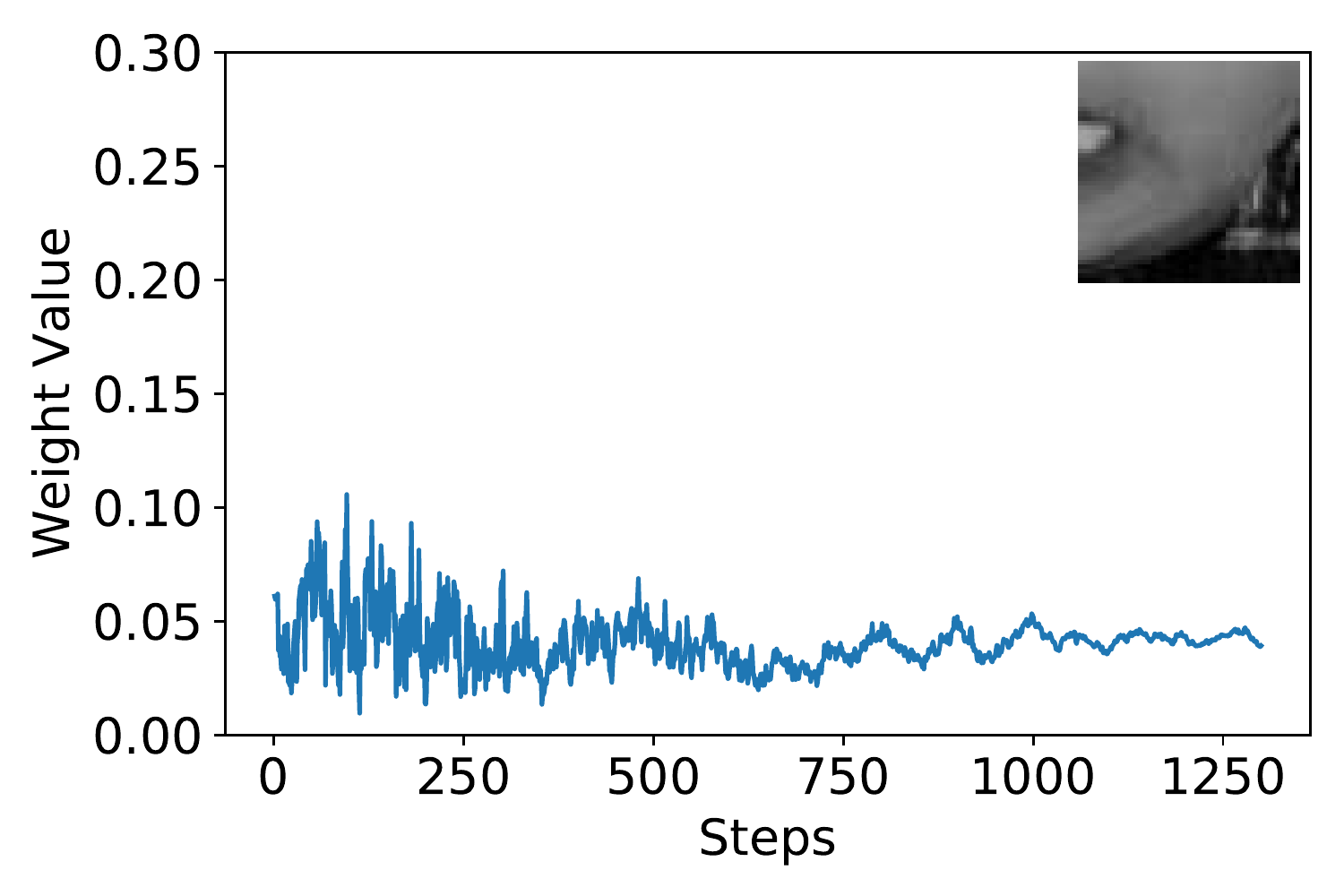} \\
		\end{minipage}
	}	
	\subfigure[]{
		\begin{minipage}[t]{0.78\linewidth}
			\centering
			\includegraphics[width=0.24\textwidth]{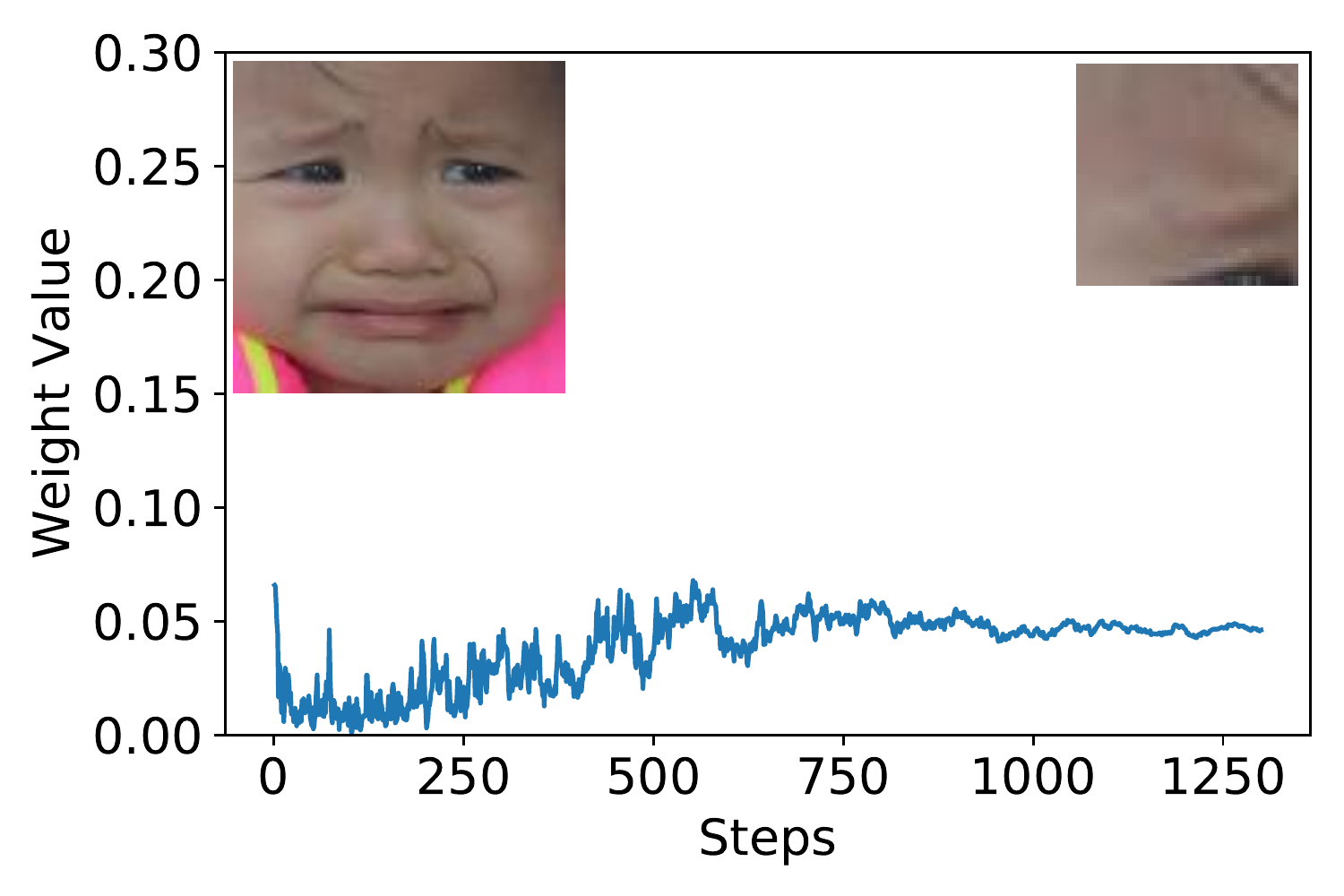}
			\includegraphics[width=0.24\textwidth]{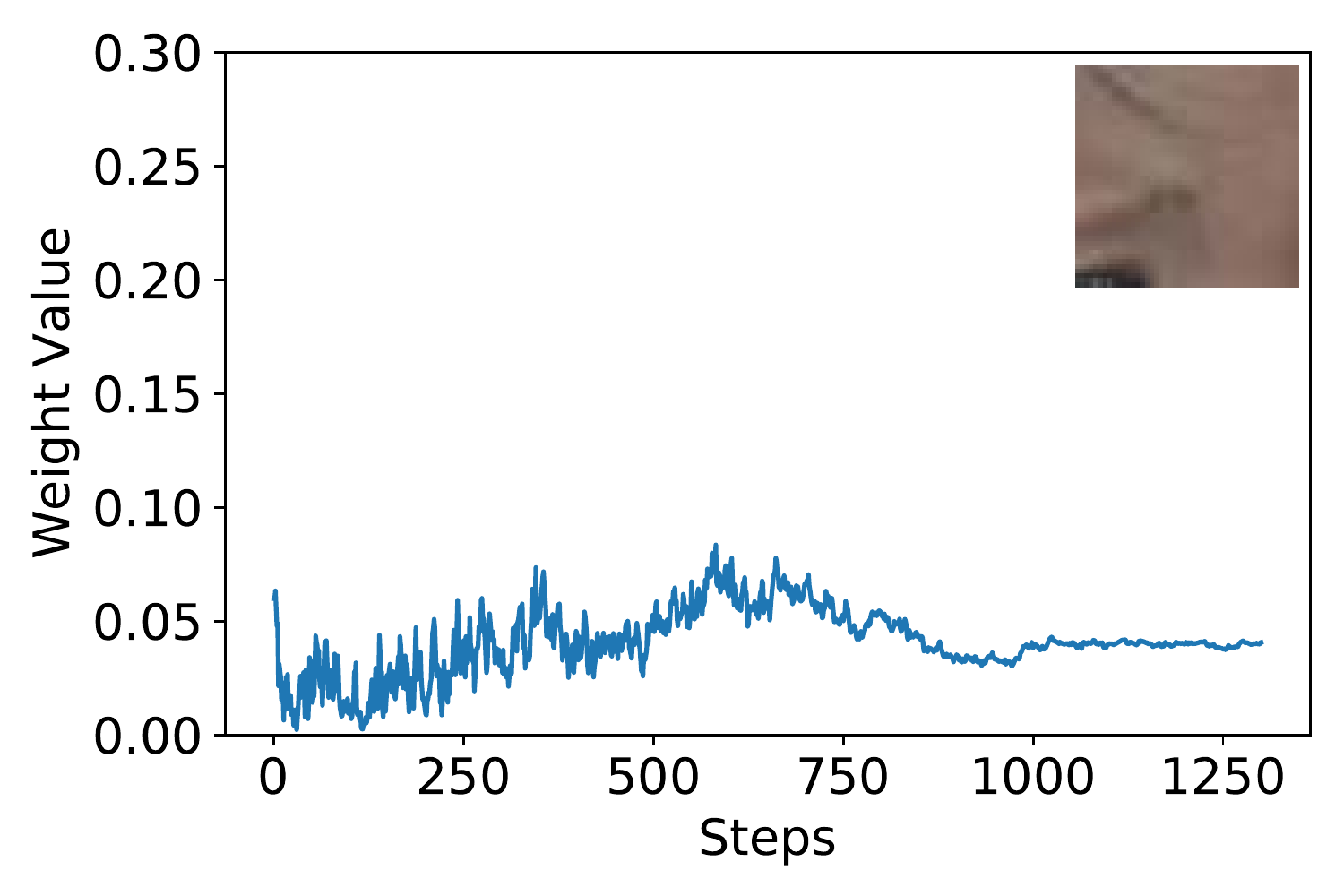}
			\includegraphics[width=0.24\textwidth]{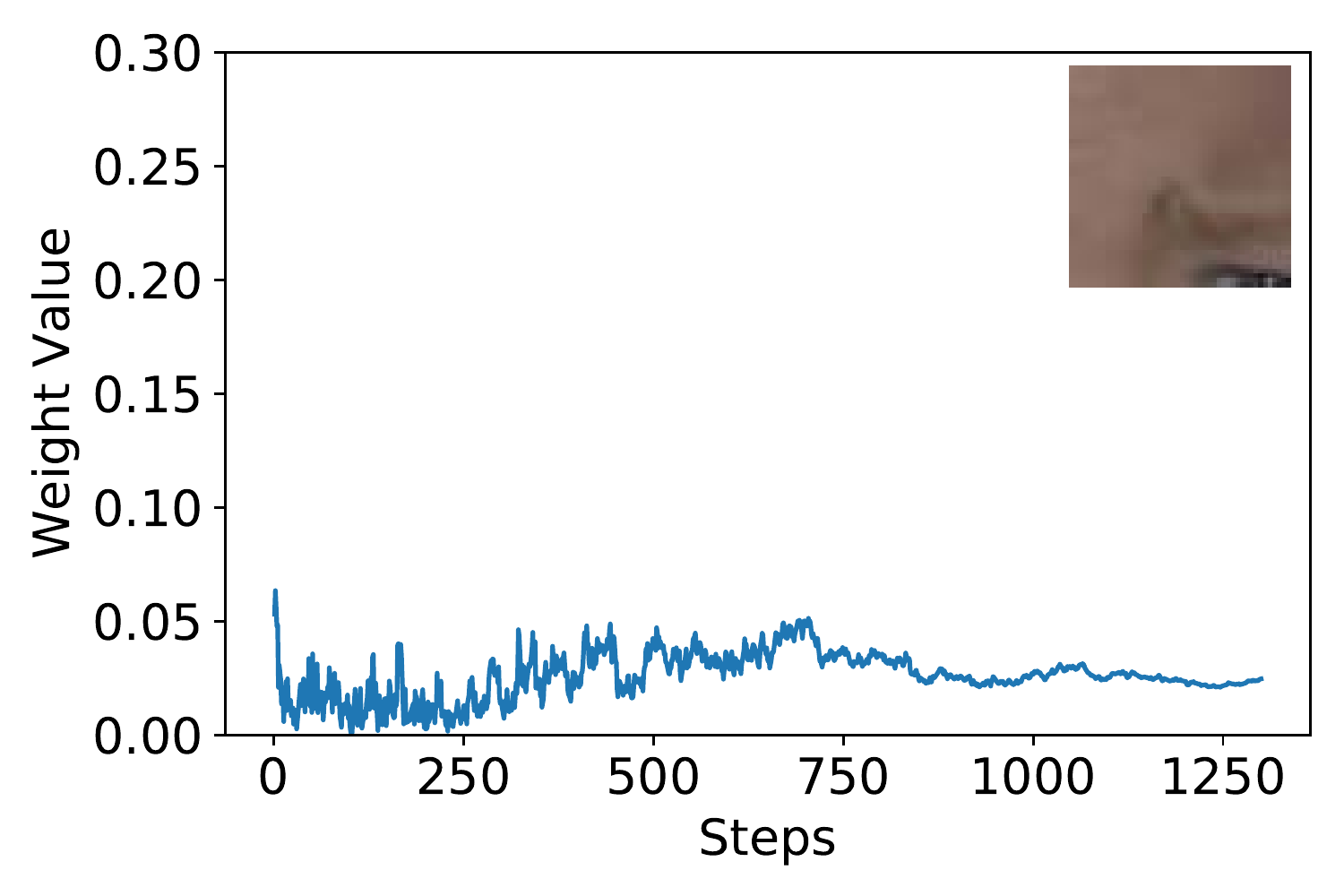}
			\includegraphics[width=0.24\textwidth]{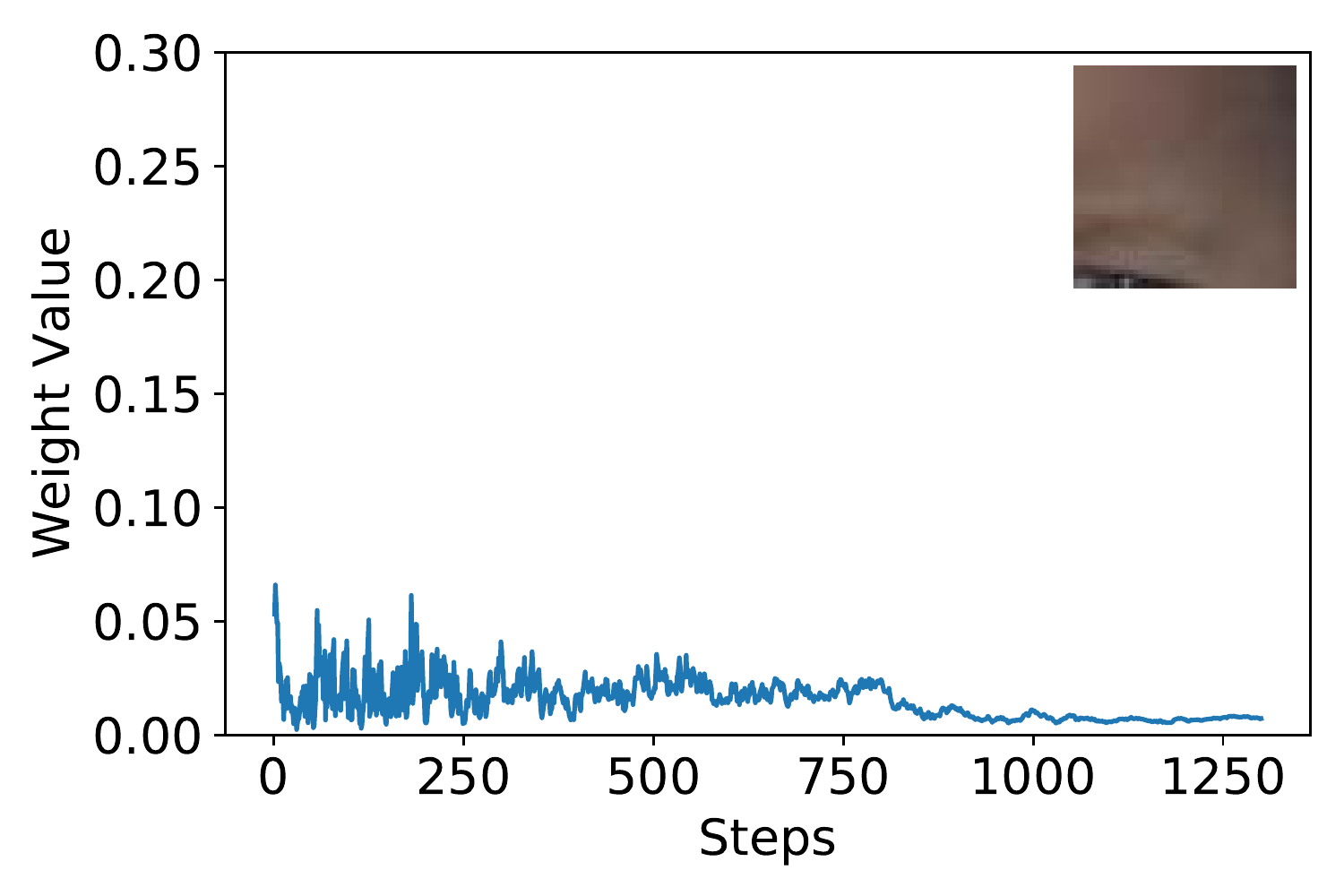}\\
			
			\includegraphics[width=0.24\textwidth]{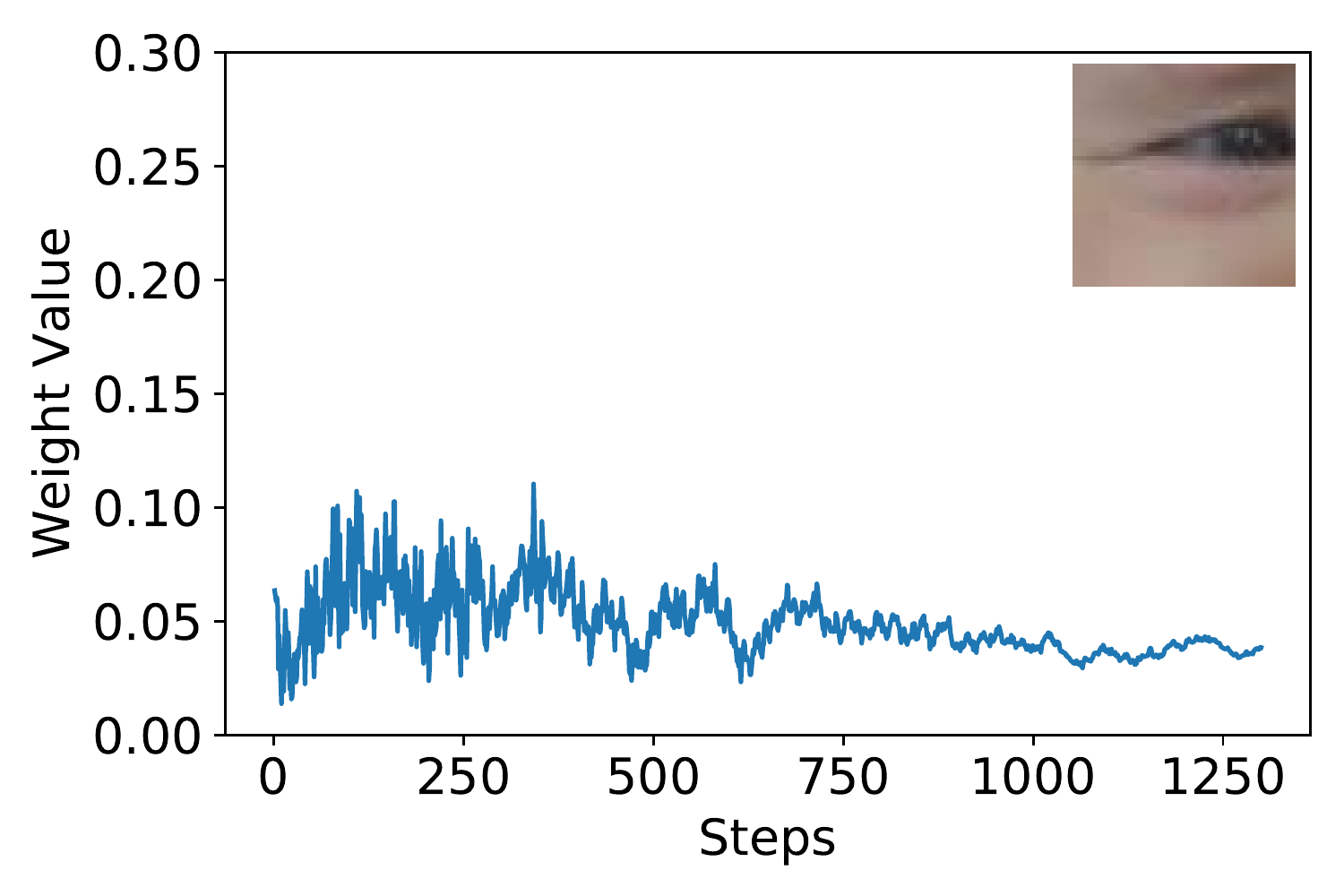}
			\includegraphics[width=0.24\textwidth]{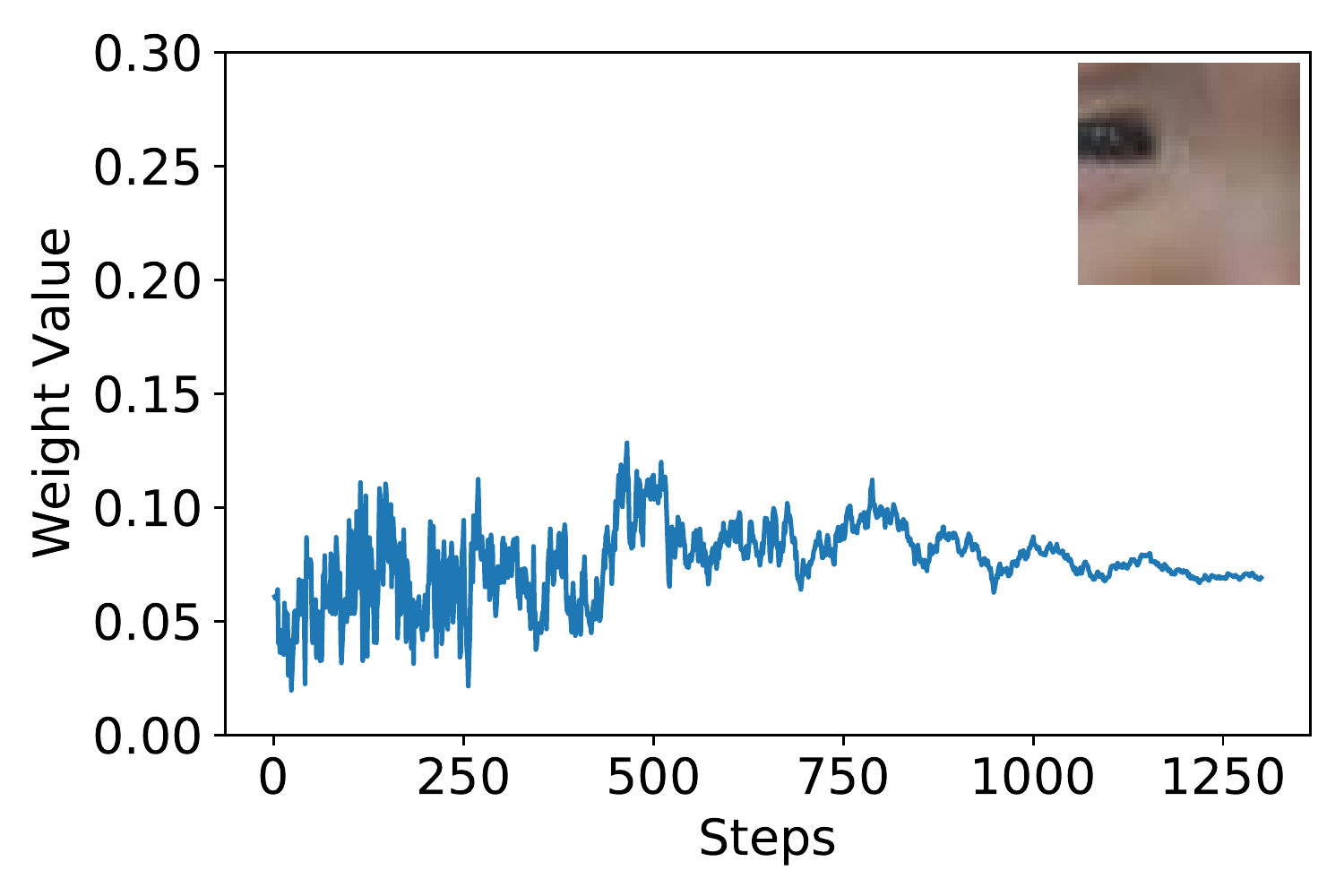}
			\includegraphics[width=0.24\textwidth]{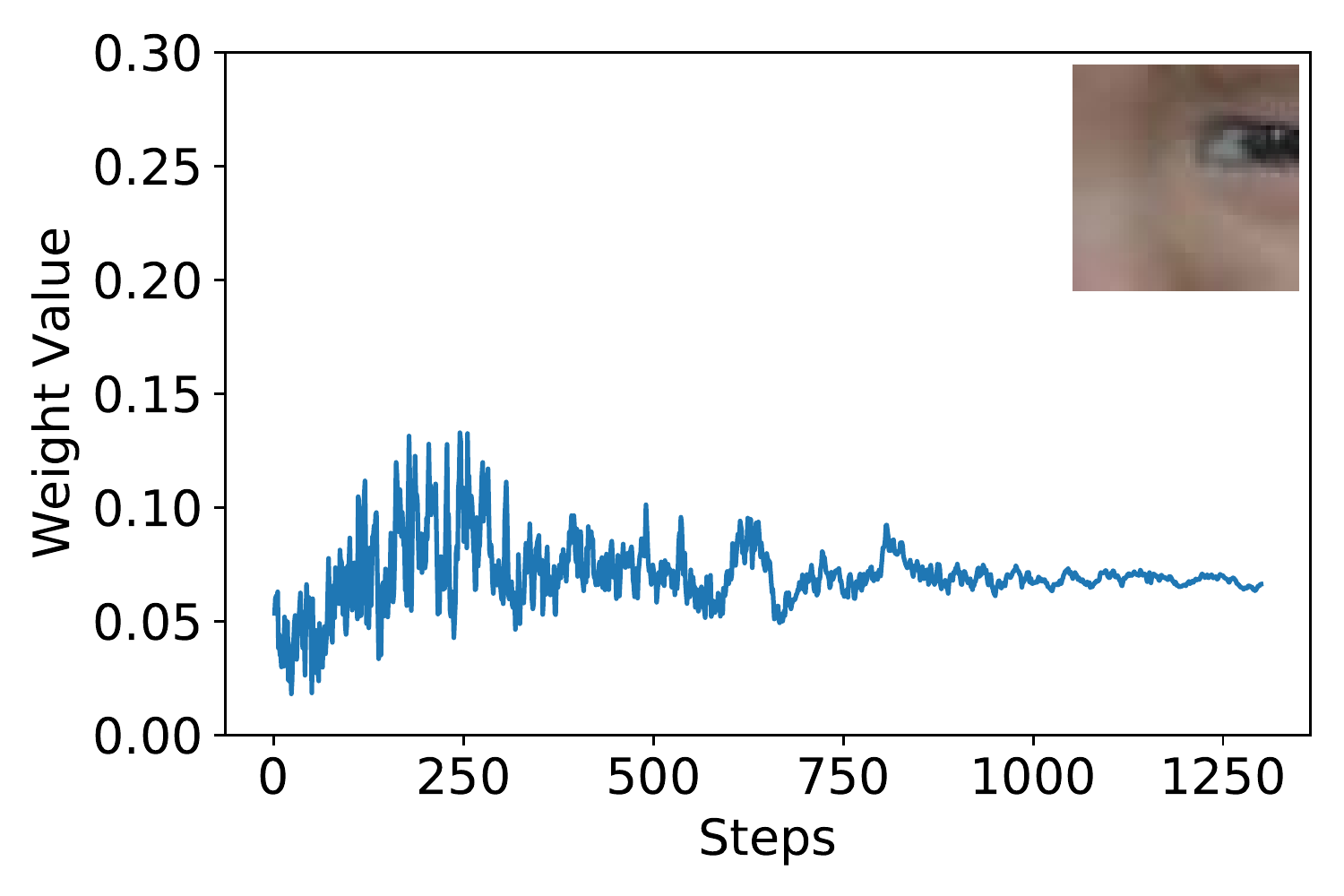}
			\includegraphics[width=0.24\textwidth]{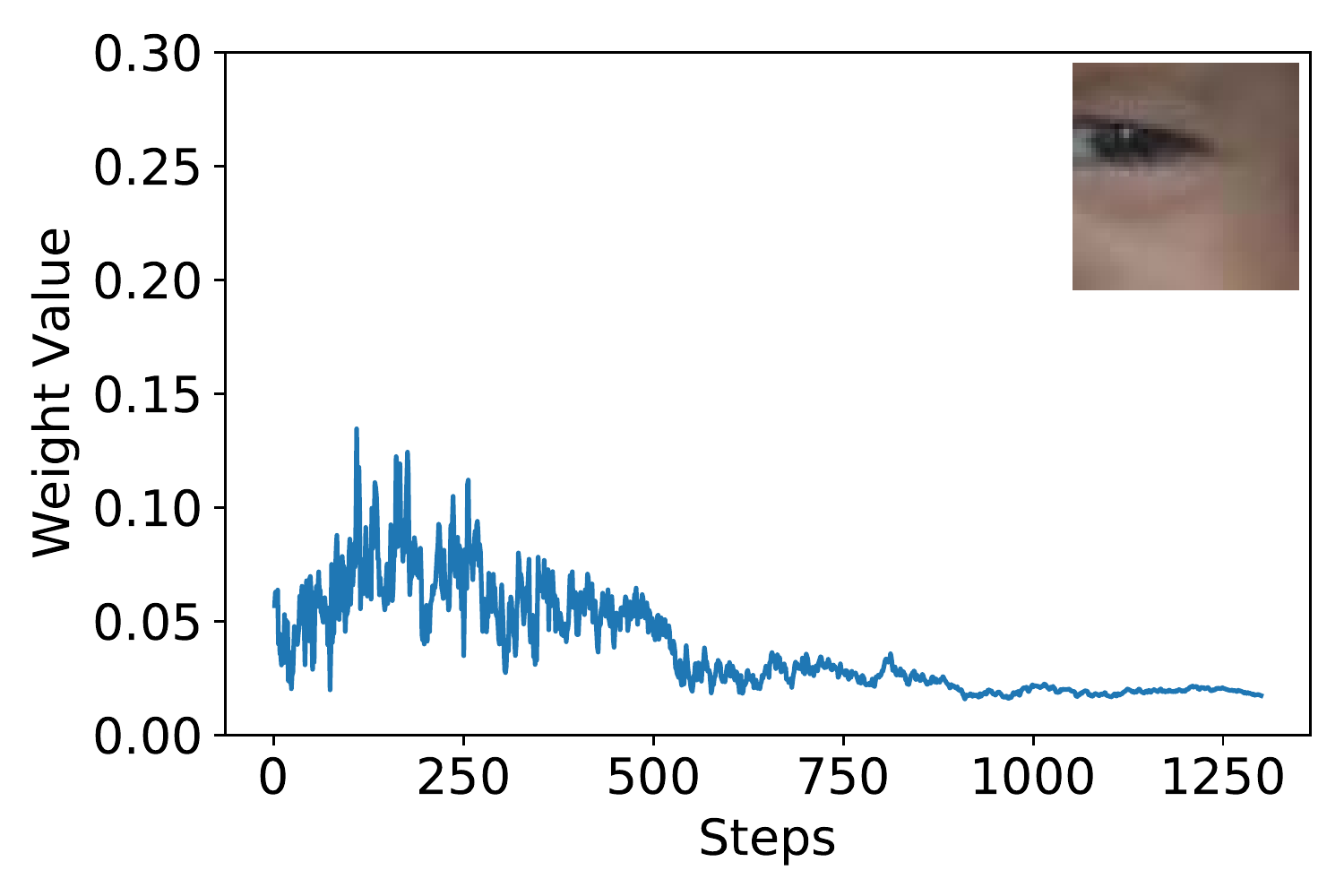}\\
			
			\includegraphics[width=0.24\textwidth]{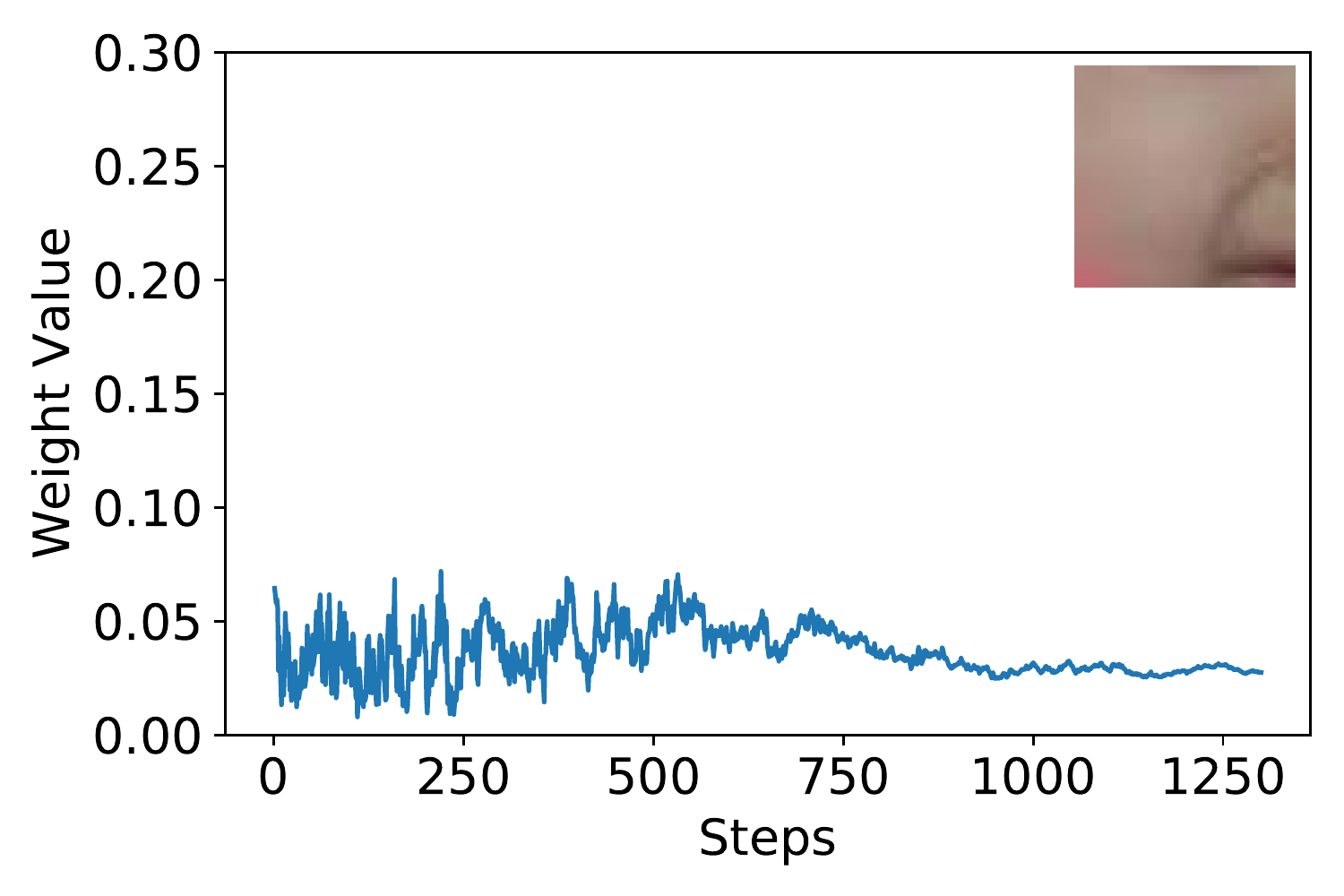}
			\includegraphics[width=0.24\textwidth]{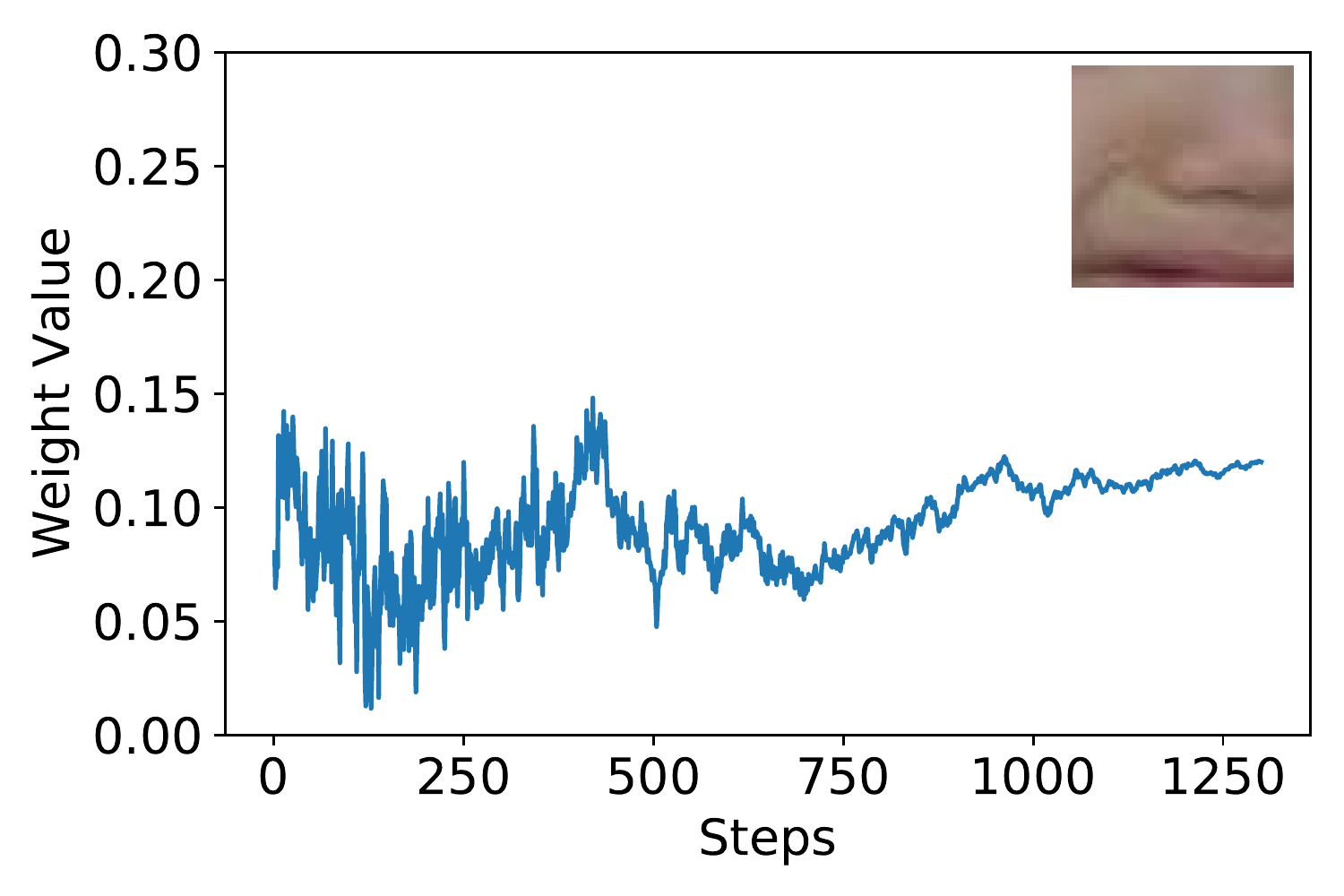}
			\includegraphics[width=0.24\textwidth]{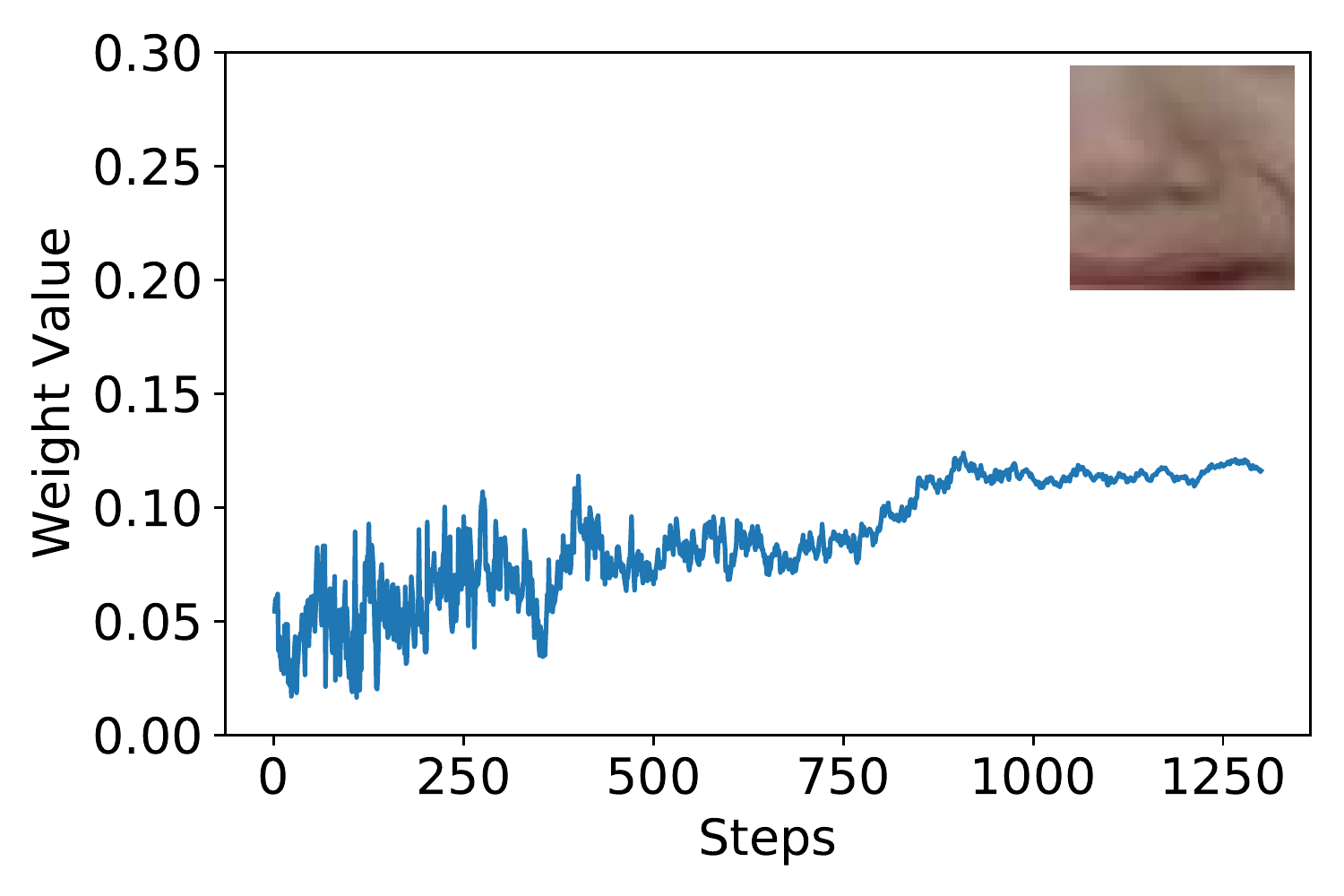}
			\includegraphics[width=0.24\textwidth]{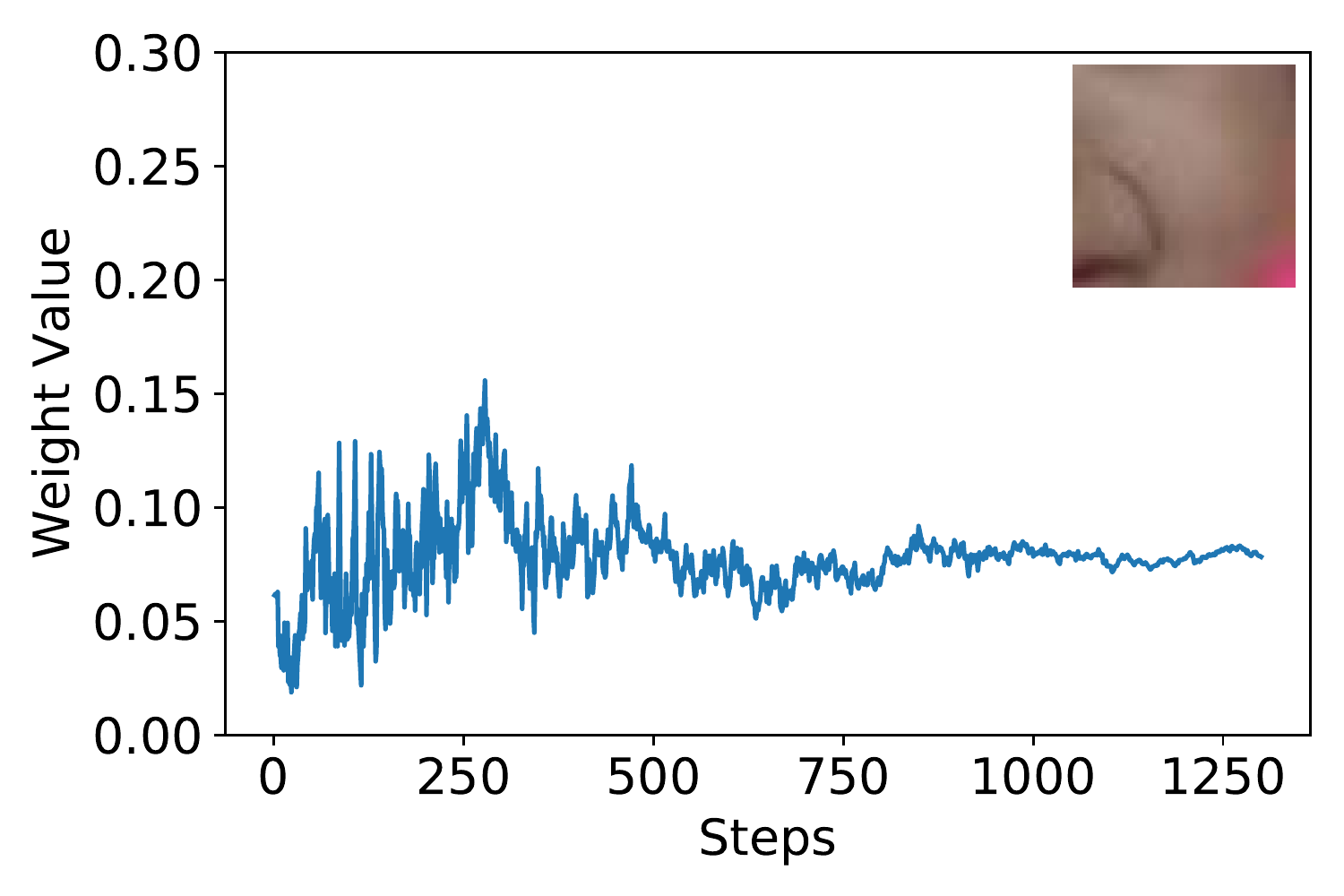}\\
			
			\includegraphics[width=0.24\textwidth]{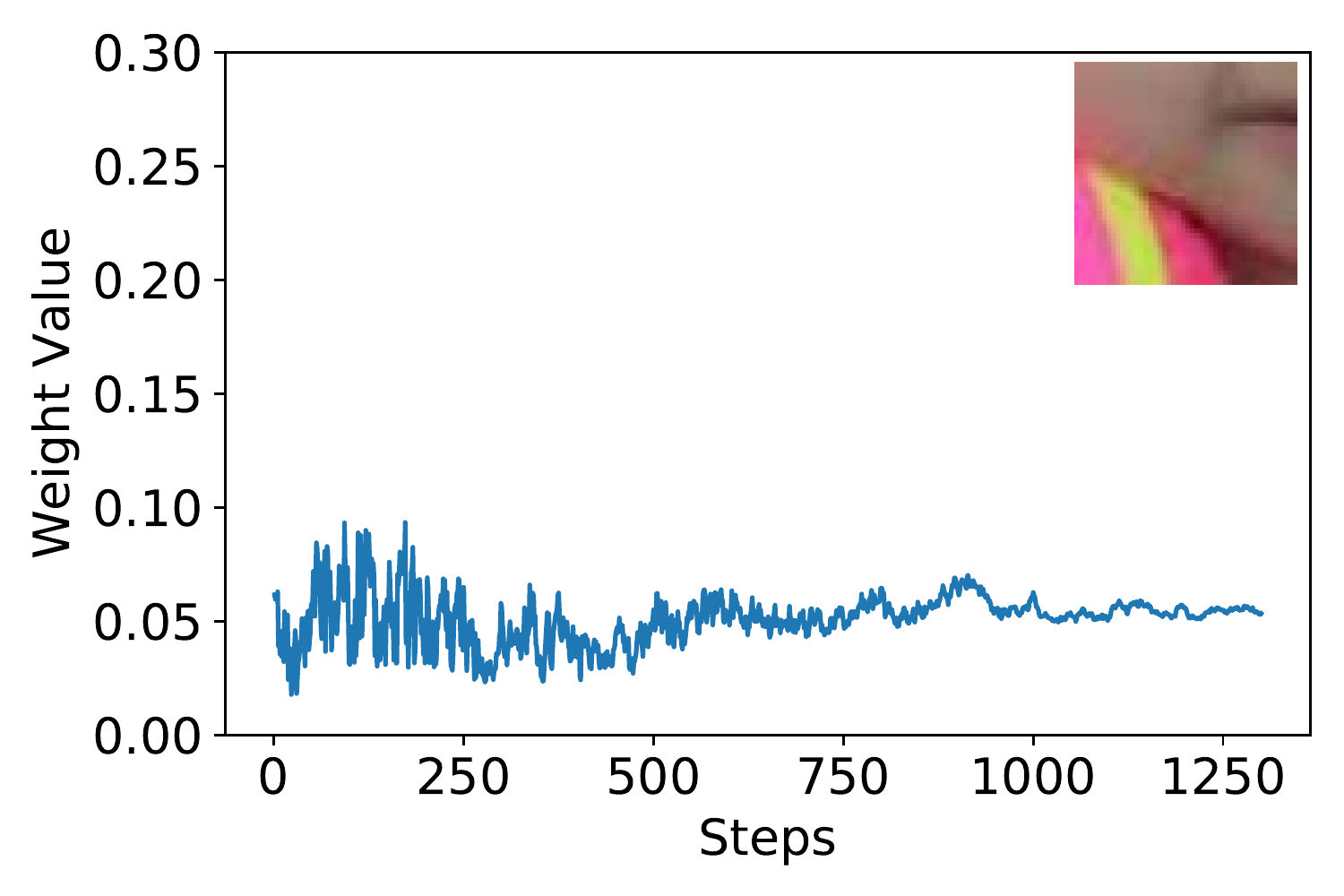}
			\includegraphics[width=0.24\textwidth]{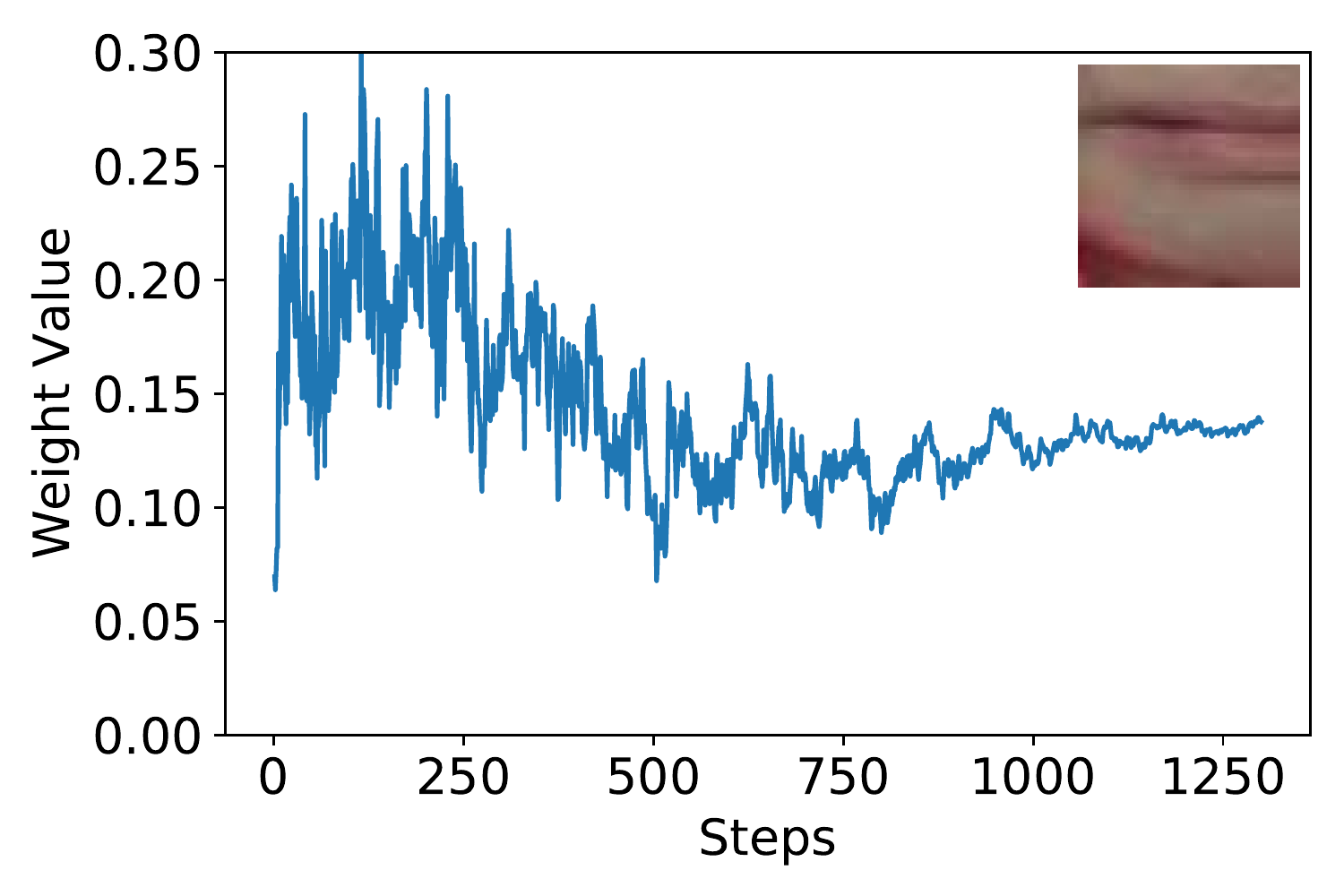}
			\includegraphics[width=0.24\textwidth]{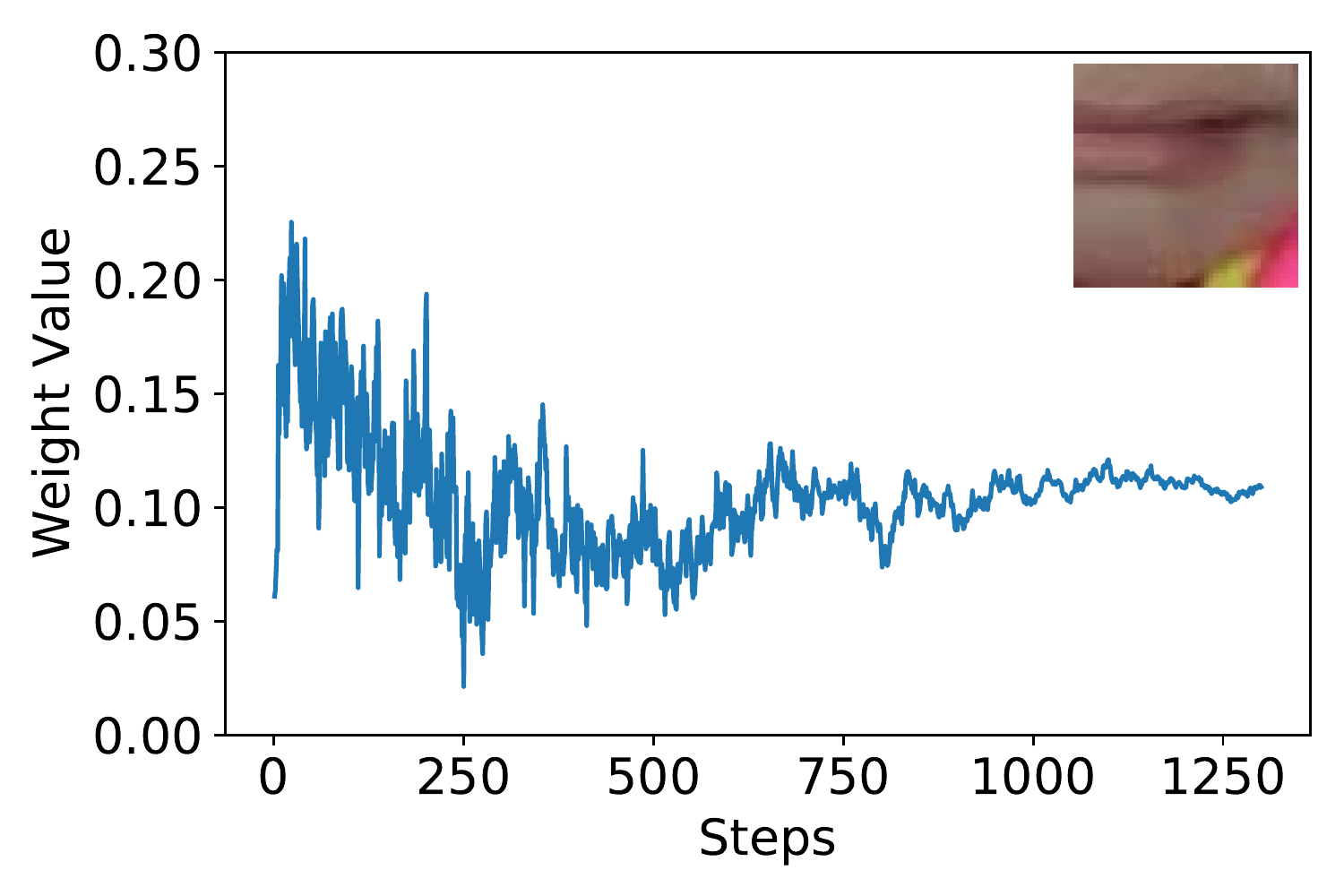}
			\includegraphics[width=0.24\textwidth]{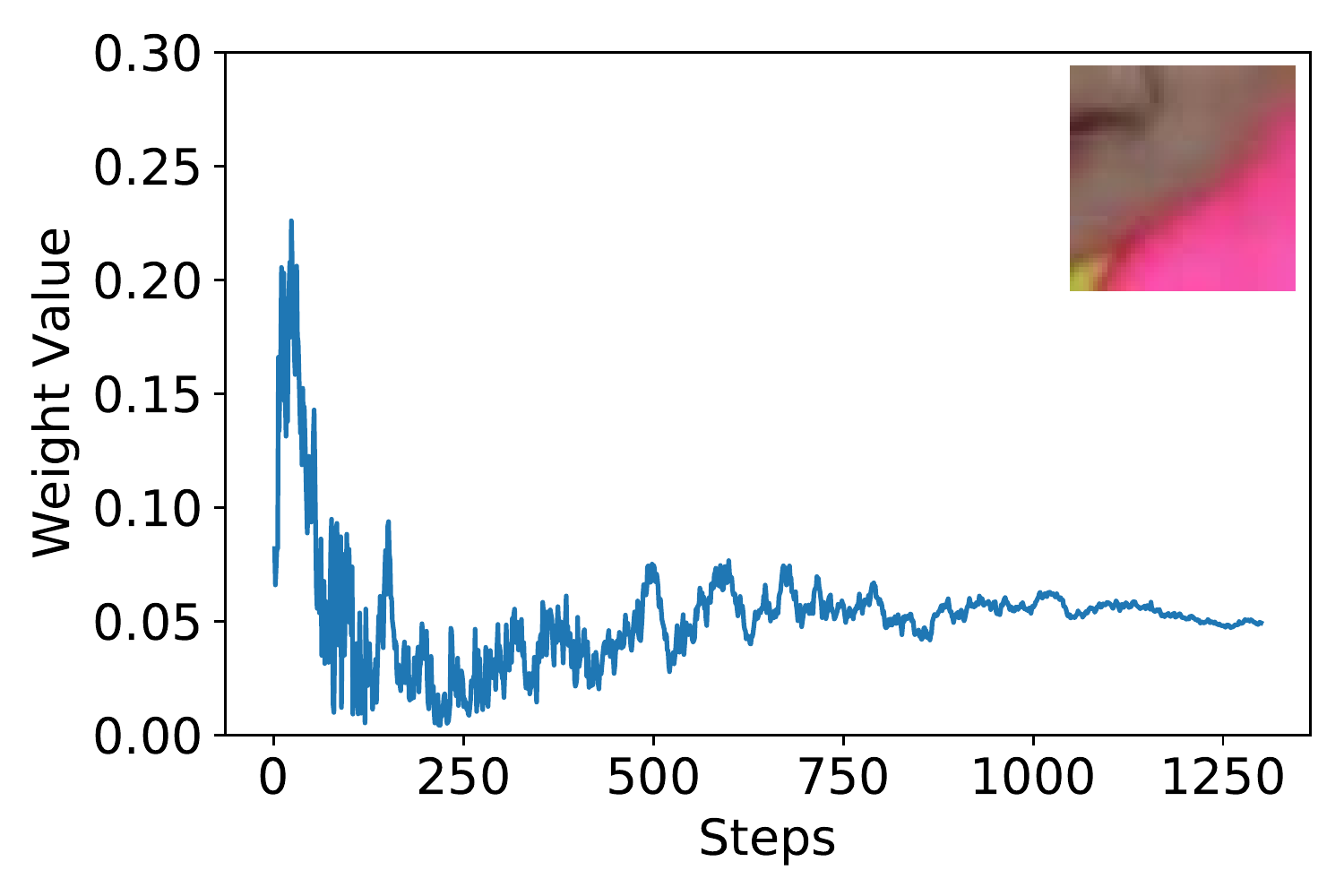}
		\end{minipage}
	}
	\caption{The change of weights ${\bf{w}}^g$ corresponding to 16 local regions in the training process of LNLAttenNet. The abscissa represents the number of iterations in the training process and the ordinate represents the magnitude of the weight corresponding to each iteration.}
	\label{weight16-plot}
\end{figure*}

\subsection{Analyses of Non-Local Attention}
In LNLAttenNet, it is achieved to adaptively enhance the feature learning of facial crucial regions by jointly optimizing for local and non-local parts, where the non-local attention network is constructed to obtain the global weights ${\bf{w}}^g$ of multiple local regions.
Actually, one purpose of our work is to explore how to automatically enhance the significance of local crucial regions in deep FER, while any landmarks are not given as the prior information of facial crucial regions.
Thus, in order to validate it, we make an analysis for the weights of 16 local regions obtained by our non-local attention for RAF-DB dataset.

First, the visualization results from 16 persons are shown in Fig.\ref{FigNonLocalWeights}. In Fig.\ref{FigNonLocalWeights}, the first and third rows show the original facial expression images, and the second and fourth rows exhibit the matrix (4$\times$4) of the final global weights ${\bf{w}}^g$ (16$\times$1) corresponding to 16 local regions.
For each matrix, the darker the color is, the higher the weight is.
From Fig.\ref{FigNonLocalWeights}, it is obvious that some crucial regions obtain higher weights and non-crucial regions get smaller weights for each facial expression.
For examples, the areas including or around eyes are given higher weights for the first person in the first row, where the maximum is given the local region located at the coordinate (2,2) including eyes.
For the sixth person in the first row, four local regions (located at (3,2), (3,3), (4,2), (4,3)) including his mouth are boosted and given higher weights.
In the third and fourth rows, the local regions located around eyes and the mouth are boosted for the second person, and the whole regions including eyes are given higher weights for the last person.
Visually, these enhanced local regions are more discriminative and significant for FER.

From Fig.\ref{FigNonLocalWeights}, it is also observed that the location of crucial regions is different for different facial images. But, our network still automatically tracks down more discriminative regions for each different face, without the supervision of any annotated crucial points. Based on this, secondely, we make an experiment to pursue the change of weights corresponding to each local region in the process of training our model.
Fig.\ref{FigChangWeights} shows the change of non-local weights in the training process.
In Fig.\ref{FigChangWeights}, the first row shows the original image and its final global weights obtained by our model, the second and third rows show the given global weights of 16 local regions in the initial, 250$^{th}$, 500$^{th}$, 750$^{th}$, 1000$^{th}$ and 1250$^{th}$ iterations, respectively, and the last row shows the final weights.
From Fig.\ref{FigChangWeights}, it is seen that the non-local weight of each local patch is same at the beginning of training, which implies that each local region is initially regarded as the equal importance.
With the training of our network, each local region is given different weights, and the higher weights are given some more discriminative regions, such as the patches (located at (4,2) and (4,3)) including the mouth shown in Fig.10(a), the patches (located at (3,2), (3,3), (4,2) and (4,4)) in Fig.10(b), et al..
It illustrates that some more crucial local regions can be adaptively enhanced in the training of our network without any landmarks.

\begin{figure*}[t]
	\centering
	\includegraphics[width=0.12\textwidth]{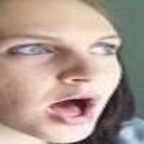}
	\includegraphics[width=0.12\textwidth]{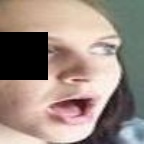}	
	\includegraphics[width=0.12\textwidth]{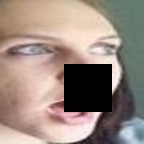}
	\includegraphics[width=0.12\textwidth]{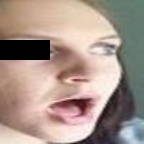}	
	\includegraphics[width=0.12\textwidth]{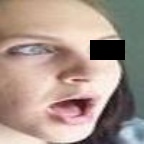}
	\includegraphics[width=0.12\textwidth]{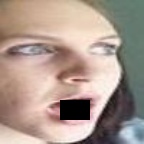}
	\includegraphics[width=0.12\textwidth]{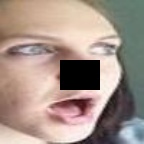}\\
	\includegraphics[width=0.12\textwidth]{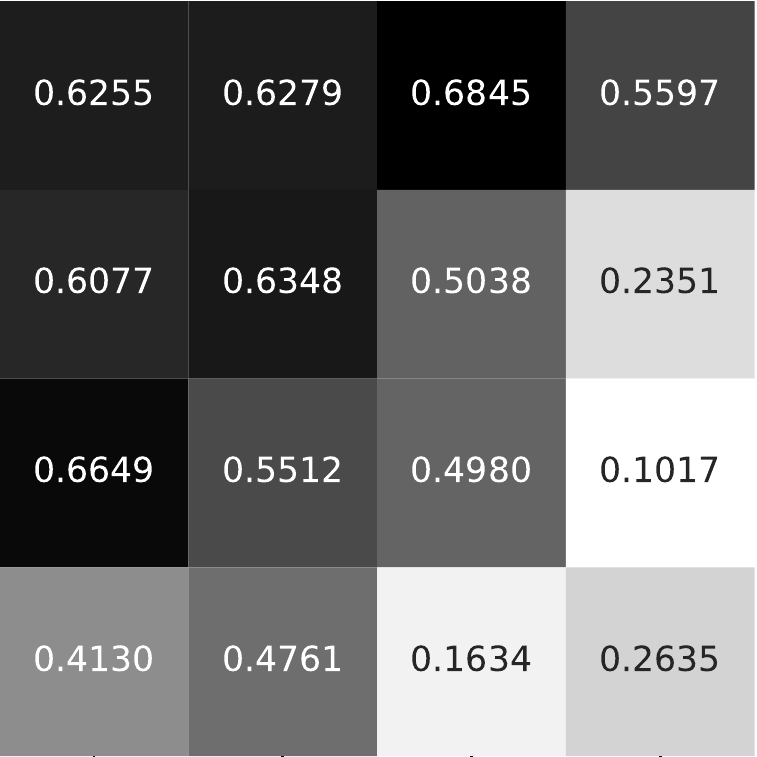}
	\includegraphics[width=0.12\textwidth]{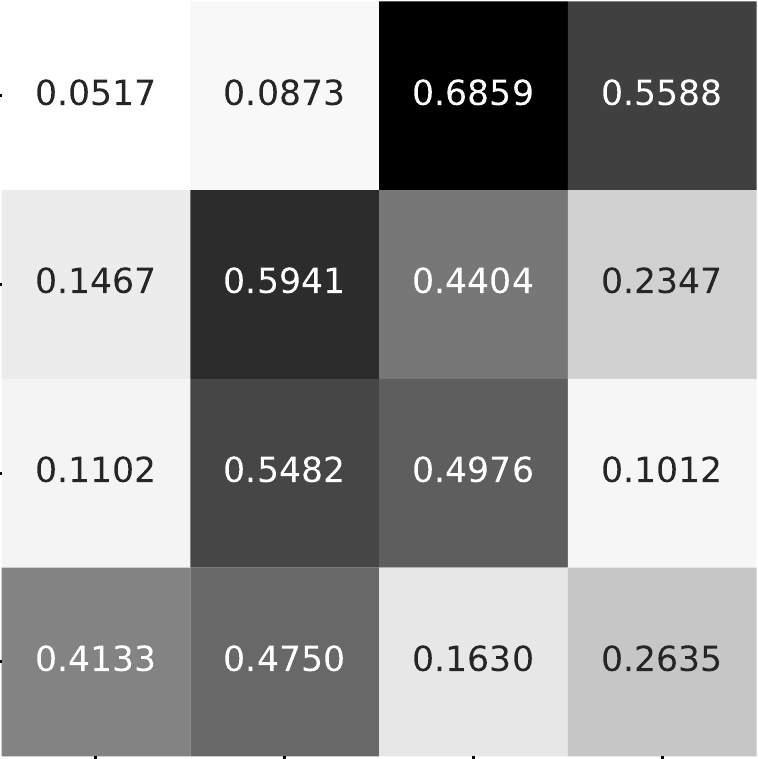}
	\includegraphics[width=0.12\textwidth]{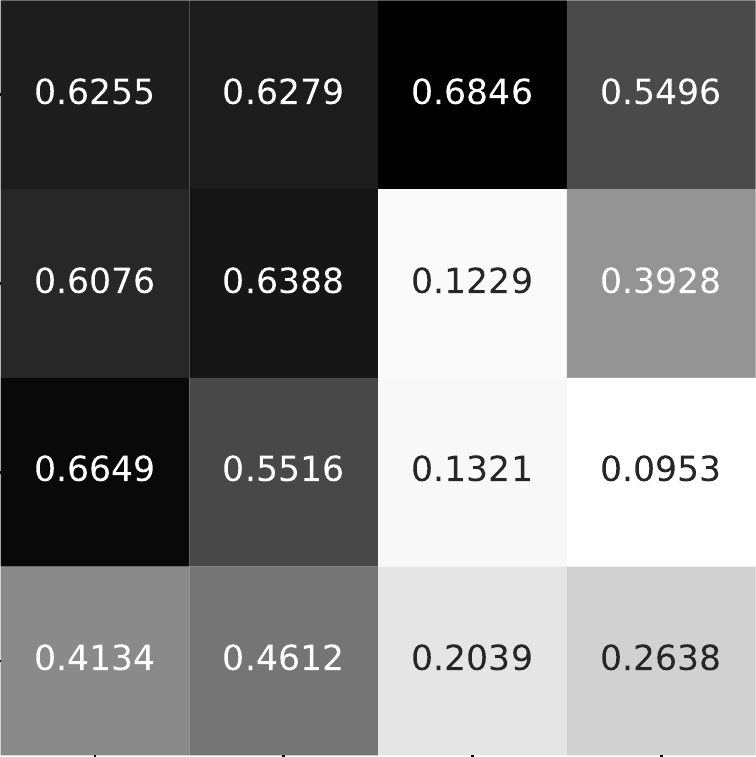}
	\includegraphics[width=0.12\textwidth]{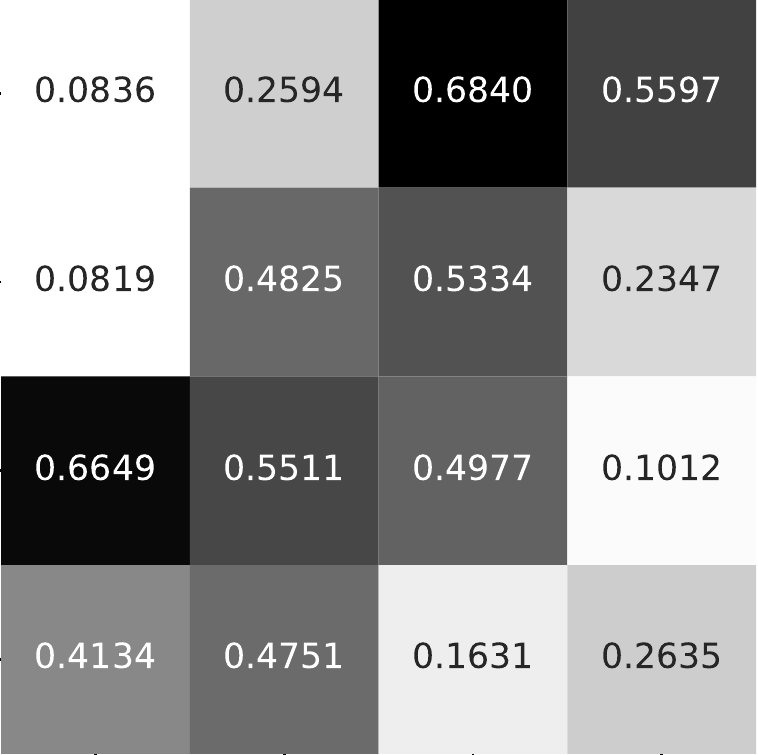}
	\includegraphics[width=0.12\textwidth]{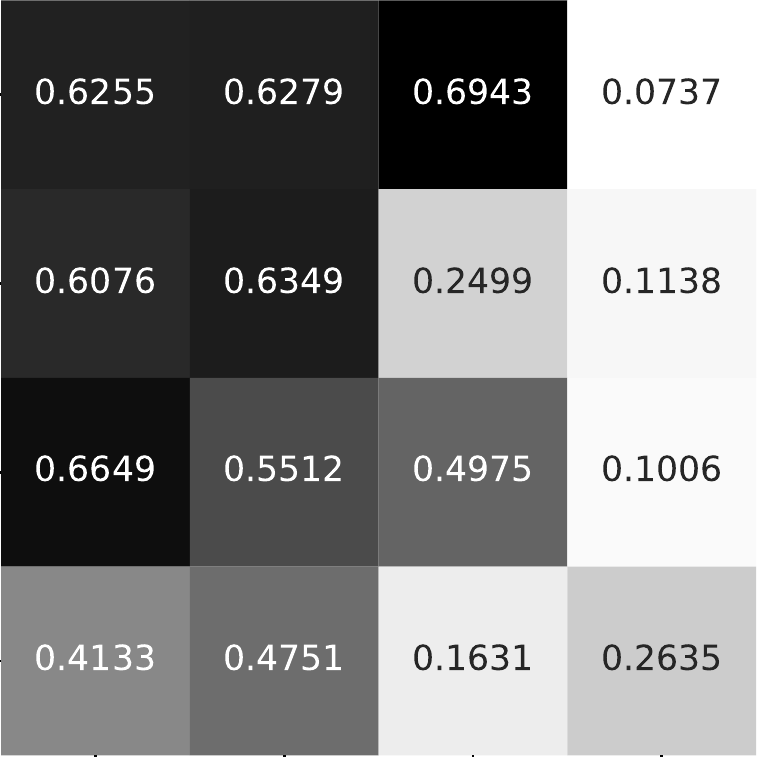}
	\includegraphics[width=0.12\textwidth]{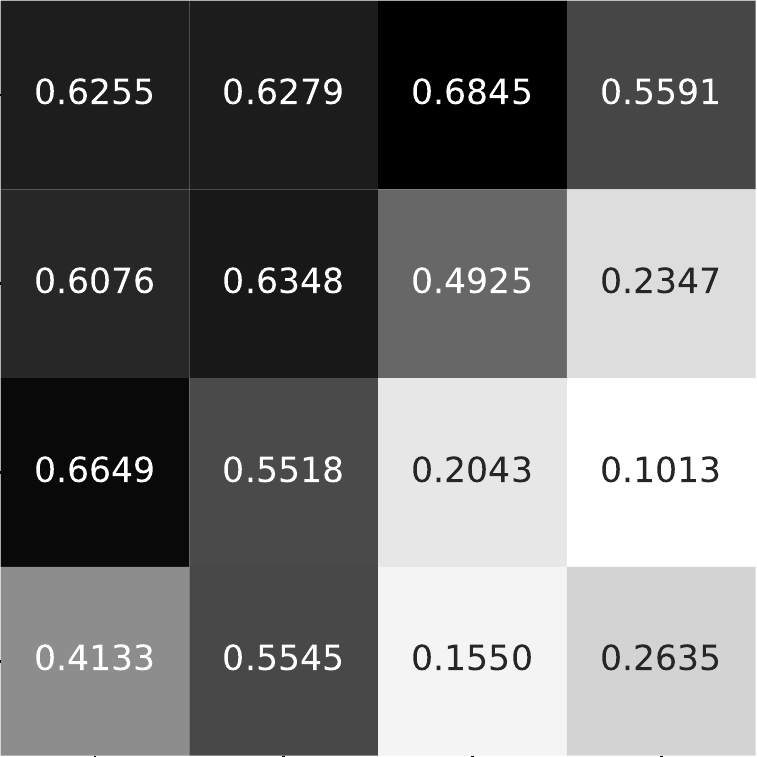}
	\includegraphics[width=0.12\textwidth]{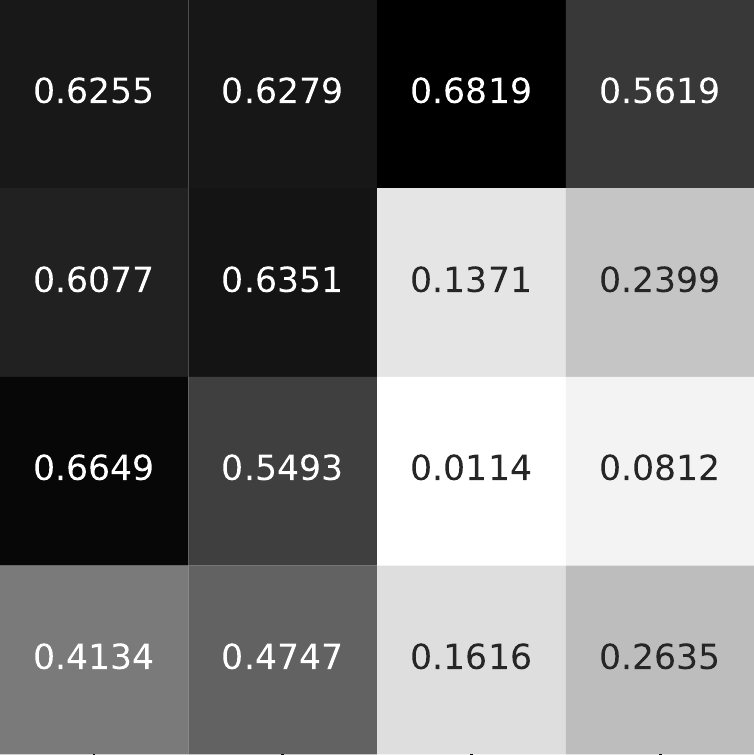}\\
	\includegraphics[width=0.12\textwidth]{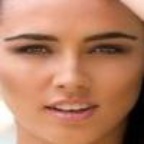}	\includegraphics[width=0.12\textwidth]{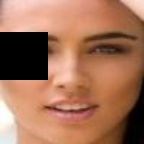}
	\includegraphics[width=0.12\textwidth]{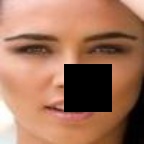}
	\includegraphics[width=0.12\textwidth]{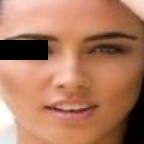}
	\includegraphics[width=0.12\textwidth]{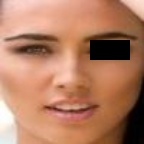}
	\includegraphics[width=0.12\textwidth]{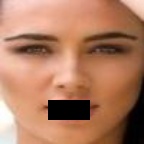}
	\includegraphics[width=0.12\textwidth]{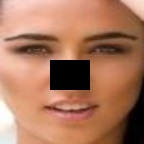}\\
	\includegraphics[width=0.12\textwidth]{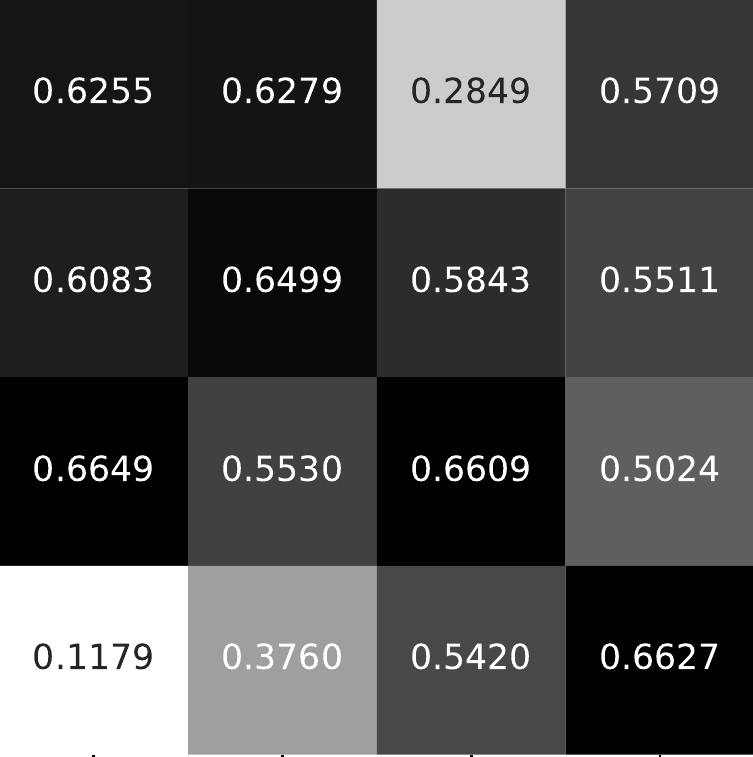}	\includegraphics[width=0.12\textwidth]{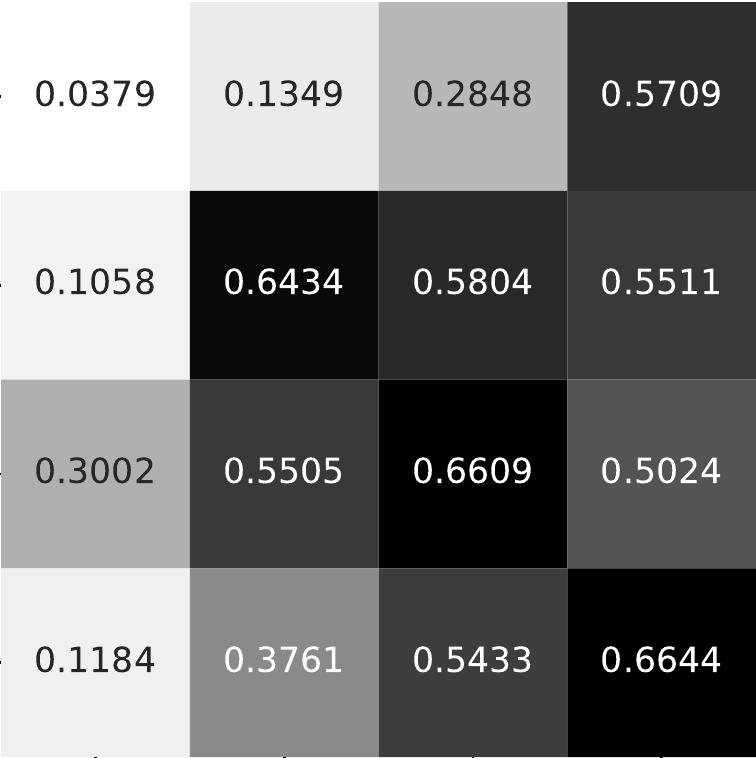}
	\includegraphics[width=0.12\textwidth]{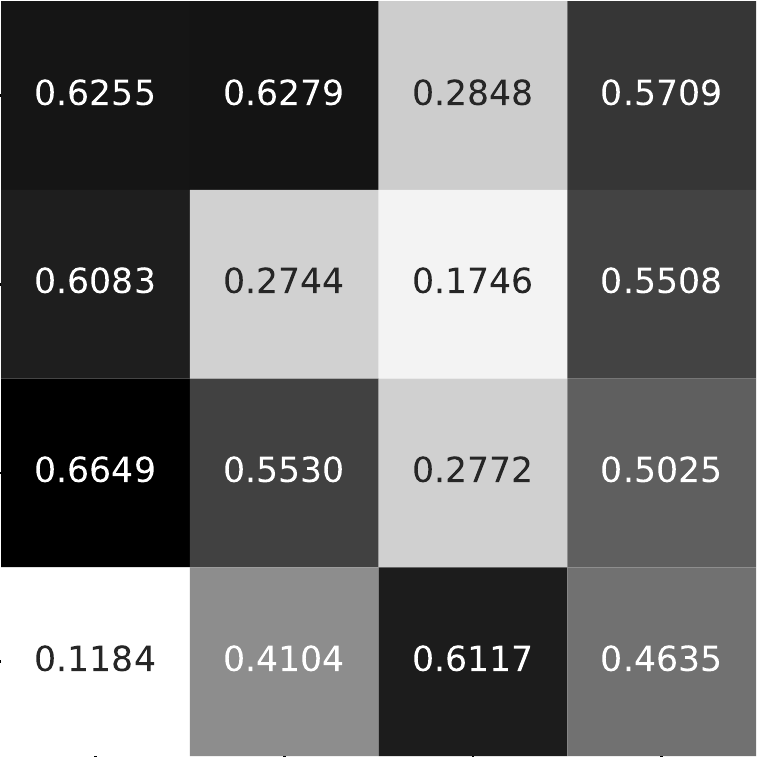}
	\includegraphics[width=0.12\textwidth]{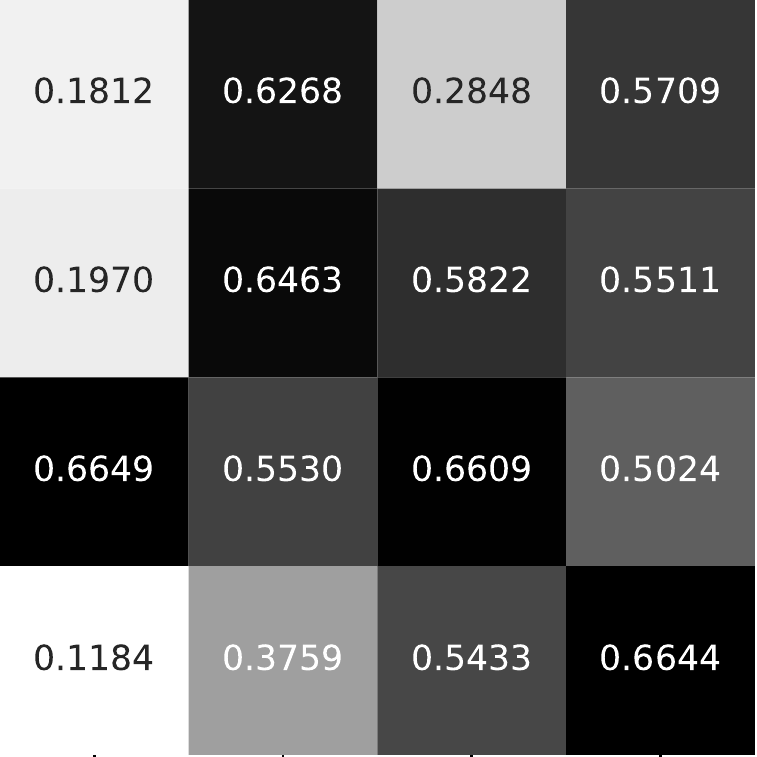}
	\includegraphics[width=0.12\textwidth]{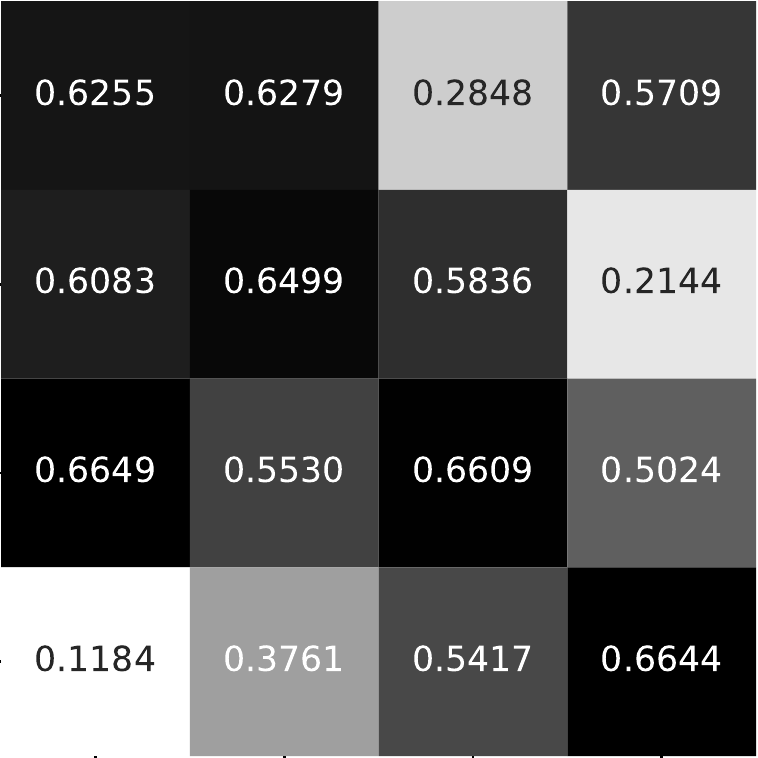}
	\includegraphics[width=0.12\textwidth]{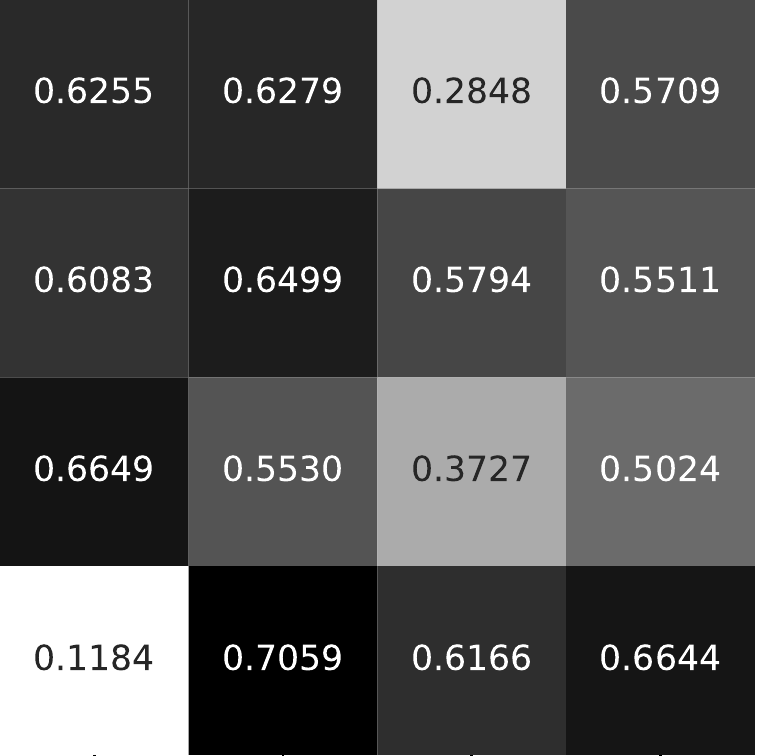}
	\includegraphics[width=0.12\textwidth]{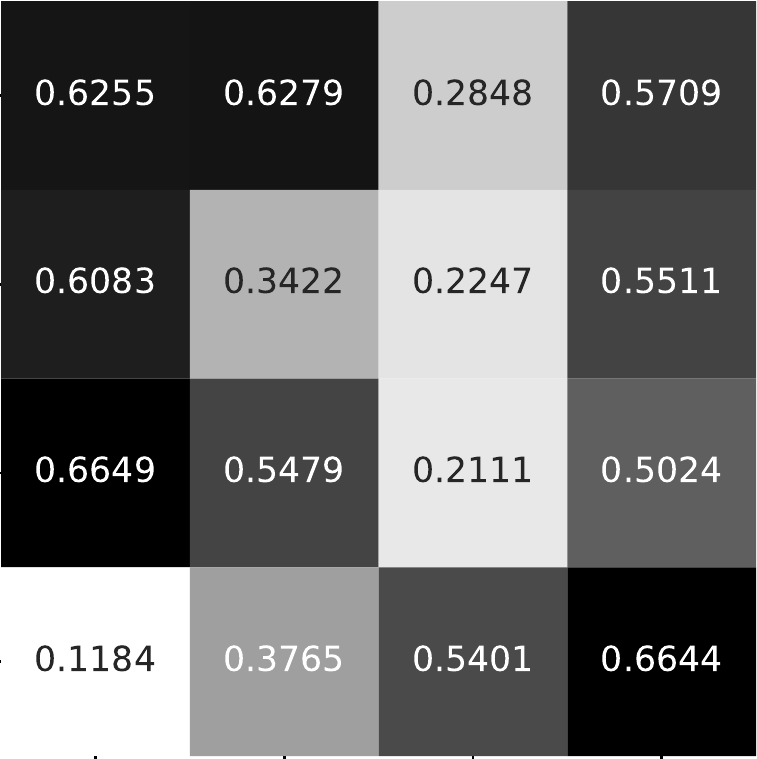}
	\caption{Local weights of 16 patches of each face on RAF-DB obtained by the proposed model. The first column shows the results corresponding to original images, and the second to seventh columns show the results corresponding to obscured images.}\label{fig5}
\end{figure*}

In order to better observe the change of weights, we also show the change of weights corresponding to 16 local regions in all iterations in Fig.\ref{weight16-plot}.
From Fig.\ref{weight16-plot}, it is seen that the weight value fluctuates at the beginning of network training and it is gradually stabilized until the end of the training.
Some patches that are visually more discriminative are lightened with higher weights and some patches located at the non crucial regions cut down with smaller weights.
In summary, the analyses for non-local weights demonstrate that the proposed method can effectively automatically enhance the significance of facial crucial regions in deep feature learning, without any given prior information of facial crucial regions.

\begin{figure}[h]
	\centering
	\includegraphics[width=0.5\textwidth]{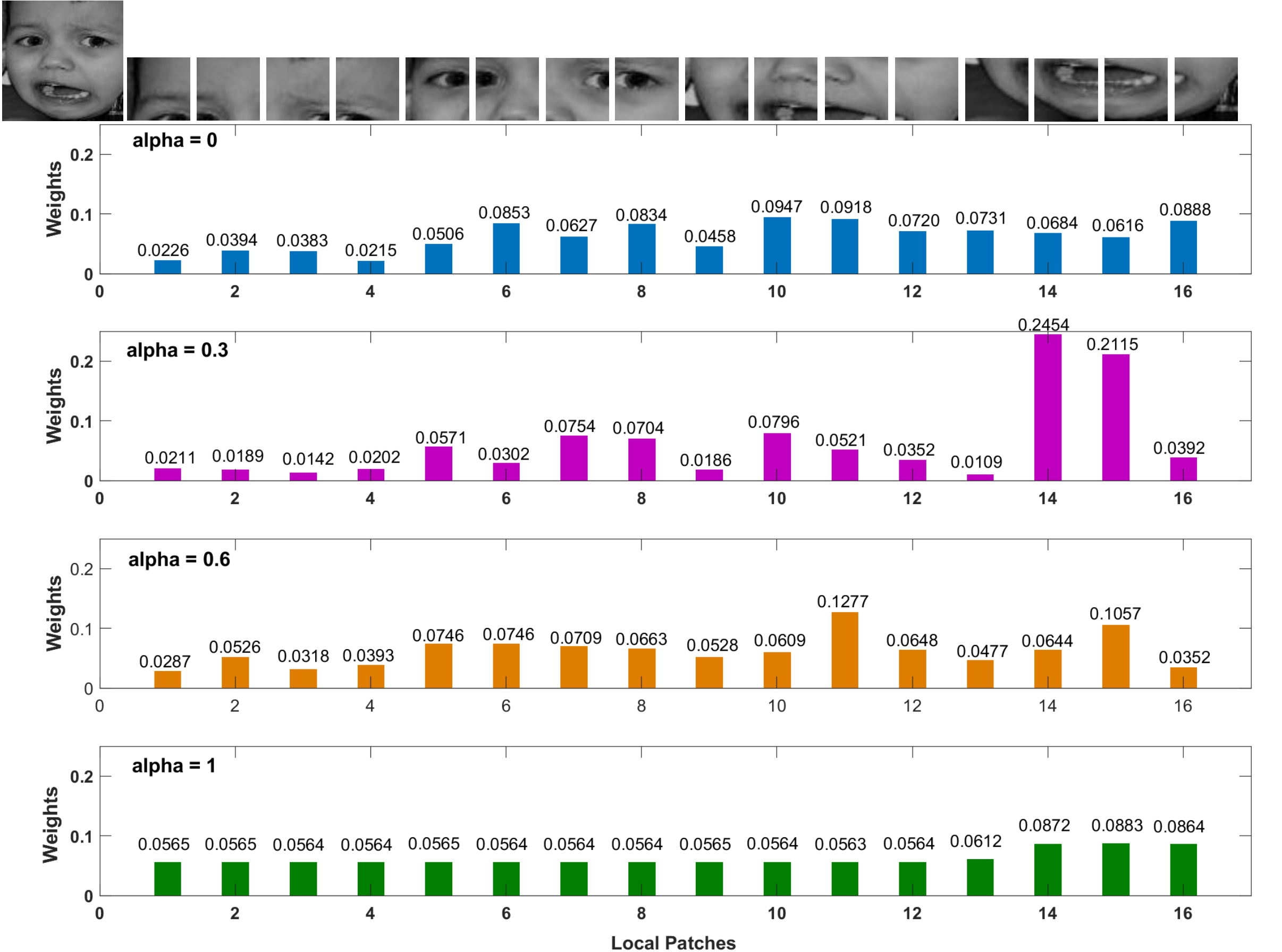}
	\caption{The change in the non-local weight at different $\alpha$}
	\label{figAlphaChange} 
\end{figure}

\subsection{Visualization of Local Attentions}
In the proposed method, the local attention is designed to deal with the problem that local regions is missed or obscured. In this part, the visualization of local attentions will be shown to validate the robustness of the proposed method for faces with missing regions, experimented on RAF-DB database. Note that the sigmoid function is employed to select the information flowing into the next layer in our local attention model.
Fig.\ref{fig5} shows visual results of local attentions obtained by our method.

In Fig.\ref{fig5}, the 1th and 3rd rows show one original facial image and six obscured images (from 2nd to 7th columns), and the 2nd and 4th rows show the weights of 16 patches of each facial image obtained by our method.
Compared with the result of the original images (shown in the first column of Fig.\ref{fig5}), it is found that the weight is weakened while one patch is obscured and the weights of other patches are unchanged.
Note that the weights of some adjacent patches are also decreased with the central patch, due to overlap pixels between two adjacent patches.
Practically, the local vector encoded based on one obscured patch is given a small weight, which effectively diminishes the influence of that obscured patch for facial expression recognition.
In short, the experimental results illustrate that the proposed method equipped with the local attention is more robust for complex facial expression databases in practice.

\subsection{Analyses for the parameter $\alpha$}
In the non-local attention network, we formulate Eq.(1) to obtain the non-local feature vector ${\bf{g}}^*$ based on the global information of facial expression, where the parameter $\alpha$ is used to traff off the feature vectors ${\bf{g}}$ and ${\bf{s}}$.
In the previous experiments, we set $\alpha = 0.7$.
Therefore, we make an analysis to observe the performance of the proposed method with different values of $\alpha$ in this part.
In this experiment, the experimental setups are same as the above experiments except $\alpha$, and $\alpha$ is set as \{0, 0.1, 0.2, ...,0.9, 1\}, respectively.
Table \ref{tab2} shows the accuracy under different $\alpha$ for five datasets.

From Table \ref{tab2}, it is seen that the accuracy is firstly increased and then decreased with a change in trend while increasing the value of $\alpha$.
According to Eq.(1), we get ${\bf{g}}^*={\bf{g}}$ if $\alpha =0$ and ${\bf{g}}^*={\bf{s}}$ if $\alpha =1$.
Combining the network optimization, it is known that the back propagation in LNLAttenNet has no constraint on ${\bf s}$ when $\alpha =0$, which implies that the same effect (or feedback) is given the non-local attention and each component of the non-local weights $\bf{w}^g$ should be random in theory.
On the contrary, $\alpha=1$ means that the back propagation has no constraint on the global vector ${\bf{g}}$, which means the back propagation in LNLAttenNet has no global information and may result in an extreme result.
Actually, as shown in Fig.\ref{figAlphaChange}, we also find that the obtained weights ($\bf{w}^g$) tend to be random under a small $\alpha$ and equal under a large $\alpha$, which effectively verifies the effect of $\alpha$ as same as the above analysis.

\begin{table*}
	\caption{Accuracy rates (\%) given by the proposed method with different $\alpha$.}
	\label{tab2}
	\begin{center}
		\begin{tabular}{|l|c|c|c|c|c|c|c|c|c|c|c|}
			\hline
			$\alpha$    &0      & 0.1    & 0.2    &0.3         & 0.4     & 0.5    & 0.6    & 0.7     & 0.8     & 0.9  & 1.0\\
			\hline\hline
			RAF         &84.09  & 85.60  & 85.69  &\bf{86.15}  & 85.59   & 85.33  &85.17   & 85.23   & 83.74   &83.54 & 83.02\\
			SFEW        &55.06  & 55.73  & 56.88  &\bf{57.80}  & 57.34   & 57.11  &56.65   & 56.88   & 55.96   &54.59 &53.67 \\
			CK+         &96.02  & 96.75  & 97.56   &98.18       & \bf{98.30}   & 97.74  &97.36   & 96.60   &96.22    &96.04 &95.28 \\
			MMI & 67.00 & 67.45 & 68.50 & 68.75 & \bf{68.88} & 68.25 & 67.50 & 67.38 & 66.93 & 66.50 & 66.25  \\
			AffectNet & 57.94 & 58.71 & \bf{59.43} & 59.28 & 58.03 & 57.80 & 56.83 & 56.86 & 56.71 & 56.66 & 56.63  \\ 			
			\hline
		\end{tabular}
	\end{center}
\end{table*}

\begin{table}
	\caption{Accuracy(\%) of the proposed method with different numbers ($M$) of patches.}
	\label{tab3}
	\begin{center}
		\begin{tabular}{|l|c|c|c|c|c|}
			\hline
			$M$           & 4    & 9    &16            & 25         & 36  \\
			\hline\hline
			RAF         & 84.97  & 85.66  &\bf{86.15}  & 85.53      & 85.63  \\
			SFEW        & 55.28  & 56.88  &57.80       & \bf{58.03} & 57.80     \\
			CK+        & 96.22   & 97.17  &\bf{98.18}  & 97.92       & 97.74   \\
			MMI & 67.60 & 67.90 & 68.75 & \bf{68.83} & 67.13 \\
			AffectNet & 58.06 & 58.43 & \bf{59.28} & 59.06 & 57.97 \\ 	
			\hline
		\end{tabular}
	\end{center}	
\end{table}

\subsection{Analyses for different M}
In our method, multiple individual networks are generated based on facial local regions, and the previous experiments are implemented with the number of local patches $M=16$. Therefore, we also make an analysis for the number ($M$) of local patches on five datasets. In this experiment, $M$ is set as 4, 9, 16, 25 and 36, respectively.
Table \ref{tab3} shows the accuracy rates with different $M$.
In this experiment, the size of the input image is 144*144 and the size of overlapping pixels between adjacent patches is around a third of the size of each patch, which is computed by
\begin{equation}
n*P_{size} - (n-1)*\gamma * P_{size} = 144,
\end{equation}
where $\gamma$ is around $1/3$, $n^2 = M$ and $P_{size}$ is the size of each patch.
Note that the parameters of our network except $M$ is set as same as previous experiments.

From Table \ref{tab3}, it is observed that the performance with more local regions is superior to with less local regions. It implies that the size of each local region is too large to attain multiple diverse local information when $M$ is set as a small value. Whereas, it is also notice that the computational complexity will be increased when $M$ is set as a high value, and thus we finally set $M=16$ to implement most experiments.

 \subsection{Analyses for Overlapped Pixels between Local Regions}
In the previous experiments, $1/3$ of whole pixels in each patch are applied as the overlapping pixels between two neighbor patches, which is a more appropriate value, since the number of pixels overlapping between the middle patch and both sides is only $2/3$ , and the information of $1/3$  of the pixels at the center of patch is still retained. If a larger number of overlapping pixels is employed, such as $1/2$, the middle patch will completely overlap with the patches on both sides. If a smaller number is used, such as $1/4$, the number of pixels in the overlapping region will be too small to solve the problem of regional connectivity. In order to analyze the influence of overlapping pixels between two patches, an experiment that other experimental settings are same to before is implemented based on RAF-DB dataset, and the result is shown in Table \ref{tab4}.
	
In Table \ref{tab4}, it shows accuracies obtained by the proposed method based on different number ($N$) of overlapping pixels.  From the results, it is seen that the performance on the test set increases slowly to plateau as the number of overlapping pixels increases. It illustrates that the more the overlapping pixels are, the larger the number of network parameters are. According to our analyses, the main reason is that it is easier to introduce redundant information between adjacent patches when the number of overlapping pixels is larger.

\begin{table}
	\caption{Accuracy(\%) of the proposed method with different overlapping numbers(N) of pixels.}
	\label{tab4}
	\begin{center}
		\begin{tabular}{|l|c|c|c|c|c|c|}
			\hline
			$N$           & 4      & 8     &12      &16        & 20      & 24        \\
			\hline
			RAF           & 84.63  & 84.96 &85.29   & 86.15    & 86.16   & 86.24   \\
			\hline
		\end{tabular}
	\end{center}	
\end{table}

\section{Conclusion}\label{Conclusion}
In this paper, we propose the LNLAttenNet method to effectively explore the significance of facial crucial regions in feature learning for FER, without any landmark information.
In LNLAttenNet, the global information of the facial expression is utilized to construct the non-local attention network, and meanwhile the local information is utilized to supervise self-information.
By the joint optimization of facial non-local and local feature vectors, LNLAttenNet can adaptively enhance more crucial regions in the process of deep feature learning.
Specifically, an ensemble of multiple networks corresponding to local regions is constructed to integrate the local feature with the non-local weights, which achieves the interactive guidance between the facial global and local information.
Experimental results also demonstrate that some local crucial regions can be effectively enhanced in feature learning by LNLAttenNet while there are not any given information of landmarks in the training model.
Moreover, the proposed method focuses on enhancing facial crucial regions in FER without any landmark information based on multiple patches, and thus we will explore it from the view of pixels for facial expressions in the further works.

\ifCLASSOPTIONcaptionsoff
  \newpage
\fi

\bibliographystyle{IEEEtran}
\bibliography{Reference}

\begin{IEEEbiography}[{\includegraphics[width=1in,height=1.25in,clip,keepaspectratio]{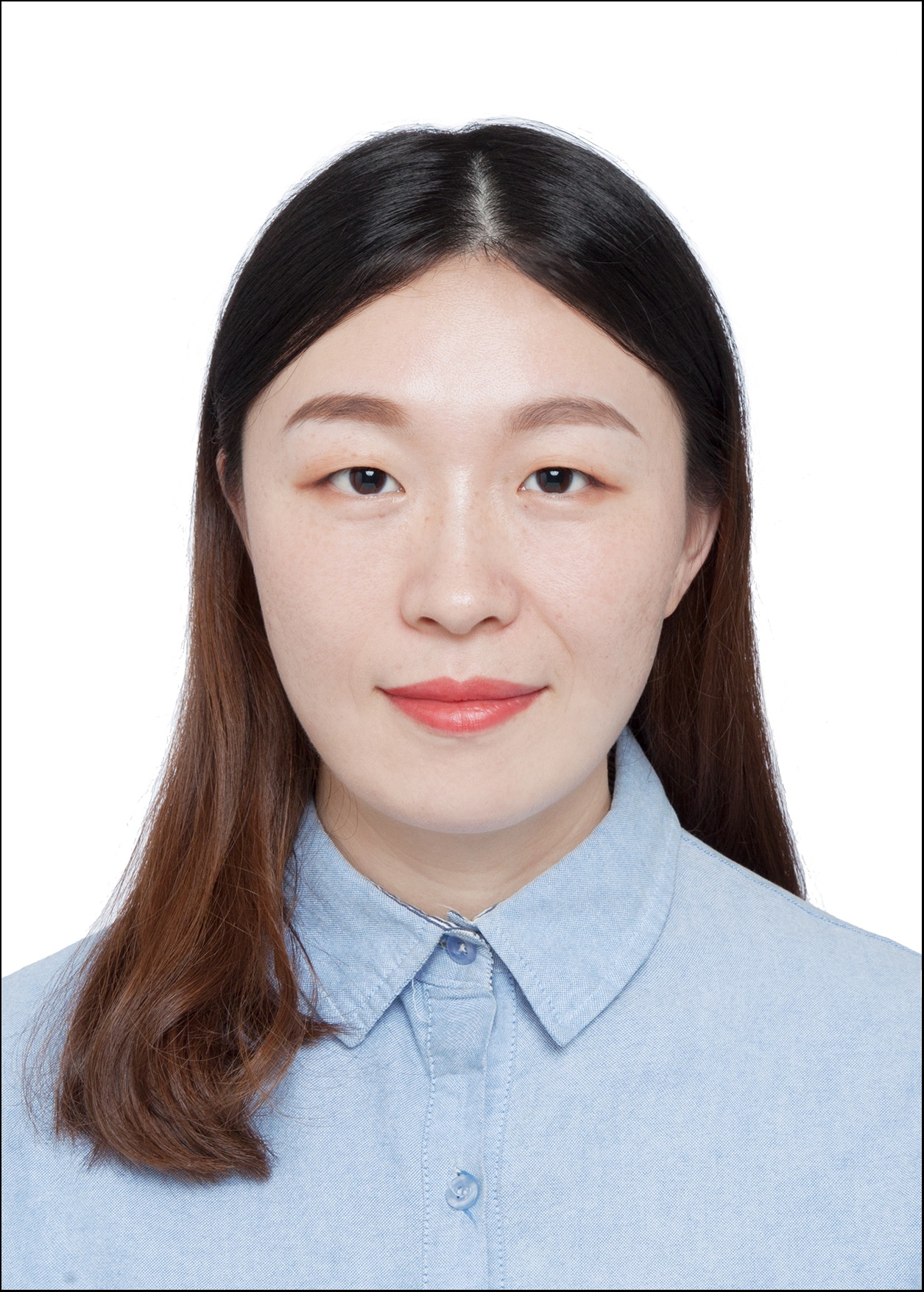}}]{Shasha Mao}
	(M'14) received the Ph.D. degree in circuit and system from Key Lab of Intelligent Perception and Image Understanding of Ministry of Education, Xidian University, Xi?an, China, in 2014.  From 2014 to 2018, she worked as a Research Fellow in Nanyang Technological University and Singapore University of Technology and Design, Singapore, respectively. She is currently an Associate Professor at the school of Artificial Intelligence, Xidian University. Her research interests include ensemble learning, deep learning, imbalanced learning, facial expression recognition, and SAR images regestration.	
\end{IEEEbiography}

\begin{IEEEbiography}[{\includegraphics[width=1in,height=1.25in,clip,keepaspectratio]{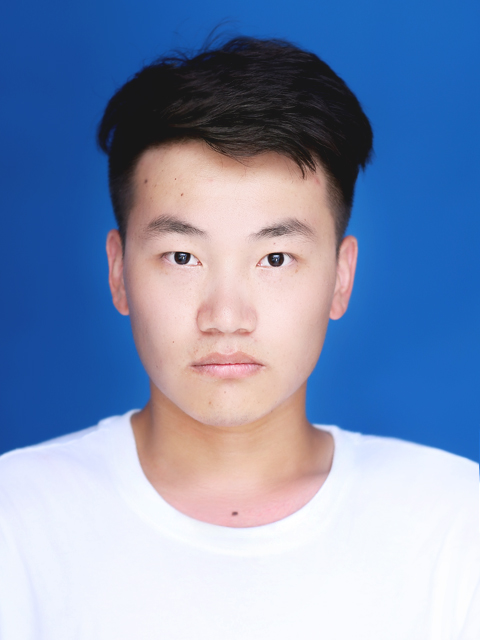}}]{Guanghui Shi}
	Guanghui Shi received the B.S. degree in Electronics and Information Engineering from Wuhan University of Technology in 2018 and received the M.S. degree in Electronics and Communication Engineering from Xidian University, Xian, China in 2021. He is currently working at the 701 Research Institute. His research interests include machine learning, deep learning and facial expression recognition.	
\end{IEEEbiography}

\begin{IEEEbiography}[{\includegraphics[width=1in,height=1.25in,clip,keepaspectratio]{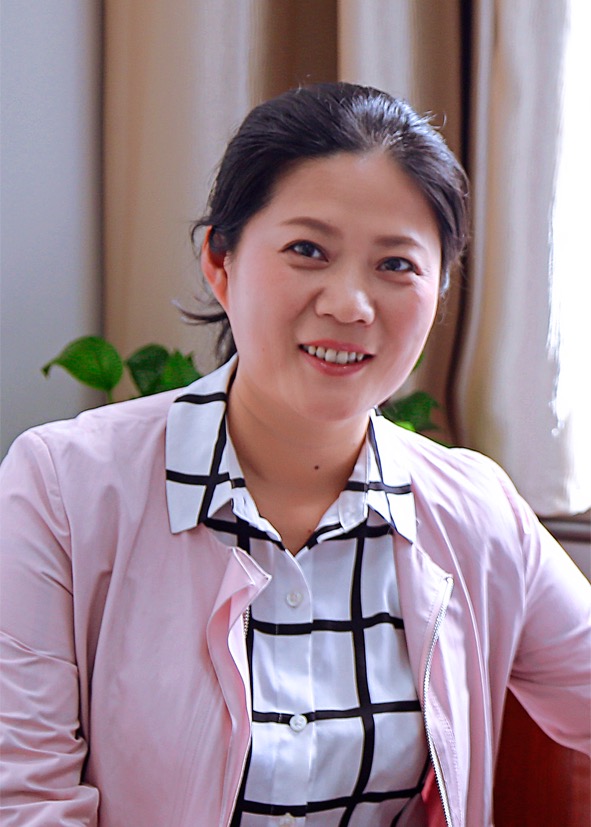}}]{Shuiping Gou}
	(M'08) received the B.S. and M.S. degrees in computer science and technology from Xidian University, Xi'an, China, in 2000 and 2003, respectively, and the Ph.D. degree in pattern recognition and intelligent system from Xidian University, in 2008. She is currently a Professor with the Key Laboratory of Intelligent Perception and Image Understanding of Ministry of Education of China, School of Artificial Intelligence, Xidian University. Her research interests include machine learning, data mining, remote sensing image analysis and medical image analysis.
\end{IEEEbiography}

\begin{IEEEbiography}[{\includegraphics[width=1in,height=1.25in,clip,keepaspectratio]{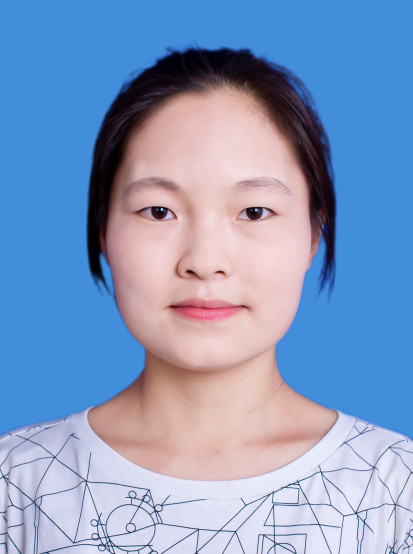}}]{Dandan Yan}
	received the B.S. degree in Computer Science and Technology from Xi?an University of Technology in July 2021. She is currently a student at the School of Artificial Intelligence, Xidian University. Her research interests include deep learning, facial expression recognition and label distribution learning.
\end{IEEEbiography}	

\begin{IEEEbiography}[{\includegraphics[width=1in,height=1.25in,clip,keepaspectratio]{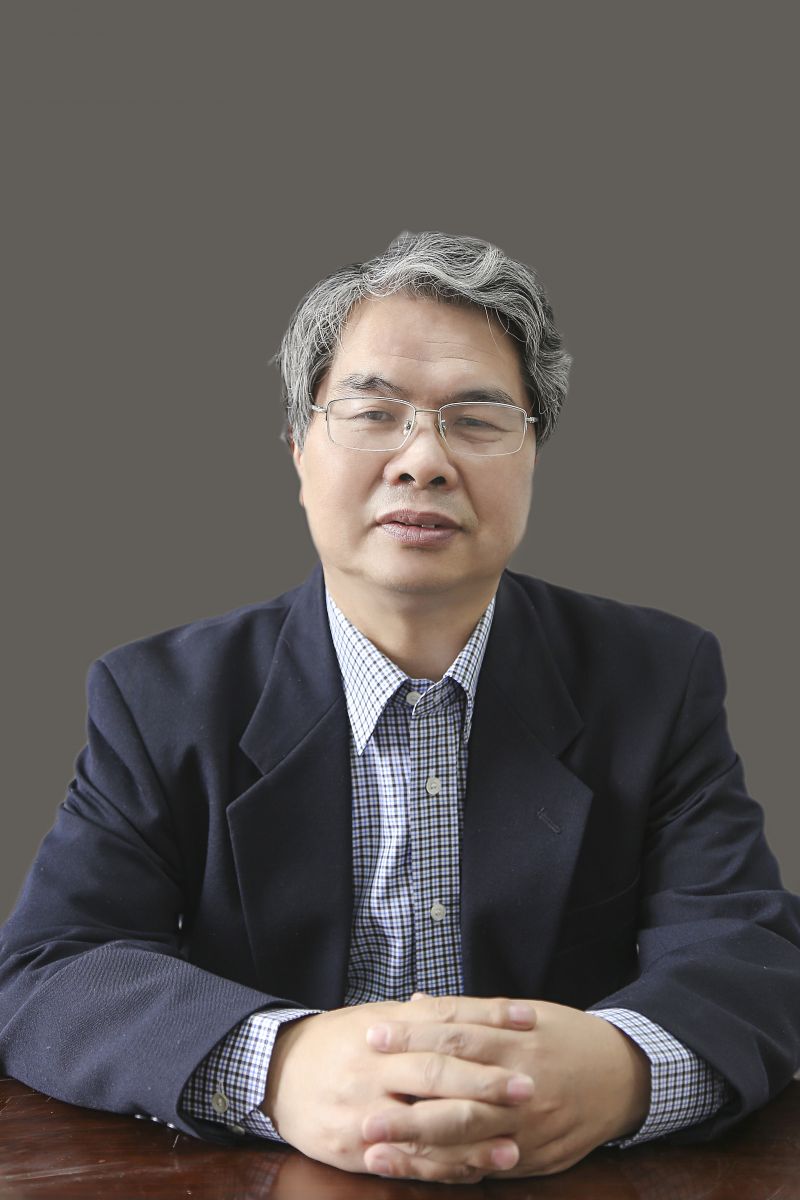}}]{Licheng Jiao}	
	(SM'89---F'17) received the B.S. degree in electronic engineering from Shanghai Jiao Tong University, Shanghai, China, in 1982, the M.S. and Ph.D. degrees in electronic engineering from Xian Jiaotong University, Xi?an, China, in 1984 and 1990, respectively. From 1990 to 1991, he was a Post-Doctoral Fellow with the National Key Laboratory for Radar Signal Processing, Xidian University, Xi?an. Since 1992, he was a Professor with the School of Electronic Engineering, Xidian University. Currently, he is a Professor with the School of Artificial Intelligence, Xidian University, and he is  also the Director of the Key Laboratory of Intelligent Perception and Image Understanding, Ministry of Education of China, Xidian University. He is in charge of about 40 important scientific research projects. He has authored or co-authored more than 20 monographs and 100 papers in international journals and conferences. His research interests include image processing, natural computation, machine learning, and intelligent information processing.
	Prof. Jiao is a member of the IEEE Xian Section Execution Committee, the Chairman of awards and recognition committee, the Vice Board Chairperson of the Chinese Association of Artificial Intelligence, the Councilor of the Chinese Institute of Electronics, the Committee Member of the Chinese Committee of Neural Networks, and an Expert of academic degrees committee of the state council.	
\end{IEEEbiography}

\begin{IEEEbiography}[{\includegraphics[width=1in,height=1.25in,clip,keepaspectratio]{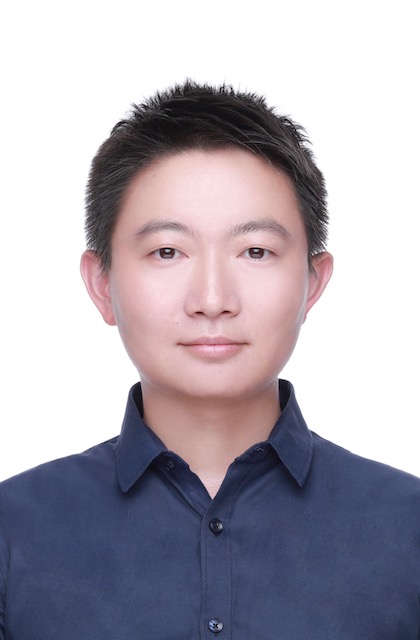}}]{Lin Xiong}
	received the Ph.D. degree in pattern recognition \& intelligent system from Key Lab of Intelligent Perception and Image Understanding of Ministry of Education, Xidian University, Xi'an, China, in 2015.
	Currently, he works as research scientist in JD Finance America Corporation. Before, he was a senior research engineer of Learning \& Vision, Core Technology Group, Panasonic R\&D Center Singapore (PRDCSG) from 2015 to 2018. His research interests include distributed model parallelism, unconstrained/large-scale face recognition, deep learning architecture engineering, person re-identification, face recognition, Riemannian manifold optimization, sparse and low-rank matrix factorization.
\end{IEEEbiography}		

\end{document}